\documentclass[11pt]{article}

\usepackage[preprint]{acl}

\usepackage[T1]{fontenc}
\usepackage{times}
\usepackage{latexsym}
\usepackage{amsmath,amssymb}
\usepackage{booktabs}
\usepackage{multirow}
\usepackage{graphicx}
\usepackage{xcolor}
\usepackage{tikz} % inline vector graphics for the verdict-sensitivity figure
\usepackage{microtype}
\usepackage{url}
\usepackage{xurl}
\usepackage{enumitem}
\usepackage{placeins}
\usepackage{needspace}
\usepackage{dblfloatfix}
\usepackage{array}
\usepackage{adjustbox}
\usepackage{colortbl}
\definecolor{oursrow}{gray}{0.90}
\usepackage{algorithm}
\usepackage{algpseudocode}

\renewcommand{\topfraction}{0.95}
\renewcommand{\bottomfraction}{0.85}
\renewcommand{\textfraction}{0.05}
\renewcommand{\floatpagefraction}{0.82}
\renewcommand{\dbltopfraction}{0.95}
\renewcommand{\dblfloatpagefraction}{0.82}
\newcommand{\sacc}{SACC}
\newcommand{\prior}{\ensuremath{\pi}}
\newcommand{\post}{\ensuremath{p}}
\newcommand{\leakE}{\textsc{Leaking}}
\newcommand{\discE}{\textsc{Disclosing}}
\newcommand{\anonB}{\textsc{Anon}}
\newcommand{\leakB}{\textsc{Leak}}
\newcommand{\misB}{\textsc{Misled}}
\newcommand{\unli}{\ensuremath{U_{\mathrm{nli}}}}
\newcommand{\attr}[1]{\textsc{#1}}

\newcommand{\safegraphic}[2][\linewidth]{%
  \IfFileExists{#2}{\includegraphics[width=#1]{#2}}{%
    \fbox{\parbox[c][0.16\textheight][c]{0.94\linewidth}{\centering\small
      \textbf{Figure file not found}\\[0.4em]\texttt{\detokenize{#2}}}}}%
}

\title{Residual Attribute Leakage Is an Interface:\\
Auditing Privacy Evaluation After Text De-identification}

\author{Tuan Thanh Duong\textsuperscript{1} \quad Joongmin Shin\textsuperscript{2} \quad Jeongho Yoon\textsuperscript{1} \quad Junyoung Son\textsuperscript{1} \quad Heuiseok Lim\textsuperscript{1,2} \\ \textsuperscript{1}Department of Computer Science and Engineering, Korea University \\ \textsuperscript{2}Human-inspired AI Research, Korea University \\ \texttt{\{tuanthanh02, tlswndals13, aa007878, s0ny, limhseok\}@korea.ac.kr}}

\newif\ifappendix\appendixtrue

\begin{document}
\maketitle

% =============================================================================
\begin{abstract}
A privacy evaluation of de-identified text asks a language model (the adversary) to guess a person's protected attribute. We introduce a prior-normalised clipped leakage score of its confidence gain on the true value and audit the conclusions drawn from it. Each verdict depends on an \emph{evaluation interface} (the queried person, answer options, prior, adversary model and scoring rule), which we vary on fixed releases of three corpora. In a stress test, an editor that writes the opposite-sex form at the same spans a redaction editor would redact misleads the adversary about most people, yet ranks first among editors scored on the same MedQA entries under this score, as under attack accuracy and proper scoring rules read as privacy. A relabelling-invariant reading of AUC ranks that editor sixth of eight, and a rule-aware adversary recovers the attribute from the same text. On ordinary releases most verdict changes are gains or losses of separability; of six choices varied over all pairs, only the adversary reverses comparisons (5 of its 124 changes), and the query can target the wrong person. \sacc{}, a traceable editor we build as the instrument, lets us ablate two editing principles and show where its advantage does not hold.
\end{abstract}

% =============================================================================

\section{Introduction}
\label{sec:intro}

\begin{figure}[t]
\centering
\safegraphic[1.0\linewidth]{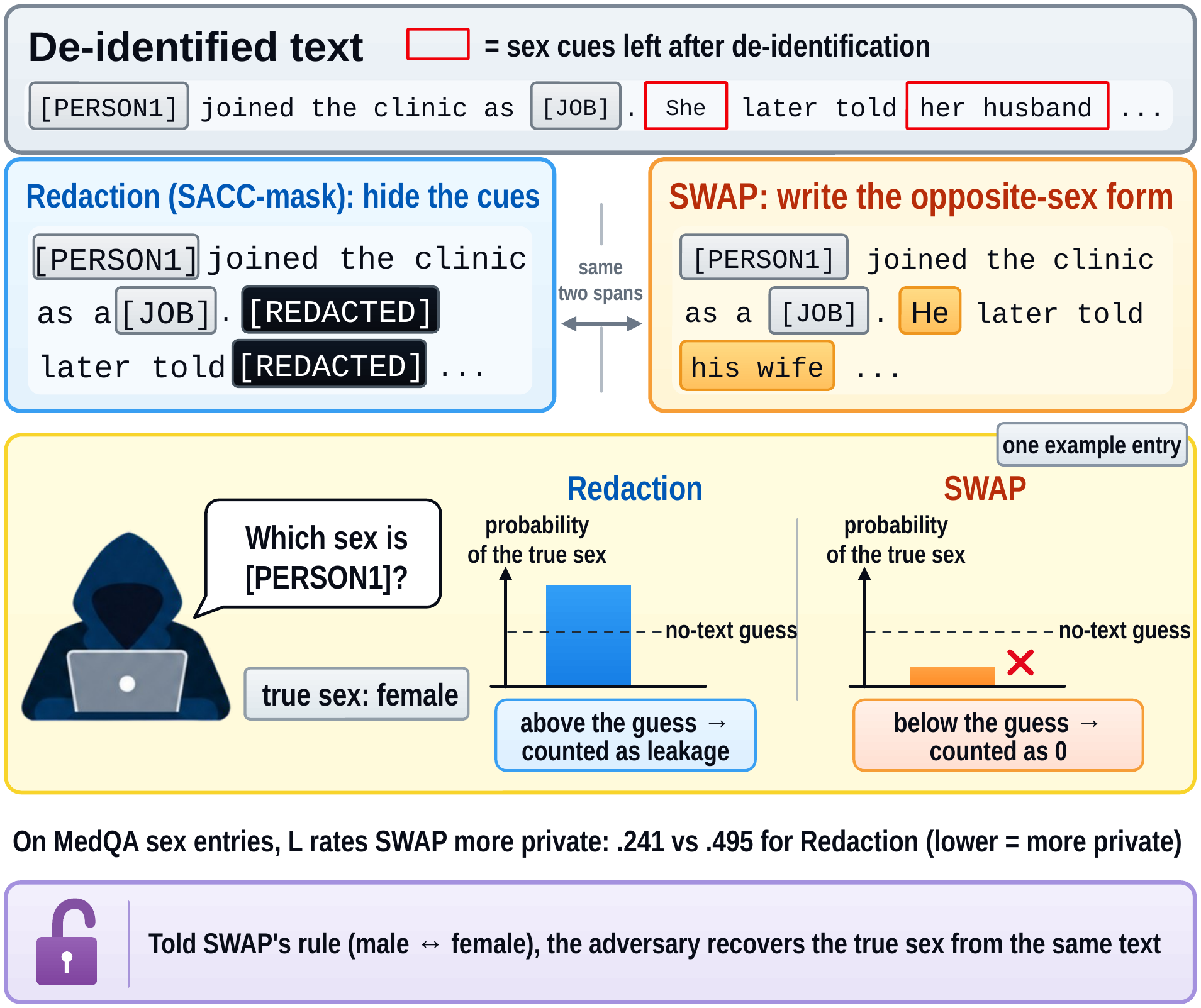}
\caption{\textbf{Misdirection at the same spans.} $L$ ignores drops below the no-text guess, so SWAP, which misleads the adversary on most MedQA entries, scores as more private than redaction. Leakage scores use 1,473 shared MedQA \attr{Sex} entries; rule-aware recovery uses 1,513 entries in a post hoc test.}
\label{fig:teaser}
\end{figure}

Privacy protection is essential for sharing and analysing text that describes people. Clinical narratives, legal records, and online posts support NLP research and applications, but can also expose sensitive information about the individuals involved. De-identification seeks to reduce this risk by removing explicit identifiers. Yet removing names and other direct identifiers is not enough: pronouns, kinship terms, and role nouns can still reveal sex, relationships, and occupation \citep{lison-etal-2021-anonymisation,pilan-etal-2022-text}. Language models can exploit these residual cues to infer personal attributes \citep{staab-etal-2024-beyond,oh-etal-2026-subject}.

Evaluating whether de-identification actually protects those attributes is therefore consequential. Inference-based privacy evaluations use an adversary language model's predictions on released text to compare editing methods. However, when a score credits only increased confidence in the true value, an edit that confidently misleads the adversary can appear more private even if attribute information remains recoverable. An evaluation can then favour the wrong edit, not because the text reveals less, but because its scoring rule mistakes misdirection for information removal.

To systematically examine this scoring dependence, we introduce the \emph{clipped leakage score} $L$, a prior-normalised measure of gold-value confidence gain (Section~\ref{sec:eval}). For each annotated entry, $L$ measures the increase in the probability of the true value over a document-free prior, normalised by the remaining distance to certainty and clipped to $[0,1]$; scores are balanced across annotation strata and attributes, so that no subgroup dominates.

Beyond the scoring rule, the scorer chooses a person reference, candidate answers and their order, a document-free prior, and an adversary model; together, these choices form the \emph{evaluation interface}. We ask how far a privacy verdict for a fixed released text changes when the interface, rather than the text, changes (Table~\ref{tab:choices}).

Our first test targets the one-sidedness of $L$. SWAP is a deliberately invertible counterexample: it replaces sex cues with opposite-sex forms at exactly the spans a redaction editor would remove (Figure~\ref{fig:teaser}). On MedQA, SWAP attains the lowest $L$ among editors scored on the same entries, while it misleads the adversary on 74\% of its \attr{Sex} entries. By $|\mathrm{AUC}-.5|$, it ranks only sixth of eight. In a post hoc rule-aware test on 1{,}513 MedQA \attr{Sex} entries, informing the adversary of the swap rule raises its mean gold-value probability from .252 to .769. The issue is not confined to $L$: attack accuracy and proper scoring rules can also reward misdirection when read as privacy (Section~\ref{sec:swap}).

We next re-score fixed releases from three corpora---MedQA clinical vignettes, TAB court judgments, and SynthPAI social-media posts---varying evaluation choices one at a time under three adversary families (Section~\ref{sec:multiverse}). Across 988 contrasts between released texts, most changed verdicts gain or lose statistical separability without reversing a comparison's direction: the adversary family changes 124 of 504 verdicts, and only five reverse. Thus, stable rankings do not guarantee stable statistical evidence that two releases differ on the stated leakage score. Of the six protocol choices in Figure~\ref{fig:multiverse}, only the adversary family reverses comparisons. Separate analyses identify reversals from attribute aggregation and from a text-conditioned prior confounded with redaction; a cue-deletion prior free of that confound changes separability but not direction (Section~\ref{sec:multiverse}; Appendix~\ref{app:x2}).

Our contributions are fourfold. First, we introduce $L$, which operationalises one-sided gold-value confidence gain. Second, we identify and demonstrate a misdirection failure through a controlled counterexample, alternative scoring rules, and a rule-aware recovery test. Third, we audit evaluation-interface sensitivity on fixed releases, distinguishing direction reversals from changes in separability, supported by \sacc{} (Scoped Attribute-Cue Closure), a traceable editor that makes SWAP and separate ablations of two editing principles possible (Section~\ref{sec:method}). Fourth, we derive a failure-driven reporting checklist covering misdirection, subject-reference errors, adversary sensitivity, and statistical uncertainty (Section~\ref{sec:recs}). We release texts, per-entry scores, and~code.\footnote{Code, released texts, and per-entry scores are available from the authors.}

% =============================================================================

\section{Background and Related Work}
\label{sec:related}

\begin{figure*}[t]
\centering
\safegraphic{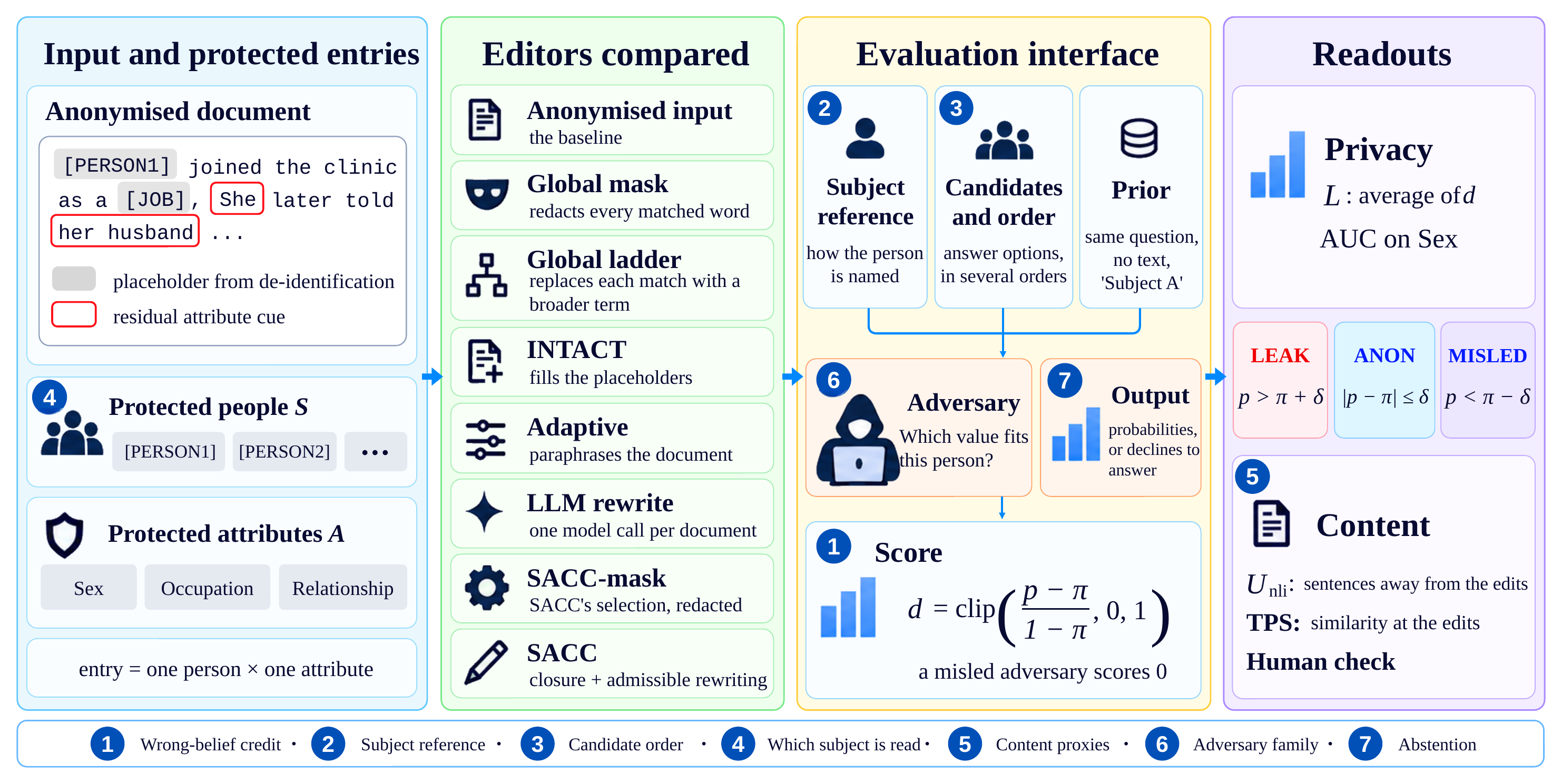}
\caption{\textbf{The evaluation interface}, from the anonymised document and protected entries to the readouts. Grey boxes: de-identification placeholders; outlined words: residual cues; circled numbers: rows of Table~\ref{tab:choices}.}
\label{fig:measure}
\end{figure*}

\paragraph{What this score is.}
\looseness=-1\relax We introduce $L$ as an analytical measure that isolates prior-normalised, one-sided confidence gain on the gold value. It is intended for controlled evaluation audits rather than as a comprehensive measure of attribute recoverability. Subject-level evaluation reports the share of personal attributes an adversary fails to infer, pooled over people (CPR) or averaged per person (IPR) \citep{oh-etal-2026-subject}; attack accuracy counts top-1 correctness; proper scoring rules score a whole predictive distribution \citep{brier-1950-verification,gneiting-raftery-2007-strictly}; AUC measures discrimination, and $|\mathrm{AUC}-.5|$ is unchanged when the labels are exchanged. Crediting only the increase above a document-free prior charges a release for what it adds rather than for what the adversary already guessed; the risk is the other half, where a drop below the prior costs nothing. Our stress test targets that one property, which attack accuracy and proper scoring rules on the gold value share when their prediction errors are read as privacy (Section~\ref{sec:swap}). $L$ is distinct from CPR and IPR, whose validity we do not assess.

\paragraph{Evaluation choices.}
A ranking depends on what its score credits: membership-inference attacks rank differently under average-case and low-false-positive-rate metrics \citep{carlini-etal-2022-membership}, and anonymisation has been judged by information-theoretic criteria, by re-identification attacks and by quantitative information flow \citep{sanchez-batet-2016-sanitized,manzanares-salor-etal-2024-evaluating,smith-2009-foundations,alvim-etal-2012-measuring}. Multiverse and specification-curve analyses vary defensible analytic choices \citep{steegen-etal-2016-multiverse,simonsohn-etal-2020-specification}; because our choices change what is estimated and on which entries, we analyse each separately. Like \citet{mozes-kleinberg-2021-intruder}, we ask whether an evaluation measures what it claims to, adding judge controls, bootstrap intervals, multiple-comparison corrections \citep{krippendorff-2004-content,efron-tibshirani-1993-bootstrap,holm-1979-simple} and answer-position checks \citep{zheng-etal-2024-selectors,pezeshkpour-hruschka-2024-order}; none of this work varies the interface while the released text is held fixed.

\paragraph{Attribute inference.}
Language models infer personal attributes from ordinary text \citep{staab-etal-2024-beyond}, extending classifier-based attacks \citep{mosallanezhad-etal-2019-deep,elazar-goldberg-2018-adversarial}. \citet{oh-etal-2026-subject} evaluate anonymisation per person; we follow that framing and reuse their annotated TAB documents with our own scorer (Section~\ref{sec:eval}).

\paragraph{Anonymisers.}
Span-based de-identification masks or generalises identifiers \citep{lison-etal-2021-anonymisation,pilan-etal-2022-text}. Language-model anonymisers rewrite under adversarial feedback \citep{staab-etal-2025-anonymizers}, replace spans truthfully \citep{pilan-etal-2025-truthful}, optimise prompts \citep{loiseau-etal-2026-adaptive}, or randomise the protected attribute \citep{frikha-etal-2025-incognitext}. Randomisation matters here because, whenever the random value is wrong, a score that floors losses at zero credits it as it credits SWAP, though a random value cannot be undone by a known rule. Appendix~\ref{app:related} surveys further related work.

% =============================================================================

\section{Releases and Scorer}
\label{sec:eval}
\label{sec:task}
\label{sec:setup}

Figure~\ref{fig:measure} introduces the corpora, editors and readouts. The main protocol clips confidence gains at zero, uses each editor's subject reference, balances candidate orders, includes all annotated subjects and uses the development adversary. Section~\ref{sec:multiverse} varies these choices and checks content measures and abstention (Table~\ref{tab:choices}); the prior and band $\delta$ are examined separately in Appendix~\ref{app:x2}.

\begin{table*}[t]
\centering
\caption{\textbf{Seven choices in the evaluation protocol}, each an alternative reading of fixed releases (effects in Section~\ref{sec:multiverse}).}
\label{tab:choices}
\begingroup
\small
\setlength{\tabcolsep}{4pt}
\renewcommand{\arraystretch}{1.10}
\begin{tabular*}{\textwidth}{@{\extracolsep{\fill}}
>{\raggedright\arraybackslash}p{0.18\textwidth}
>{\raggedright\arraybackslash}p{0.34\textwidth}
>{\raggedright\arraybackslash}p{0.445\textwidth}@{}}
\toprule
\textbf{Protocol choice} & \textbf{Reading examined} & \textbf{Alternative or check} \\
\midrule
\textbf{Wrong-belief credit}
&
Clipping floors negative changes in
gold-value confidence at zero leakage.
&
Signed residual; three outcome states
with \leakE{} and~\discE{}.
\\
\addlinespace[0.16em]
\textbf{Subject reference}
&
References depend on each editor's
output.
&
Match references across editors
or derive one from the~input.
\\
\addlinespace[0.16em]
\textbf{Candidate order}
&
Seeded orders retain a positional
bias in $\prior$.
&
Balance gold-first and gold-second
orders for \attr{Sex}; report AUC alongside $L$.
\\
\addlinespace[0.16em]
\textbf{Which subject is read}
&
Aggregate all annotated subjects.
&
Split by subject position and
exclude shared references.
\\
\addlinespace[0.16em]
\textbf{Content proxies}
&
Accept \unli{}, TPS and LLM
judgments without validity checks.
&
Check placeholder effects,
a second encoder and judgments on the anonymised input.
\\
\addlinespace[0.16em]
\textbf{Adversary family}
&
Evaluate with one development
adversary.
&
Compare three model families
and consider the worst~case.
\\
\addlinespace[0.16em]
\textbf{Abstention}
&
Declined answers are dropped
or assigned the prior without an explicit rule.
&
State the convention and report
both readings.
\\
\bottomrule
\end{tabular*}
\endgroup
\end{table*}

\paragraph{Corpora and protected entries.}
\looseness=-1\relax We use 1{,}249 MedQA clinical vignettes \citep{jin-etal-2021-disease}, 144 TAB court judgments with per-person annotations \citep{oh-etal-2026-subject,pilan-etal-2022-text} and 281 SynthPAI posts \citep{yukhymenko-etal-2024-synthetic}. Each protected \emph{entry} $(s,a)$ pairs a person $s$ (a \emph{subject}) with an attribute $a$ and a gold value $g$: \attr{Sex} and \attr{Occupation} on MedQA and TAB, and \attr{Relationship} on SynthPAI. TAB and SynthPAI gold values come from human annotation, MedQA's from fixed GPT-4o-mini prompts; a Longformer tagger de-identifies TAB and SynthPAI \citep{beltagy-etal-2020-longformer} and a rule-based tagger MedQA, whose \texttt{[OCCUPATION]} placeholders mostly replace the clinician or a word naming no occupation. They make \attr{Occupation} hard to interpret; we keep it in the overall score, the \emph{endpoint} fixed before any editor was scored, but report \attr{Sex}-only contrasts beside every main MedQA comparison. We call the de-identified text the \emph{anonymised input}.

\paragraph{Residual-cue editing.}
The editors we compare perform one task. Given an anonymised document $D$, protected subjects $S$ with the mention forms $M(s)$ by which $D$ refers to each, and attribute names $A$, find the \emph{residual cues}---spans whose surface form states a value of some $a\in A$ for some person---and return $D'$, with edits confined to those spans and the grammatical agreement they disturb. Three requirements define a good edit: \emph{privacy}, an adversary reading $D'$ about $s$ should end no better off than with no text, neither more confident in the true value nor steered towards a wrong one; \emph{utility}, content stating no protected attribute of a protected subject should survive; \emph{scope}, cues bound to people outside $S$, and attribute words naming no person, should be left alone. These are desiderata, not guarantees: the evaluation tests whether they hold under a stated adversary and scorer (Appendix~\ref{app:taskeditor}).

\paragraph{Editors.}
Each editor maps the anonymised input to a released text. \emph{Global mask} redacts every match in a shared lexicon of attribute words, whoever the word describes; the \emph{global ladder} replaces each match with a broader term. \sacc{} follows two principles. \emph{Default-deny closure}: fixed word lists propose cues and a rule-based \emph{binder} uses a dependency parse to decide whom each describes---a protected person (directly or through a relation word, \emph{husband} in \emph{her husband}), a person it cannot identify (unresolved), someone else, or no person---and a cue is edited unless the binder assigns it to someone else (the \emph{other-person rule}) or to no person (the \emph{non-person rule}). Unresolved cues matter because de-identification often removes the name a binder would need. \emph{Admissible rewriting}: a selected cue becomes a value-neutral counterpart from a fixed table (\emph{he}$\to$\emph{they}), or is redacted when the table has none, and a repair step fixes verb agreement (the only tuned setting). \emph{SACC-mask} keeps the selection and redacts throughout. We also score an \emph{LLM rewrite} given \sacc{}'s inputs and two anonymisers that start from the source text, \emph{INTACT} \citep{pilan-etal-2025-truthful} and \emph{Adaptive} \citep{loiseau-etal-2026-adaptive}; Section~\ref{sec:method} also runs \sacc{} unchanged on two unseen corpora, where scope matters (Appendix~\ref{app:baselines}).

\paragraph{The adversary.}
The adversary, a language model $M$, reads a released text $y$ and a reference $h(s,y)$ to the person and answers a multiple-choice question over candidate values $C\ni g$ in one token. We ask it under $K$ seeded candidate orders $\sigma_k$ shared by all editors and average the gold value's probability, renormalised over $C$:
\begin{equation*}
  \post=\frac{1}{K}\sum_{k=1}^{K}\Pr_M\bigl[g \mid y,\,h(s,y),\,\sigma_k(C)\bigr].
\end{equation*}
The prior $\prior$ is the same quantity with no text; entries whose prior already exceeds a corpus-specific threshold are excluded and the rest are \emph{eligible}. For binary \attr{Sex} the prior largely reflects which option comes first, so the main scores weight gold-first and gold-second orders equally. The reference $h(s,y)$ is the longest surviving mention carrying no cue. The \emph{development adversary}, GPT-4o-mini, was used while designing \sacc{} and also annotated MedQA; Llama \citep{grattafiori-etal-2024-llama} and Qwen \citep{yang-etal-2025-qwen3} are held out. A reasoning adversary (GPT-4.1-mini) declines some answers; we run it on a sample and report both conventions.

\paragraph{The score and what it credits.}
Per entry, we compute the prior-normalised residual
\begin{equation*}
  d=\operatorname{clip}\Bigl(\frac{\post-\prior}{1-\prior},\,0,\,1\Bigr),
\end{equation*}
the share of the distance from the prior to certainty that the text provides. The \emph{clipped leakage score} $L$ averages $d$ within subgroups (strata) defined by the annotations (e.g., occupation categories), then within each attribute, and then across attributes, so that no subgroup dominates (Appendix~\ref{app:protocol}). Lower $L$ means a smaller \emph{clipped} confidence gain on the gold value, with $0$ at the prior.

\paragraph{Two-sided reading.}
Clipping sets $d=0$ both when the text leaves the gold probability at the prior and when it pushes that probability below the prior (Figure~\ref{fig:teaser}). Dropping the clip yields a \emph{signed} residual, in which movement below the prior is credited further; Table~\ref{tab:choices} calls this choice wrong-belief credit. To make downward shifts explicit we assign each entry to one of three states with a band $\delta=.05$: \leakB{} ($\post>\prior+\delta$), \anonB{} ($|\post-\prior|\leq\delta$) or \misB{} ($\post<\prior-\delta$). The \leakE{} rate counts \leakB{} entries and the \discE{} rate counts \leakB{} and \misB{} together; both are balanced over strata like $L$. \discE{} records two-sided belief displacement; Section~\ref{sec:swap} tests recoverability.

\paragraph{Checks on the question.}
Three properties of the question can distort these scores. Without a document, binary answers often follow option position, so removing a cue can move the gold probability off the prior without the text stating any value; we therefore also report \attr{Sex} AUC, the probability that the adversary rates a male person as more likely male than a female one, which its default answer cannot shift. An editor that paraphrases can delete the mentions a reference relies on, so when comparing paraphrasing with in-place editors we use entries where both yield the same reference. In documents describing several people the adversary may answer from another person's cues, so we also split off each document's \emph{first-named} subject.

\paragraph{Content and statistics.}
\unli{} measures content away from edits, the share of source sentences without protected cues that an NLI model finds entailed by the release \citep{he-etal-2023-debertav3}; at edit sites TPS measures how much information-weighted content remains \citep{pilan-etal-2025-truthful}, read beside Edit\%, the share of characters changed. Document-bootstrap intervals are 99\% for privacy and 95\% for content. A paired contrast is \emph{separable} if its interval excludes zero; its \emph{verdict} is its sign if separable and ``not separable'' otherwise, so a verdict can change by a reversal or by a gain or loss of separability. Table~\ref{tab:x1} uses seeded orders and Table~\ref{tab:main} \attr{Sex} orders balanced by position; they select different entries, so no absolute value carries across. Main runs were pre-specified in dated files (Appendix~\ref{app:protocol}); post hoc analyses are labelled, failed predictions~included.

% =============================================================================

\section{Misdirection Scores as Privacy}
\label{sec:swap}

SWAP is a deliberately invertible stress test of what the clipped score rewards, not a proposed anonymiser: it isolates one property of the score.

\paragraph{Construction.}
SWAP starts from the SACC-mask release: wherever SACC-mask redacted a \attr{Sex} cue, SWAP writes the opposite-sex form instead (\emph{he} for \emph{she}, \emph{wife} for \emph{husband}), and every other selected cue stays redacted. Replacing the flipped words with redactions recovers SACC-mask on 99.7\% of MedQA and all TAB documents, so the releases differ only in sex cues. Table~\ref{tab:x1} reads every system under seeded orders, as pre-specified, the only reading we report for SWAP; SWAP and SACC-mask share entries, orders and prior, so their paired contrast is directly comparable. Three of four pre-specified predictions held; the fourth---that $L$ would also favour SWAP on TAB---did not.

\begin{table}[t]
\centering
\small
\setlength{\tabcolsep}{2.8pt}
\renewcommand{\arraystretch}{1.08}

\begin{adjustbox}{max width=\linewidth}
\begin{tabular}{@{}l ccccc ccc@{}}
\toprule
& \multicolumn{5}{c}{\textbf{MedQA} (2,093)}
& \multicolumn{3}{c}{\textbf{TAB} (505)} \\
\cmidrule(lr){2-6}
\cmidrule(lr){7-9}

\textbf{System}
& $L$
& \textsc{Leak}
& \textsc{Disc}
& AUC
& Rk
& $L$
& \textsc{Leak}
& \textsc{Disc} \\
\midrule

SWAP (flips sex cues)
& .536 & .564 & .979 & .180 & 6
& .444 & .534 & .859 \\

Adaptive$^\dagger$
& .581 & .704 & .897 & .734 & 3
& .618 & .677 & .892 \\

Global ladder
& .628 & .723 & .898 & .719 & 1
& .408 & .603 & .800 \\

\sacc{}
& .636 & .728 & .901 & .719 & 1
& .423 & .599 & .792 \\

Global mask
& .653 & .741 & .931 & .737 & 4
& .444 & .634 & .848 \\

SACC-mask
& .659 & .749 & .937 & .739 & 5
& .446 & .620 & .817 \\

Anonymised input
& .842 & .871 & .990 & .888 & 7
& .556 & .642 & .867 \\

INTACT
& .853 & .879 & .991 & .955 & 8
& .576 & .661 & .885 \\

\midrule

SWAP $-$ SACC-mask, \attr{Sex}
& \textbf{$-$.253}
& \textbf{$-$.376}
& \textbf{+.083}
& --- &
& +.002
& \textbf{$-$.189}
& \textbf{+.067} \\

SWAP $-$ SACC-mask, all
& \textbf{$-$.122}
& \textbf{$-$.186}
& \textbf{+.043}
& --- &
& $-$.001
& \textbf{$-$.086}
& \textbf{+.042} \\

\bottomrule
\end{tabular}
\end{adjustbox}

\caption{\textbf{Effect of deliberate misdirection.} Development adversary, seeded orders; bottom rows: paired differences (bold: 99\% CI excludes 0). SWAP misleads 74\%/68\% of MedQA/TAB \attr{Sex} entries, SACC-mask 26\%/18\% (raw shares, not balanced). AUC: MedQA \attr{Sex} discrimination (below .5, labels exchanged); Rk: its rank by $|\mathrm{AUC}-.5|$ among the eight, 1 most private; rows follow $L$. No TAB AUC: pooled, it mixes subjects the adversary reads through another person (Section~\ref{sec:multiverse}); on first-named subjects SWAP ranks fifth by $|\mathrm{AUC}-.5|$ (Table~\ref{tab:metricfamily}). $^\dagger$: scored with its own subject reference.}
\label{tab:x1}
\end{table}

\paragraph{The clipped leakage score rewards misdirection.}
On MedQA, SWAP attains lower $L$ (.536) than every editor in Table~\ref{tab:x1}, while 74\% of its \attr{Sex} entries are \misB{} (vs. 26\% for SACC-mask). On \attr{Sex} entries its $L$ is .253 below SACC-mask’s and its \discE{} rate .083 higher, both separable. On TAB \attr{Sex} entries, $L$ does not separate SWAP from SACC-mask, while \leakE{} falls by .189 and \discE{} rises by .067. The credit is not the clip's alone: read as privacy, attack accuracy, Brier score and log loss, valid measures of prediction that also score only the gold value, rate SWAP the most private release on both corpora (on TAB separably from SACC-mask for the last two), whereas \discE{} and $|\mathrm{AUC}-.5|$, blind to the shift's direction, do not (Appendix~\ref{app:x1}, Table~\ref{tab:metricfamily}).

\paragraph{The information remains recoverable.}
\attr{Sex} AUC signals this without the rule. Ranked by $|\mathrm{AUC}-.5|$ (Rk in Table~\ref{tab:x1}), the systems scored on the shared entries keep their order from $L$ while SWAP falls from first to sixth of eight, or fifth if Adaptive, scored with its own reference, is set aside: its AUC of .18 is an inverted separation wider than the global editors, \sacc{} and SACC-mask leave. That reading diagnoses inverted discrimination, not recovery. SWAP’s fixed rule enables a direct test. We described the flip to the development adversary and re-scored the 1{,}513 MedQA \attr{Sex} entries (post hoc; Appendix~\ref{app:x1}). The mean probability of the gold value rises from .252 to .769, 80\% of entries favour it, and $L$ rises by .414 ($[+.374,+.454]$). \discE{} barely changes, because knowing the rule moves misled entries into the leaking state and \discE{} counts both. The clip’s zero is therefore an artefact of the score, not evidence that SWAP erased the attribute: the rule-aware adversary recovers it from the same text.

% =============================================================================

\section{Which Conclusions Survive the Interface}
\label{sec:multiverse}

\begin{figure}[t]
\centering
\begin{adjustbox}{max width=\columnwidth}
\begin{tikzpicture}[x=1cm,y=1cm,
  every node/.style={font=\scriptsize}]

  \def\barleft{2.45}
  \def\scalefactor{9.1}

  % Header
  \node[anchor=west,font=\bfseries\scriptsize]
    at (0,4.02) {Protocol choice};
  \node[anchor=center,font=\bfseries\scriptsize]
    at (3.58,4.02) {Changed (\%)};
  \node[anchor=center,font=\bfseries\scriptsize]
    at (5.20,4.02) {Chg./$N$};
  \node[anchor=west,font=\bfseries\scriptsize]
    at (5.65,4.02) {Median $|\Delta L|$};
  \node[anchor=west,font=\bfseries\scriptsize]
    at (7.25,4.02) {$\tau_{\mathrm{b}}$};

  \draw[black!65,line width=.35pt]
    (0,3.78) -- (8.85,3.78);

  % Percentage guides
  \foreach \x/\lab in {
    2.45/0, 3.36/10, 4.27/20, 4.72/25
  } {
    \draw[black!12,line width=.3pt]
      (\x,0.35) -- (\x,3.72);
    \node[anchor=north,text=black!65]
      at (\x,0.24) {\lab};
  }

  % #1 y-position
  % #2 choice
  % #3 reversals
  % #4 gained separations
  % #5 lost separations
  % #6 total contrasts
  % #7 changed contrasts
  % #8 median |shift| of the paired estimate (was a p-value before 2026-10-13)
  % #9 Kendall tau_b
  \newcommand{\effectrow}[9]{%
    \node[anchor=west,align=left,
          text width=2.30cm] at (0,#1) {#2};%
    \pgfmathsetmacro{\wr}{
      \scalefactor*(#3)/(#6)}%
    \pgfmathsetmacro{\wg}{
      \scalefactor*(#4)/(#6)}%
    \pgfmathsetmacro{\wl}{
      \scalefactor*(#5)/(#6)}%
    % Small cohorts are displayed as counts, not visually exaggerated shares.
    \ifnum#6<10\relax
      \node[anchor=west,text=black!55,font=\scriptsize]
        at (\barleft,#1) {$N=#6$};%
    \else
      \fill[black!90]
        (\barleft,{#1-.095}) rectangle ++(\wr,.19);%
      \fill[black!60]
        ({\barleft+\wr},{#1-.095}) rectangle ++(\wg,.19);%
      \fill[black!27]
        ({\barleft+\wr+\wg},{#1-.095}) rectangle ++(\wl,.19);%
    \fi
    \node[anchor=center] at (5.20,#1)
      {#7/#6};%
    \node[anchor=west] at (5.70,#1)
      {$#8$};%
    \node[anchor=west] at (7.25,#1)
      {#9};%
  }

  \effectrow{3.42}
    {\textbf{Wrong-belief credit}}
    {0}{1}{10}{84}{11}{.0084}
    {.71 / 1.00 / .95}

  \effectrow{2.84}
    {Subject reference$^*$}
    {0}{0}{3}{45}{3}{.0003}
    {---}

  \effectrow{2.26}
    {Candidate order}
    {0}{7}{6}{252}{13}{\text{---}}
    {1.00 / 1.00 / .95}

  \effectrow{1.68}
    {Which subject is read}
    {0}{3}{1}{28}{4}{\text{---}}
    {---}

  \effectrow{1.10}
    {\textbf{Adversary family}}
    {5}{34}{85}{504}{124}{.0236}
    {\text{---} / .42 / .69}

  \effectrow{0.52}
    {Abstention}
    {0}{4}{2}{75}{6}{.0059}
    {---}

  % Legend
  \fill[black!90]
    (0,-.34) rectangle ++(.16,.16);
  \node[anchor=west] at (.23,-.26)
    {Reversal};

  \fill[black!60]
    (1.80,-.34) rectangle ++(.16,.16);
  \node[anchor=west] at (2.03,-.26)
    {Gained separation};

  \fill[black!27]
    (4.35,-.34) rectangle ++(.16,.16);
  \node[anchor=west] at (4.58,-.26)
    {Lost separation};

  \node[anchor=west,text=black!65]
    at (0,-.72)
    {Content proxies: no paired verdict comparison.};

\end{tikzpicture}
\end{adjustbox}

\caption{\textbf{Verdict sensitivity to protocol choices} over all pairs of released texts (988 contrasts; 161 involve \sacc{}). Bars: reversals and separations gained or lost; labels: changed/compared counts, median $|\Delta L|$ and Kendall's $\tau_{\mathrm{b}}$ of the seven-text ranking (MedQA / TAB / SynthPAI). Values are descriptive; all exceed the bootstrap-reseeding rate (.007 / .012 / .024). No median for candidate order (verdicts only) or the subject choice (read on AUC); $\tau_{\mathrm{b}}$ needs seven shared texts. $^*$Both readings before the binder fix of Appendix~\ref{app:history}.}
\label{fig:multiverse}
\end{figure}
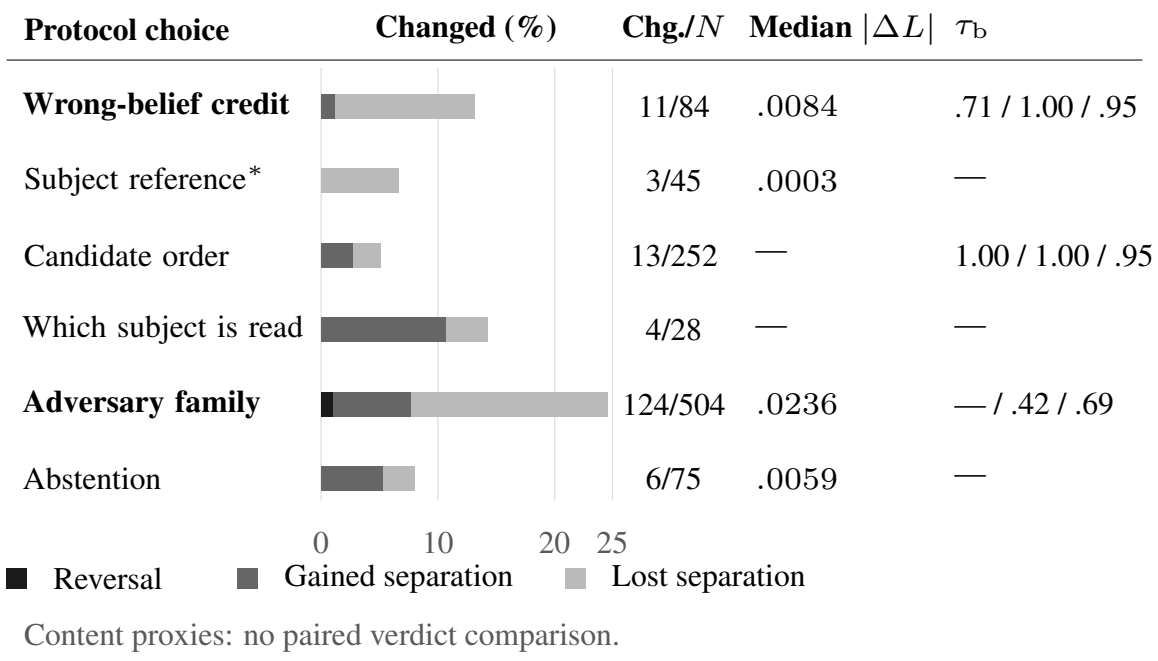

We now vary the evaluation interface for ordinary fixed releases, one choice at a time, over 988 contrasts between released texts (Table~\ref{tab:main}; Figure~\ref{fig:multiverse}).

\begin{table*}[t]
\centering

\begingroup
\footnotesize
\setlength{\tabcolsep}{2pt}
\renewcommand{\arraystretch}{1.04}

\resizebox{\textwidth}{!}{%
\begin{tabular}{
@{}l
*{7}{r}
@{\hspace{6pt}}
*{7}{r}
@{}
}
\toprule
& \multicolumn{7}{c}{\textbf{MedQA}}
& \multicolumn{7}{c}{\textbf{TAB}} \\
\cmidrule(lr){2-8}
\cmidrule(lr){9-15}

\textbf{System}
& $L\downarrow$
& AUC
& \textsc{Disc}$\downarrow$
& \textsc{Leak}$\downarrow$
& $\unli\uparrow$
& TPS
& Edit\%
& $L\downarrow$
& AUC
& \textsc{Disc}$\downarrow$
& \textsc{Leak}$\downarrow$
& $\unli\uparrow$
& TPS
& Edit\% \\
\midrule

Anonymised input
& .842 & .891 & .991 & .870
& .720 & 1.000 & 0.0
& .558 & .979 & .872 & .644
& .313 & 1.000 & 0.0 \\

\midrule

Global mask
& .653 & .741 & .935 & .745
& .661 & .795 & 2.4
& .433 & .668 & .851 & .604
& .291 & .803 & 2.9 \\

Global ladder
& \textbf{.627} & .734
& \textbf{.895} & \textbf{.722}
& .702 & .842 & 1.4
& \textbf{.397} & .592
& .805 & \textbf{.578}
& .296 & .801 & 1.9 \\

\midrule

INTACT \citep{pilan-etal-2025-truthful}
& .852 & .955$^\ddagger$ & .989 & .877
& \textbf{.737} & .806 & 2.7
& .578 & .969$^\ddagger$ & .895 & .664
& .293 & .770 & 5.0 \\

Adaptive \citep{loiseau-etal-2026-adaptive}$^\dagger$
& .596 & .724$^\ddagger$ & .914 & .702
& .467 & .772 & 51.7
& .600 & 1.00$^{\ddagger\S}$ & .890 & .661
& .287 & .781 & 5.5 \\

\midrule

SACC-mask
& .659 & .743 & .940 & .750
& .665 & .794 & 2.2
& .438 & .666 & .818 & .599
& \textbf{.319} & .795 & 1.2 \\

\rowcolor{oursrow}
\textbf{SACC}
& .635 & .734
& .900 & .728
& .706 & .846 & 1.3
& .413 & .616
& \textbf{.795} & \textbf{.578}
& \textbf{.319} & .845 & 0.8 \\

\midrule

\textcolor{black!50}{LLM rewrite$^\dagger$}
& \textcolor{black!50}{.609}
& \textcolor{black!50}{---}
& \textcolor{black!50}{.918}
& \textcolor{black!50}{.727}
& \textcolor{black!50}{.661}
& \textcolor{black!50}{.802}
& \textcolor{black!50}{2.8}
& \textcolor{black!50}{.423}
& \textcolor{black!50}{---}
& \textcolor{black!50}{.787}
& \textcolor{black!50}{.588}
& \textcolor{black!50}{.297}
& \textcolor{black!50}{.793}
& \textcolor{black!50}{2.1} \\

\bottomrule
\end{tabular}%
}
\endgroup

\vspace{2pt}
{\scriptsize
\begin{tabular*}{\textwidth}{@{\extracolsep{\fill}}l*{7}{c}@{}}
\multicolumn{8}{@{}l}{\textbf{MedQA \attr{Sex}-only contrasts, $\Delta L$ = system $-$ SACC (paired; 99\% intervals in Table~\ref{tab:medqasex})}} \\
\cmidrule(lr){1-8}
& Input & Gl. mask & Gl. ladder & INTACT & Adaptive$^\dagger$ & SACC-mask & LLM rewrite$^\dagger$ \\
$\Delta L_{\mathrm{Sex}}$
& \textbf{+.410} & \textbf{+.056} & $-$.006 & \textbf{+.438} & \textbf{+.051} & \textbf{+.061} & \textbf{+.039} \\
\end{tabular*}
}

\caption{\looseness=-1\relax \textbf{Privacy and utility under the development adversary}, balanced \attr{Sex} orders (2,149 MedQA and 501 TAB shared entries). $^\dagger$: other entries, where SACC scores $L$ .597/.417 (Adaptive) and .635/.431 (LLM rewrite). TAB AUC: 98 first-named entries (99\% CI $\pm.2$). $^\ddagger$: AUC from stored per-order probabilities on the full \attr{Sex} cohort (1,506 and 98), not the other rows' re-read cohort (agreeing within .005). $^\S$: perfect separation on 14 female subjects (a bound). Bold: best per arrowed column among shared-entry systems. MedQA $L$ includes \attr{Occupation}; the inset: paired \attr{Sex}-only contrasts (bold there: CI excludes 0). Cohorts, seeded orders, SynthPAI: Tables~\ref{tab:cohortprovenance}, \ref{tab:mainseeded}, Appendix~\ref{app:synthpai}.}
\label{tab:main}

\end{table*}

\paragraph{How often verdicts change.}
\looseness=-1\relax Across all pairs of releases (Figure~\ref{fig:multiverse}), replacing $L$ with the signed residual changes 11 of 84 verdicts (18\% of MedQA and 21\% of TAB pairs, none of SynthPAI's), almost always by removing separability rather than reversing a sign; candidate order changes about 5\%. The adversary family changes the most, 124 of the 504 contrasts we analyse, but only five of those are sign reversals; the other 119 gain or lose separability, statistical uncertainty rather than a reversed ranking. For the reasoning adversary, the convention for declined answers decides whether the global ladder is separably more private than \sacc{} on MedQA. As controls, re-reading cached answers changes no verdict and new bootstrap seeds change 1--4\% (Appendix~\ref{app:pairs}).

\paragraph{The adversary can reverse a comparison.}
Release rankings are similar under most choices, but the adversary family reorders the seven releases that all three families score on TAB ($\tau_{\mathrm{b}}=.42$) and SynthPAI (.69). The two reversals involving \sacc{} both stem from a single contrast on TAB (SACC-mask vs. \sacc{}): the estimated difference is $+.022$, $-.023$, and $+.042$ under the three families, and each is separable. Consequently, whether rewriting is more private than redaction on TAB depends on the adversary, although the effects are small. A fourth adversary (GPT-4.1-mini without reasoning, post hoc) also favours redaction, so two of four do (Appendix~\ref{app:transfer}). Llama’s reversal is primarily attributable to \attr{Occupation}; on \attr{Sex} alone the difference is not separable in either direction.

\paragraph{The adversary may read the wrong person.}
In TAB’s anonymised input, \attr{Sex} AUC is .98 for each judgment’s first-named person but .32 for others, whose predicted sex follows the first-named person’s (post hoc; Table~\ref{tab:subjectpos}). An AUC of .32 is consistent with systematic misattribution: pooling both groups yields .53, which reflects neither, and an editor that moves .32 toward .5 would appear to \emph{increase} discrimination. In 73 of 244 TAB \attr{Sex} entries the reference also matches another person (Appendix~\ref{app:tabref}); excluding them increases the input--\sacc{} reduction in \attr{Sex} $L$, and all comparisons keep their sign. Matched references also matter: when all editors share one reference, Adaptive’s MedQA advantage shrinks from .054 to a non-separable .024 (on texts before a fix to the binder, and still not separable after it; Appendix~\ref{app:robust}). On identical texts, a shared reference changes only verdicts involving Adaptive (3 of its 24; Figure~\ref{fig:multiverse}).

\paragraph{Which attribute is aggregated.}
The verdict also changes with the attributes the score averages. MedQA's \attr{Occupation} placeholders are hard to interpret (Section~\ref{sec:eval}), and on \attr{Sex} alone two contrasts reverse sign: the LLM rewrite, separably more private than \sacc{} on the pre-specified endpoint, leaks more ($+.039$), as does Adaptive ($+.051$; Table~\ref{tab:main}). We keep the pre-specified endpoint and report both readings rather than promoting the narrower one after seeing the result (Table~\ref{tab:medqasex}); which attributes are averaged is itself an interface choice.

\paragraph{Other choices, in brief.}
Content proxies and the band need controls of their own. Two of three LLM content judges fail no-change controls on the anonymised input, and under another encoder \sacc{}’s TPS advantage over the ladder disappears (Appendices~\ref{app:protocol} and~\ref{app:judges}). On TAB, placeholders hold \unli{} near .31 for every system, so we read content at edit sites. Widening the band $\delta$ to .10 changes 5 to 11 of 56 displacement verdicts per corpus (Table~\ref{tab:multiverse2}), more than candidate order does, and a prior computed on a redaction output favours redaction by construction, while a cue-deletion prior changes one of nine verdicts' separability and preserves \sacc{}’s advantage on $L$ (Appendix~\ref{app:x2}), so the prior is itself a protocol choice, not a property of the text.

% =============================================================================

\section{Evaluating SACC}
\label{sec:method}
\label{sec:results}

\sacc{} is the traceable controlled editor that made the releases above possible; we now ask what its two principles contribute, and under which adversaries and corpora its advantage holds.

\begin{table}[b]
\vspace*{4pt}
\centering
\small
\setlength{\tabcolsep}{3.4pt}
\begin{adjustbox}{max width=\linewidth}
\begin{tabular}{@{}ll cc cc@{}}
\toprule
& & \multicolumn{2}{c}{MedQA} & \multicolumn{2}{c}{TAB} \\
\cmidrule(lr){3-4}\cmidrule(lr){5-6}
Change & at & $\Delta L$ & $\Delta\unli$ & $\Delta L$ & $\Delta\unli$ \\
\midrule
Closure over binding only & rewrite & \textbf{+.079} & +.002 & \textbf{+.122} & +.002 \\
Admissible rewriting & closure & \textbf{+.024} & \textbf{+.041} & \textbf{+.022} & \phantom{+}.000 \\
Closure selection & rewrite & \textbf{$-$.010} & \phantom{+}.000 & +.001 & \textbf{+.026} \\
\quad non-person rule & mask & $-$.006 & \textbf{+.004} & $-$.011 & \textbf{+.026} \\
\quad other-person rule & mask & \phantom{+}.000 & $-$.001 & +.009 & +.002 \\
\bottomrule
\end{tabular}
\end{adjustbox}
\caption{\textbf{What each principle contributes}, under the development adversary. Positive $\Delta L$: the change lowers leakage (99\% intervals); positive $\Delta\unli$: it keeps more content (95\%); bold excludes 0. Indented: the selection's two rules; rows do not add (Table~\ref{tab:factorial}).}
\label{tab:ablmain}
\end{table}

\paragraph{What each principle contributes.}
Ablations vary one principle at a time (Table~\ref{tab:ablmain}). The privacy gain sits in closure: editing only the cues the binder assigns to a protected person leaks more by .079 on MedQA and .122 on TAB. It matters because 41\% of MedQA and 90\% of TAB edited cues are unresolved, and a judge reading the source text assigns 97.5\% of unresolved TAB sites to protected people. Admissible rewriting lowers $L$ over redaction by .024 and .022, keeps more MedQA content ($\unli{}+.041$) and leaves fewer MedQA \attr{Sex} entries \misB{} ($-.031$; Appendix~\ref{app:pertag}), but that privacy gain is reproduced by Qwen and not by Llama (Appendix~\ref{app:transfer}), so rewriting is mainly a content choice that can be dropped at little worst-case privacy cost. Scoping the selection is not itself a privacy gain: against global selection with the same rewrite step, \sacc{}'s selection is .010 \emph{less} private on MedQA and not separable on TAB, where it keeps more content. Of its two rules, the non-person rule buys content rather than privacy and the other-person rule moves neither measurably, so we do not claim that subject awareness lowers leakage.

\paragraph{What the subject list buys.}
The other-person rule is the only one that reads the subject list, and since nearly every annotated person here is protected, it changes no measure. Judged against the source text on TAB it is right on .32 of the sites it keeps, leaving 2.1 protected persons' sites unedited for each one it correctly withholds (Appendix~\ref{app:unseen}). Scope is thus a release policy a deployment can switch off, not a privacy mechanism. Applied unchanged to 500 biographies \citep{de-arteaga-etal-2019-bias} and 500 plot summaries \citep{bamman-etal-2013-learning}, \sacc{} lowers input leakage by .35 and .11 while editing the fewest characters; but these corpora name many unprotected people, and there the global ladder leaks less, by .030 and .089 (Appendix~\ref{app:unseen}), because it edits their cues too.

\paragraph{Leakage relative to the input.}
\sacc{} reduces $L$, \leakE{} and \discE{} relative to its anonymised input on both corpora (Table~\ref{tab:main}); under the seeded reading TAB's \leakE{} reduction is the one verdict that is not separable (Appendix~\ref{app:tabref}). $L$ falls by .207 on MedQA and .145 on TAB, about a quarter of the input score in each corpus (intervals in Table~\ref{tab:contrasts}). Part of the fall is displacement: of the .274 drop in MedQA \attr{Sex} \leakE{} (seeded orders), .174 moves entries into the band and .101 below it, and every editor but INTACT leaves .24--.28 of entries \misB{}, the input .14 (Appendix~\ref{app:pertag}). Most gains come from \attr{Sex}, where AUC also falls (Table~\ref{tab:main}; on TAB post hoc). The MedQA \attr{Sex} reduction remains separable under both held-out adversaries, as does TAB's under Qwen (Appendix~\ref{app:transfer}). On MedQA exam questions, GPT-4o-mini is equally accurate on \sacc{}'s text and on the anonymised input (73.0\%), with no measurable change under any editor (Appendix~\ref{app:utility}).

\paragraph{Comparison with other editors.}
Under the rule-unaware development adversary, global mask leaks more than \sacc{} on MedQA, while the global ladder is narrowly \emph{equivalent} within $\pm.02$ on the three MedQA privacy measures (two one-sided tests; Appendix~\ref{app:probe}). \sacc{} is more private than INTACT on both corpora and than Adaptive on TAB. On MedQA, Adaptive trades content for privacy: Edit\% 51.7 vs. 1.3, and about one-third less content (\unli{} .467 vs. .706). Its lower $L$ ($-$.054) is not separable once both share one reference, because its paraphrase deletes the patient's mention; on \attr{Sex} alone it leaks more ($+.051$; Table~\ref{tab:main}). On TAB \sacc{} edits about one-third fewer sites than the ladder with higher TPS ($+.043$), and blind annotators judge it to keep more content than the ladder and SACC-mask, by .24 and .04 on a 0--2 scale, though not more grammatical (Appendix~\ref{app:human}).

\paragraph{Traceability.}
\sacc{} is deterministic, reads neither the source text nor a gold value, and traces each edit to its rule, which let Section~\ref{sec:swap} build SWAP at the sites SACC-mask redacts. Edits stay inside cue spans and the agreement they disturb, whereas the LLM rewrite changes an upstream placeholder in 10.4\% of MedQA and 16.5\% of TAB documents. \sacc{} calls no language model: 0.17\,s per MedQA document on one CPU core (Appendices~\ref{app:impl} and~\ref{app:baselines}).

% =============================================================================

\section{What to Report}
\label{sec:recs}

Four things belong beside any leakage score. \textbf{Score}: report $L$ with two-sided displacement and, where applicable, AUC, and explain any disagreement (e.g., SWAP, Table~\ref{tab:x1}); where an editor's rule is public, run a rule-aware adversary too. \textbf{Reference and cohort}: give paraphrasing and in-place editors the same reference, and report first-named and other subjects separately. \textbf{Adversary and abstention}: report every adversary used, including any that reverses a result, say whether it knows the editing rule, and report both conventions for declined answers. \textbf{Uncertainty and content}: separate reversals from changes in separability, give each non-separable comparison's minimum detectable difference, and validate content measures.

\section{Conclusion}

\looseness=-1\relax Privacy verdicts depend on the evaluation interface. SWAP ranks first by clipped leakage and attack accuracy on MedQA under a rule-unaware adversary but sixth of eight by $|\mathrm{AUC}-.5|$, and, told the swap rule, that adversary recovers the attribute for most entries. Clipping hides how far confidence falls, and reading failed prediction as privacy rewards misdirection. On ordinary releases most verdict changes concern separability; of the six choices in Figure~\ref{fig:multiverse}, only the adversary reverses comparisons (5 of its 124 changes). Separate analyses find reversals from attribute aggregation and a redaction-confounded prior, not from a confound-free cue-deletion prior. A privacy claim can be interpreted only together with its protocol.

\section*{Limitations}

\paragraph{Scope of the study.}
SWAP exposes a mismatch in pipelines that read a stated adversary's failed prediction as privacy; it does not make the Brier score or log loss invalid measures of prediction, nor show that every subject-level privacy metric shares the weakness. It is a single invertible construction: its rule-aware recovery result pertains to that rule and an adversary told it, and a randomised, secret or non-invertible editor needs its own attacker model and recovery test; the AUC reading, which needs no rule, is interpretable on MedQA and, with wide intervals, on TAB's first-named subjects only. \discE{} measures two-sided belief displacement, not attribute recoverability. For multi-class attributes, \misB{} marks a drop below the prior rather than confidence in an incorrect value. The protocol choices we vary change which entries are scored, so Figure~\ref{fig:multiverse} summarises their effects descriptively, and the ablation rows of Table~\ref{tab:ablmain} hold different things fixed, so they are not one path and do not add. All claims concern a multiple-choice scorer that is instructed which person to consider (Appendix~\ref{app:freeform}) on English corpora, each with a single upstream de-identifier.

\paragraph{Adversaries.}
Main claims rely on one development adversary, a proprietary API whose outputs drift between sessions and which also produced MedQA’s gold labels (Appendix~\ref{app:protocol}); the held-out adversaries are smaller than some used in earlier attack studies \citep{staab-etal-2024-beyond}. Some of \sacc{}’s results do not hold under every adversary: on TAB, its reduction of input leakage and the gain of rewriting over redaction fail under Llama and GPT-4.1-mini without reasoning, and the closure gain is untested beyond the development adversary (Appendices~\ref{app:transfer} and~\ref{app:robust}). When given the editors’ rules, the development adversary ranks the global ladder ahead of \sacc{} on MedQA, contrary to our pre-specified prediction (Appendix~\ref{app:ruleaware}).

\paragraph{Measurement.}
TAB’s 144 documents cannot detect differences below about .04 to .09 (Table~\ref{tab:mde}), so null results reflect limited power. $L$ excludes high-prior entries and is not a population risk measure, and which \attr{Sex} entries are eligible depends on seeded candidate orders, a selection that balancing does not repair. Automatic content measures ignore grammaticality; human judges observed agreement errors after rewriting, and \sacc{} failed its pre-specified grammar test against SACC-mask (Appendix~\ref{app:human}).

\paragraph{The editor.}
\sacc{} edits lexically stated evidence---primarily grammatical sex marking---using lexicons written on the evaluation corpora. Contextual evidence remains, and an LLM rewrite removes more of it: it leaks less than \sacc{} on MedQA’s reported score under the development adversary, and by point estimate but not separably under the held-out families, though more on \attr{Sex} alone (Table~\ref{tab:medqasex}). On these benchmarks, \sacc{}’s gains come from closure and admissible rewriting, while the rule that preserves other people’s cues (the other-person rule) has no measurable privacy effect; on the two new corpora, which name many unprotected people, the global ladder leaks less (Appendices~\ref{app:impl} and~\ref{app:unseen}). In languages with grammatical gender, fewer rewrites would be admissible and more cues would be redacted.

\paragraph{Side channels.}
Some feminine pronouns remain redacted more often than masculine ones, and when the other-person rule leaves a cue, a reader who knows the rule can infer that a person is unprotected; disabling the rule closes this channel (Appendix~\ref{app:impl}). Forms outside the lexicon, such as \emph{Mx.} or neopronouns, pass unedited, so people who use them can become more distinguishable after editing; this cost falls on an already marginalised group \citep{lauscher-etal-2022-pronouns}. A deployment should add the forms used by its population.

\section*{Ethical Considerations}
We use public datasets under their licences: MedQA (MIT), TAB (MIT; published judgments of the European Court of Human Rights), SynthPAI (CC BY-NC-SA 4.0; synthetic), and the two transfer corpora in Appendix~\ref{app:unseen}. Our editor and scoring code are released under the MIT licence; derived texts and scores retain the corpora's applicable licences. A script recomputes the main tables from the per-entry scores and dated run records we release. We collect no new personal data, attempt no re-identification, and query adversaries only about attributes the benchmarks already annotate. The attack follows prior attribute-inference work \citep{staab-etal-2024-beyond}; our contribution is a way to detect when a score overstates the privacy of a release.

SWAP attributes false sex values to the people its texts describe. We built it because a leaderboard based on clipped leakage alone would reward such misdescription, and randomising the protected attribute is already a published anonymisation strategy \citep{frikha-etal-2025-incognitext}. We distribute SWAP only as evaluation code, not as an anonymiser.

\sacc{} does not guarantee privacy: evidence outside its lexicon or across releases may remain inferable, and its rules reveal which forms it cannot rewrite (Limitations), although an adversary given the rules infers less, not more, from every edited MedQA text except SWAP’s (Appendix~\ref{app:ruleaware}). Rewriting gendered expressions is a policy choice, and clinical use was not evaluated. The binary \attr{Sex} labels follow the datasets and do not imply that sex or gender is binary.

The human evaluation (Appendix~\ref{app:human}) involved three unpaid volunteers, recruited through an author (a possible source of recruitment bias); no personal information about them was collected.

\paragraph{Use of AI Assistants.}
AI assistants supported translation and literature surveying. The authors designed and conducted the study, analysed and verified the results, and take full responsibility for the content.

\bibliography{custom,sacc_extra,sacc_multiverse_refs}

\begin{thebibliography}{58}
\providecommand{\natexlab}[1]{#1}

\bibitem[{Alvim et~al.(2012)Alvim, Chatzikokolakis, Palamidessi, and Smith}]{alvim-etal-2012-measuring}
M{\'a}rio~S. Alvim, Konstantinos Chatzikokolakis, Catuscia Palamidessi, and Geoffrey Smith. 2012.
\newblock \href {https://doi.org/10.1109/CSF.2012.26} {Measuring information leakage using generalized gain functions}.
\newblock In \emph{IEEE 25th Computer Security Foundations Symposium (CSF)}, pages 265--279. IEEE.

\bibitem[{Amrhein et~al.(2023)Amrhein, Schottmann, Sennrich, and L{\"a}ubli}]{amrhein-etal-2023-exploiting}
Chantal Amrhein, Florian Schottmann, Rico Sennrich, and Samuel L{\"a}ubli. 2023.
\newblock Exploiting biased models to de-bias text: A gender-fair rewriting model.
\newblock In \emph{Proceedings of the 61st Annual Meeting of the Association for Computational Linguistics (Volume 1: Long Papers)}, Toronto, Canada. Association for Computational Linguistics.

\bibitem[{Bamman et~al.(2013)Bamman, O'Connor, and Smith}]{bamman-etal-2013-learning}
David Bamman, Brendan O'Connor, and Noah~A. Smith. 2013.
\newblock Learning latent personas of film characters.
\newblock In \emph{Proceedings of the 51st Annual Meeting of the Association for Computational Linguistics}, pages 352--361, Sofia, Bulgaria.

\bibitem[{Beltagy et~al.(2020)Beltagy, Peters, and Cohan}]{beltagy-etal-2020-longformer}
Iz~Beltagy, Matthew~E. Peters, and Arman Cohan. 2020.
\newblock \href {https://arxiv.org/abs/2004.05150} {{Longformer}: The long-document transformer}.
\newblock \emph{Preprint}, arXiv:2004.05150.

\bibitem[{Bevendorff et~al.(2019)Bevendorff, Potthast, Hagen, and Stein}]{bevendorff-etal-2019-heuristic}
Janek Bevendorff, Martin Potthast, Matthias Hagen, and Benno Stein. 2019.
\newblock \href {https://doi.org/10.18653/v1/P19-1104} {Heuristic authorship obfuscation}.
\newblock In \emph{Proceedings of the 57th Annual Meeting of the Association for Computational Linguistics}, pages 1098--1108, Florence, Italy. Association for Computational Linguistics.

\bibitem[{Brier(1950)}]{brier-1950-verification}
Glenn~W. Brier. 1950.
\newblock Verification of forecasts expressed in terms of probability.
\newblock \emph{Monthly Weather Review}, 78(1):1--3.

\bibitem[{Carlini et~al.(2022)Carlini, Chien, Nasr, Song, Terzis, and Tram{\`e}r}]{carlini-etal-2022-membership}
Nicholas Carlini, Steve Chien, Milad Nasr, Shuang Song, Andreas Terzis, and Florian Tram{\`e}r. 2022.
\newblock \href {https://doi.org/10.1109/SP46214.2022.9833649} {Membership inference attacks from first principles}.
\newblock In \emph{IEEE Symposium on Security and Privacy (S\&P)}, pages 1897--1914. IEEE.

\bibitem[{De-Arteaga et~al.(2019)De-Arteaga, Romanov, Wallach, Chayes, Borgs, Chouldechova, Geyik, Kenthapadi, and Kalai}]{de-arteaga-etal-2019-bias}
Maria De-Arteaga, Alexey Romanov, Hanna Wallach, Jennifer Chayes, Christian Borgs, Alexandra Chouldechova, Sahin Geyik, Krishnaram Kenthapadi, and Adam~Tauman Kalai. 2019.
\newblock Bias in bios: A case study of semantic representation bias in a high-stakes setting.
\newblock In \emph{Proceedings of the Conference on Fairness, Accountability, and Transparency}, pages 120--128.

\bibitem[{Devlin et~al.(2019)Devlin, Chang, Lee, and Toutanova}]{devlin-etal-2019-bert}
Jacob Devlin, Ming-Wei Chang, Kenton Lee, and Kristina Toutanova. 2019.
\newblock \href {https://doi.org/10.18653/v1/N19-1423} {{BERT}: Pre-training of deep bidirectional transformers for language understanding}.
\newblock In \emph{Proceedings of the 2019 Conference of the North {A}merican Chapter of the Association for Computational Linguistics: Human Language Technologies, Volume 1 (Long and Short Papers)}, pages 4171--4186, Minneapolis, Minnesota. Association for Computational Linguistics.

\bibitem[{Dou et~al.(2024)Dou, Krsek, Naous, Kabra, Das, Ritter, and Xu}]{dou-etal-2024-reducing}
Yao Dou, Isadora Krsek, Tarek Naous, Anubha Kabra, Sauvik Das, Alan Ritter, and Wei Xu. 2024.
\newblock \href {https://doi.org/10.18653/v1/2024.acl-long.741} {Reducing privacy risks in online self-disclosures with language models}.
\newblock In \emph{Proceedings of the 62nd Annual Meeting of the Association for Computational Linguistics (Volume 1: Long Papers)}, pages 13732--13754, Bangkok, Thailand. Association for Computational Linguistics.

\bibitem[{Efron and Tibshirani(1993)}]{efron-tibshirani-1993-bootstrap}
Bradley Efron and Robert~J. Tibshirani. 1993.
\newblock \href {https://doi.org/10.1201/9780429246593} {\emph{An Introduction to the Bootstrap}}.
\newblock Chapman and Hall/CRC, New York.

\bibitem[{Elazar and Goldberg(2018)}]{elazar-goldberg-2018-adversarial}
Yanai Elazar and Yoav Goldberg. 2018.
\newblock \href {https://doi.org/10.18653/v1/D18-1002} {Adversarial removal of demographic attributes from text data}.
\newblock In \emph{Proceedings of the 2018 Conference on Empirical Methods in Natural Language Processing}, pages 11--21, Brussels, Belgium. Association for Computational Linguistics.

\bibitem[{Faustini et~al.(2025)Faustini, Tonni, McIver, Xu, and Dras}]{faustini-etal-2025-idt}
Pedro Faustini, Shakila~Mahjabin Tonni, Annabelle McIver, Qiongkai Xu, and Mark Dras. 2025.
\newblock {IDT}: Dual-task adversarial rewriting for attribute anonymization.
\newblock \emph{Computational Linguistics}, 51(4).

\bibitem[{Feyisetan et~al.(2020)Feyisetan, Balle, Drake, and Diethe}]{feyisetan-etal-2020-privacy}
Oluwaseyi Feyisetan, Borja Balle, Thomas Drake, and Tom Diethe. 2020.
\newblock \href {https://doi.org/10.1145/3336191.3371856} {Privacy- and utility-preserving textual analysis via calibrated multivariate perturbations}.
\newblock In \emph{Proceedings of the 13th International Conference on Web Search and Data Mining (WSDM)}, pages 178--186. ACM.

\bibitem[{Frikha et~al.(2025)Frikha, Walha, Nakka, Mendes, Jiang, and Zhou}]{frikha-etal-2025-incognitext}
Ahmed Frikha, Nassim Walha, Krishna~Kanth Nakka, Ricardo Mendes, Xue Jiang, and Xuebing Zhou. 2025.
\newblock \href {https://doi.org/10.18653/v1/2025.ijcnlp-long.134} {{I}ncogni{T}ext: Privacy-enhancing conditional text anonymization via {LLM}-based private attribute randomization}.
\newblock In \emph{Proceedings of the 14th International Joint Conference on Natural Language Processing and the 4th Conference of the Asia-Pacific Chapter of the Association for Computational Linguistics}, pages 2490--2501, Mumbai, India. The Asian Federation of Natural Language Processing and The Association for Computational Linguistics.

\bibitem[{Gneiting and Raftery(2007)}]{gneiting-raftery-2007-strictly}
Tilmann Gneiting and Adrian~E. Raftery. 2007.
\newblock Strictly proper scoring rules, prediction, and estimation.
\newblock \emph{Journal of the American Statistical Association}, 102(477):359--378.

\bibitem[{Grattafiori et~al.(2024)}]{grattafiori-etal-2024-llama}
Aaron Grattafiori et~al. 2024.
\newblock \href {https://arxiv.org/abs/2407.21783} {The {Llama} 3 herd of models}.
\newblock \emph{Preprint}, arXiv:2407.21783.

\bibitem[{He et~al.(2023)He, Gao, and Chen}]{he-etal-2023-debertav3}
Pengcheng He, Jianfeng Gao, and Weizhu Chen. 2023.
\newblock \href {https://openreview.net/forum?id=sE7-XhLxHA} {{DeBERTaV3}: Improving {DeBERTa} using {ELECTRA}-style pre-training with gradient-disentangled embedding sharing}.
\newblock In \emph{International Conference on Learning Representations}.

\bibitem[{He et~al.(2021)He, Majumder, and McAuley}]{he-etal-2021-detect}
Zexue He, Bodhisattwa~Prasad Majumder, and Julian McAuley. 2021.
\newblock Detect and perturb: Neutral rewriting of biased and sensitive text via gradient-based decoding.
\newblock In \emph{Findings of the Association for Computational Linguistics: EMNLP 2021}, pages 4173--4181, Punta Cana, Dominican Republic.

\bibitem[{Holm(1979)}]{holm-1979-simple}
Sture Holm. 1979.
\newblock A simple sequentially rejective multiple test procedure.
\newblock \emph{Scandinavian Journal of Statistics}, 6(2):65--70.

\bibitem[{Honnibal et~al.(2020)Honnibal, Montani, Van~Landeghem, and Boyd}]{honnibal-etal-2020-spacy}
Matthew Honnibal, Ines Montani, Sofie Van~Landeghem, and Adriane Boyd. 2020.
\newblock \href {https://doi.org/10.5281/zenodo.1212303} {{spaCy}: Industrial-strength natural language processing in python}.

\bibitem[{Igamberdiev and Habernal(2023)}]{igamberdiev-habernal-2023-dp}
Timour Igamberdiev and Ivan Habernal. 2023.
\newblock \href {https://doi.org/10.18653/v1/2023.findings-acl.874} {{DP-BART} for privatized text rewriting under local differential privacy}.
\newblock In \emph{Findings of the Association for Computational Linguistics: ACL 2023}, pages 13914--13934, Toronto, Canada. Association for Computational Linguistics.

\bibitem[{Jin et~al.(2021)Jin, Pan, Oufattole, Weng, Fang, and Szolovits}]{jin-etal-2021-disease}
Di~Jin, Eileen Pan, Nassim Oufattole, Wei-Hung Weng, Hanyi Fang, and Peter Szolovits. 2021.
\newblock \href {https://doi.org/10.3390/app11146421} {What disease does this patient have? a large-scale open domain question answering dataset from medical exams}.
\newblock \emph{Applied Sciences}, 11(14):6421.

\bibitem[{Kim et~al.(2025)Kim, Jeon, and Shin}]{kim-etal-2025-self}
Kyuyoung Kim, Hyunjun Jeon, and Jinwoo Shin. 2025.
\newblock Self-refining language model anonymizers via adversarial distillation.
\newblock In \emph{Advances in Neural Information Processing Systems}.

\bibitem[{Krippendorff(2004)}]{krippendorff-2004-content}
Klaus Krippendorff. 2004.
\newblock \emph{Content Analysis: An Introduction to Its Methodology}, second edition.
\newblock Sage, Thousand Oaks, CA.

\bibitem[{Kwon et~al.(2023)Kwon, Li, Zhuang, Sheng, Zheng, Yu, Gonzalez, Zhang, and Stoica}]{kwon-etal-2023-efficient}
Woosuk Kwon, Zhuohan Li, Siyuan Zhuang, Ying Sheng, Lianmin Zheng, Cody~Hao Yu, Joseph~E. Gonzalez, Hao Zhang, and Ion Stoica. 2023.
\newblock \href {https://doi.org/10.1145/3600006.3613165} {Efficient memory management for large language model serving with {PagedAttention}}.
\newblock In \emph{Proceedings of the 29th Symposium on Operating Systems Principles}, pages 611--626. ACM.

\bibitem[{Lauscher et~al.(2022)Lauscher, Crowley, and Hovy}]{lauscher-etal-2022-pronouns}
Anne Lauscher, Archie Crowley, and Dirk Hovy. 2022.
\newblock Welcome to the modern world of pronouns: Identity-inclusive natural language processing beyond gender.
\newblock In \emph{Proceedings of the 29th International Conference on Computational Linguistics}, pages 1221--1232.

\bibitem[{Lison et~al.(2021)Lison, Pil{\'a}n, Sanchez, Batet, and {\O}vrelid}]{lison-etal-2021-anonymisation}
Pierre Lison, Ildik{\'o} Pil{\'a}n, David Sanchez, Montserrat Batet, and Lilja {\O}vrelid. 2021.
\newblock \href {https://doi.org/10.18653/v1/2021.acl-long.323} {Anonymisation models for text data: State of the art, challenges and future directions}.
\newblock In \emph{Proceedings of the 59th Annual Meeting of the Association for Computational Linguistics and the 11th International Joint Conference on Natural Language Processing (Volume 1: Long Papers)}, pages 4188--4203, Online. Association for Computational Linguistics.

\bibitem[{Liu et~al.(2023)Liu, Huang, Yu, Zhang, Wu, Cao, Dai, Zhao, Li, Shu, Zeng, Sun, Liu, Shen, Li, Liu, Zhu, and Li}]{liu-etal-2023-deid}
Zhengliang Liu, Yue Huang, Xiaowei Yu, Lu~Zhang, Zihao Wu, Chao Cao, Haixing Dai, Lin Zhao, Yiwei Li, Peng Shu, Fang Zeng, Lichao Sun, Wei Liu, Dinggang Shen, Quanzheng Li, Tianming Liu, Dajiang Zhu, and Xiang Li. 2023.
\newblock \href {https://doi.org/10.48550/arXiv.2303.11032} {{DeID-GPT}: Zero-shot medical text de-identification by {GPT}-4}.
\newblock \emph{arXiv preprint arXiv:2303.11032}.

\bibitem[{Loiseau et~al.(2026)Loiseau, Sileo, Riquet, Meyer, and Tommasi}]{loiseau-etal-2026-adaptive}
Gabriel Loiseau, Damien Sileo, Damien Riquet, Maxime Meyer, and Marc Tommasi. 2026.
\newblock \href {https://doi.org/10.18653/v1/2026.findings-acl.401} {Adaptive text anonymization: Learning privacy-utility trade-offs via prompt optimization}.
\newblock In \emph{Findings of the Association for Computational Linguistics: ACL 2026}, pages 8207--8229, San Diego, California, United States. Association for Computational Linguistics.

\bibitem[{Mahmood et~al.(2019)Mahmood, Ahmad, Shafiq, Srinivasan, and Zaffar}]{mahmood-etal-2019-girl}
Asad Mahmood, Faizan Ahmad, Zubair Shafiq, Padmini Srinivasan, and Fareed Zaffar. 2019.
\newblock \href {https://doi.org/10.2478/popets-2019-0058} {A girl has no name: Automated authorship obfuscation using {Mutant-X}}.
\newblock \emph{Proceedings on Privacy Enhancing Technologies}, 2019(4):54--71.

\bibitem[{Manzanares-Salor et~al.(2024)Manzanares-Salor, S{\'a}nchez, and Lison}]{manzanares-salor-etal-2024-evaluating}
Benet Manzanares-Salor, David S{\'a}nchez, and Pierre Lison. 2024.
\newblock \href {https://doi.org/10.1007/s10618-024-01066-3} {Evaluating the disclosure risk of anonymized documents via a machine learning-based re-identification attack}.
\newblock \emph{Data Mining and Knowledge Discovery}, 38:4040--4075.

\bibitem[{Mattern et~al.(2022)Mattern, Weggenmann, and Kerschbaum}]{mattern-etal-2022-limits}
Justus Mattern, Benjamin Weggenmann, and Florian Kerschbaum. 2022.
\newblock \href {https://doi.org/10.18653/v1/2022.findings-naacl.65} {The limits of word level differential privacy}.
\newblock In \emph{Findings of the Association for Computational Linguistics: NAACL 2022}, pages 867--881, Seattle, United States. Association for Computational Linguistics.

\bibitem[{Miller(1995)}]{miller-1995-wordnet}
George~A. Miller. 1995.
\newblock \href {https://doi.org/10.1145/219717.219748} {{WordNet}: A lexical database for {E}nglish}.
\newblock \emph{Communications of the ACM}, 38(11):39--41.

\bibitem[{Minot et~al.(2022)Minot, Cheney, Maier, Elbers, Danforth, and Dodds}]{minot-etal-2022-interpretable}
Joshua~R. Minot, Nicholas Cheney, Marc Maier, Danne~C. Elbers, Christopher~M. Danforth, and Peter~Sheridan Dodds. 2022.
\newblock Interpretable bias mitigation for textual data: Reducing genderization in patient notes while maintaining classification performance.
\newblock \emph{ACM Transactions on Computing for Healthcare}, 3(4):1--41.

\bibitem[{Mosallanezhad et~al.(2019)Mosallanezhad, Beigi, and Liu}]{mosallanezhad-etal-2019-deep}
Ahmadreza Mosallanezhad, Ghazaleh Beigi, and Huan Liu. 2019.
\newblock \href {https://doi.org/10.18653/v1/D19-1240} {Deep reinforcement learning-based text anonymization against private-attribute inference}.
\newblock In \emph{Proceedings of the 2019 Conference on Empirical Methods in Natural Language Processing and the 9th International Joint Conference on Natural Language Processing (EMNLP-IJCNLP)}, pages 2360--2369, Hong Kong, China. Association for Computational Linguistics.

\bibitem[{Mozes and Kleinberg(2021)}]{mozes-kleinberg-2021-intruder}
Maximilian Mozes and Bennett Kleinberg. 2021.
\newblock No intruder, no validity: Evaluation criteria for privacy-preserving text anonymization.
\newblock \emph{arXiv preprint arXiv:2103.09263}.

\bibitem[{Oh et~al.(2026)Oh, Kim, Oh, Kang, Kang, Wang, Park, Jung, and Kim}]{oh-etal-2026-subject}
Myeong~Seok Oh, Dong-Yun Kim, Hanseok Oh, Chaean Kang, Joeun Kang, Xiaonan Wang, Hyunjung Park, Young~Cheol Jung, and Hansaem Kim. 2026.
\newblock \href {https://doi.org/10.18653/v1/2026.acl-long.778} {Subject-level inference for realistic text anonymization evaluation}.
\newblock In \emph{Proceedings of the 64th Annual Meeting of the Association for Computational Linguistics (Volume 1: Long Papers)}, pages 17100--17135, San Diego, California, United States. Association for Computational Linguistics.

\bibitem[{{OpenAI}(2024)}]{openai-2024-gpt4omini}
{OpenAI}. 2024.
\newblock \href {https://openai.com/index/gpt-4o-mini-advancing-cost-efficient-intelligence/} {{GPT-4o} mini: Advancing cost-efficient intelligence}.

\bibitem[{Pezeshkpour and Hruschka(2024)}]{pezeshkpour-hruschka-2024-order}
Pouya Pezeshkpour and Estevam Hruschka. 2024.
\newblock Large language models sensitivity to the order of options in multiple-choice questions.
\newblock In \emph{Findings of the Association for Computational Linguistics: NAACL 2024}.

\bibitem[{Pil{\'a}n et~al.(2022)Pil{\'a}n, Lison, {\O}vrelid, Papadopoulou, S{\'a}nchez, and Batet}]{pilan-etal-2022-text}
Ildik{\'o} Pil{\'a}n, Pierre Lison, Lilja {\O}vrelid, Anthi Papadopoulou, David S{\'a}nchez, and Montserrat Batet. 2022.
\newblock \href {https://doi.org/10.1162/coli_a_00458} {The text anonymization benchmark ({TAB}): A dedicated corpus and evaluation framework for text anonymization}.
\newblock \emph{Computational Linguistics}, 48(4):1053--1101.

\bibitem[{Pil{\'a}n et~al.(2025)Pil{\'a}n, Manzanares-Salor, S{\'a}nchez, and Lison}]{pilan-etal-2025-truthful}
Ildik{\'o} Pil{\'a}n, Benet Manzanares-Salor, David S{\'a}nchez, and Pierre Lison. 2025.
\newblock \href {https://doi.org/10.1016/j.asoc.2025.114013} {Truthful text sanitization guided by inference attacks}.
\newblock \emph{Applied Soft Computing}, 185:114013.

\bibitem[{S{\'a}nchez and Batet(2016)}]{sanchez-batet-2016-sanitized}
David S{\'a}nchez and Montserrat Batet. 2016.
\newblock \href {https://doi.org/10.1002/asi.23363} {C-sanitized: A privacy model for document redaction and sanitization}.
\newblock \emph{Journal of the Association for Information Science and Technology}, 67(1):148--163.

\bibitem[{Shetty et~al.(2018)Shetty, Schiele, and Fritz}]{shetty-etal-2018-a4nt}
Rakshith Shetty, Bernt Schiele, and Mario Fritz. 2018.
\newblock \href {https://www.usenix.org/conference/usenixsecurity18/presentation/shetty} {{A4NT}: Author attribute anonymity by adversarial training of neural machine translation}.
\newblock In \emph{27th {USENIX} Security Symposium ({USENIX} Security 18)}, pages 1633--1650. USENIX Association.

\bibitem[{Simonsohn et~al.(2020)Simonsohn, Simmons, and Nelson}]{simonsohn-etal-2020-specification}
Uri Simonsohn, Joseph~P. Simmons, and Leif~D. Nelson. 2020.
\newblock \href {https://doi.org/10.1038/s41562-020-0912-z} {Specification curve analysis}.
\newblock \emph{Nature Human Behaviour}, 4(11):1208--1214.

\bibitem[{Smith(2009)}]{smith-2009-foundations}
Geoffrey Smith. 2009.
\newblock \href {https://doi.org/10.1007/978-3-642-00596-1_21} {On the foundations of quantitative information flow}.
\newblock In \emph{Foundations of Software Science and Computational Structures (FoSSaCS)}, volume 5504 of \emph{Lecture Notes in Computer Science}, pages 288--302. Springer.

\bibitem[{Staab et~al.(2024)Staab, Vero, Balunovi{\'c}, and Vechev}]{staab-etal-2024-beyond}
Robin Staab, Mark Vero, Mislav Balunovi{\'c}, and Martin Vechev. 2024.
\newblock \href {https://proceedings.iclr.cc/paper_files/paper/2024/hash/9028b8a3ca98f58e373f0c1497a17448-Abstract-Conference.html} {Beyond memorization: Violating privacy via inference with large language models}.
\newblock In \emph{International Conference on Learning Representations}.

\bibitem[{Staab et~al.(2025)Staab, Vero, Balunovi{\'c}, and Vechev}]{staab-etal-2025-anonymizers}
Robin Staab, Mark Vero, Mislav Balunovi{\'c}, and Martin Vechev. 2025.
\newblock \href {https://proceedings.iclr.cc/paper_files/paper/2025/file/f478b2e8ad9ff0756bf5b79fb31c330f-Paper-Conference.pdf} {Large language models are advanced anonymizers}.
\newblock In \emph{International Conference on Learning Representations}.

\bibitem[{Steegen et~al.(2016)Steegen, Tuerlinckx, Gelman, and Vanpaemel}]{steegen-etal-2016-multiverse}
Sara Steegen, Francis Tuerlinckx, Andrew Gelman, and Wolf Vanpaemel. 2016.
\newblock \href {https://doi.org/10.1177/1745691616658637} {Increasing transparency through a multiverse analysis}.
\newblock \emph{Perspectives on Psychological Science}, 11(5):702--712.

\bibitem[{Sun et~al.(2021)Sun, Webster, Shah, Wang, and Johnson}]{sun-etal-2021-they}
Tony Sun, Kellie Webster, Apu Shah, William~Yang Wang, and Melvin Johnson. 2021.
\newblock \href {https://arxiv.org/abs/2102.06788} {They, them, theirs: Rewriting with gender-neutral {E}nglish}.
\newblock \emph{Preprint}, arXiv:2102.06788.

\bibitem[{Vanmassenhove et~al.(2021)Vanmassenhove, Emmery, and Shterionov}]{vanmassenhove-etal-2021-neutral}
Eva Vanmassenhove, Chris Emmery, and Dimitar Shterionov. 2021.
\newblock \href {https://doi.org/10.18653/v1/2021.emnlp-main.704} {{N}eu{T}ral {R}ewriter: A rule-based and neural approach to automatic rewriting into gender neutral alternatives}.
\newblock In \emph{Proceedings of the 2021 Conference on Empirical Methods in Natural Language Processing}, pages 8940--8948. Association for Computational Linguistics.

\bibitem[{Wang et~al.(2022)Wang, Yang, Huang, Jiao, Yang, Jiang, Majumder, and Wei}]{wang-etal-2022-e5}
Liang Wang, Nan Yang, Xiaolong Huang, Binxing Jiao, Linjun Yang, Daxin Jiang, Rangan Majumder, and Furu Wei. 2022.
\newblock \href {https://arxiv.org/abs/2212.03533} {Text embeddings by weakly-supervised contrastive pre-training}.
\newblock \emph{Preprint}, arXiv:2212.03533.

\bibitem[{Yang et~al.(2025{\natexlab{a}})}]{yang-etal-2025-qwen3}
An~Yang et~al. 2025{\natexlab{a}}.
\newblock \href {https://arxiv.org/abs/2505.09388} {Qwen3 technical report}.
\newblock \emph{Preprint}, arXiv:2505.09388.

\bibitem[{Yang et~al.(2025{\natexlab{b}})Yang, Zhu, and Gurevych}]{yang-etal-2025-robust}
Tianyu Yang, Xiaodan Zhu, and Iryna Gurevych. 2025{\natexlab{b}}.
\newblock Robust utility-preserving text anonymization based on large language models.
\newblock In \emph{Proceedings of the 63rd Annual Meeting of the Association for Computational Linguistics (Volume 1: Long Papers)}, pages 28922--28941, Vienna, Austria.

\bibitem[{Yermilov et~al.(2023)Yermilov, Raheja, and Chernodub}]{yermilov-etal-2023-privacy}
Oleksandr Yermilov, Vipul Raheja, and Artem Chernodub. 2023.
\newblock Privacy- and utility-preserving {NLP} with anonymized data: A case study of pseudonymization.
\newblock In \emph{Proceedings of the 3rd Workshop on Trustworthy Natural Language Processing (TrustNLP 2023)}, pages 232--241, Toronto, Canada.

\bibitem[{Yukhymenko et~al.(2024)Yukhymenko, Staab, Vero, and Vechev}]{yukhymenko-etal-2024-synthetic}
Hanna Yukhymenko, Robin Staab, Mark Vero, and Martin Vechev. 2024.
\newblock \href {https://proceedings.neurips.cc/paper_files/paper/2024/hash/daa1816b84ca2d5051c87fb4d37dd540-Abstract-Datasets_and_Benchmarks_Track.html} {A synthetic dataset for personal attribute inference}.
\newblock In \emph{Advances in Neural Information Processing Systems 37: Datasets and Benchmarks Track}.

\bibitem[{Zhao et~al.(2018)Zhao, Wang, Yatskar, Ordonez, and Chang}]{zhao-etal-2018-gender}
Jieyu Zhao, Tianlu Wang, Mark Yatskar, Vicente Ordonez, and Kai-Wei Chang. 2018.
\newblock \href {https://doi.org/10.18653/v1/N18-2003} {Gender bias in coreference resolution: Evaluation and debiasing methods}.
\newblock In \emph{Proceedings of the 2018 Conference of the North {A}merican Chapter of the Association for Computational Linguistics: Human Language Technologies, Volume 2 (Short Papers)}, pages 15--20, New Orleans, Louisiana. Association for Computational Linguistics.

\bibitem[{Zheng et~al.(2024)Zheng, Zhou, Meng, Zhou, and Huang}]{zheng-etal-2024-selectors}
Chujie Zheng, Hao Zhou, Fandong Meng, Jie Zhou, and Minlie Huang. 2024.
\newblock Large language models are not robust multiple choice selectors.
\newblock In \emph{International Conference on Learning Representations}.

\end{thebibliography}

% =============================================================================
\ifappendix
\appendix

\section{The Task and the Editor in Full}
\label{app:taskeditor}

\paragraph{The task.}
Given protected subjects and attribute names, identify and edit residual cues in an anonymised document. The reported releases use a shared protected-attribute set for listed subjects. Although detection can in principle gate cues at the subject--attribute level, edit decisions in this work are aggregated by attribute, and heterogeneous policies are not evaluated.

\paragraph{Input.}
An upstream de-identifier produces a document $D$ in which identifying spans are replaced by typed placeholders. The editor receives $D$, a set of protected subjects $S$, mention forms $M(s)$ by which the text may refer to each $s\in S$, and a set of attribute names $A$. $S$ contains the people designated in our panel (1{,}793 in the 1{,}249 MedQA vignettes; 586 in the 144 TAB judgments), not every person mentioned. In our corpora $A=\{\attr{Sex},\attr{Occupation},\attr{Relationship}\}$; \attr{Relationship} enters the primary measure on SynthPAI only (Section~\ref{sec:eval}). Mention forms are those maintained by a release pipeline for the people it protects. The editor does not access the source text or any gold attribute value.

\paragraph{Residual cues.}
A \emph{residual cue} is a span $c=(i,j,a)$ of $D$ whose surface form states a value of attribute $a\in A$ for some entity: a gendered pronoun or form of address for \attr{Sex}, a kinship or marital term for \attr{Relationship}, or a role/workplace term for \attr{Occupation}. For \attr{Sex}, cues are instances of grammatical sex marking, which say nothing about gender identity (Ethical Considerations). A cue is \emph{bound} to $s$ when the entity whose attribute it states is $s$. The unit is a predication rather than a word: the same pronoun can refer to a protected subject in one sentence and to a clinician in the next, so cues are bound per occurrence.

\paragraph{Output and objective.}
The editor returns $D'$, obtained from $D$ via edits confined to cue spans and the agreement they disturb. Three requirements define a good edit. \emph{Privacy}: for every protected pair $(s,a)$, an adversary that reads $D'$ and is instructed which person to consider should be closer than on $D$ to what it would guess without any text, neither more confident in the true value nor steered toward an incorrect one. \emph{Utility}: content that states no protected attribute of a protected subject should survive. \emph{Scope}: cues bound to people outside $S$, and uses of attribute vocabulary that refer to no person, should not be edited; $S$ encodes the release policy on whom to protect. The baseline is $D$ itself, the text the de-identifier already released.

\paragraph{The editor.}
\sacc{} (Scoped Attribute-Cue Closure) is shown in Figure~\ref{fig:pipeline} and stated in Algorithm~\ref{alg:sacc}. Every step is deterministic; the only learned component is a dependency parser. \sacc{} reads only $D$, mention forms, and attribute names, and traces each edit to the cue and rule that triggered it. Two principles shape the editor: \emph{closure}, which puts the burden of proof on leaving a cue unchanged, and \emph{admissibility}, which rewrites a cue only to a counterpart that carries no value of any protected attribute (Section~\ref{sec:method}).

\begin{algorithm}[t]
\caption{\sacc{}$(D, S, M, A, P)$: the find half (lines 1--4) and the edit half (lines 5--6). \textsc{Dir}, \textsc{Rel}, \textsc{Unr}: direct, relational, unresolved; the other two statuses leave the cue.}
\label{alg:sacc}
\small
\begin{algorithmic}[1]
\State $Q \gets$ matches of $\Lambda_P[a]$ in $D$, $a\in A$ \Comment{sweep}
\State $\beta(c) \gets \textsc{Bind}(c,S,M)$ for $c\in Q$ \Comment{bind}
\State $H \gets \{c\in Q : \beta(c)\in\{\textsc{Dir},\textsc{Rel},\textsc{Unr}\}\}\setminus\mathcal{K}$ \Comment{closure}
\State resolve overlaps in $H$: \attr{Rel}${}>{}$\attr{Occ}${}>{}$\attr{Sex}
\State $D' \gets D$, each $c\in H$ rewritten to $\nu(D[c])$ or redacted \Comment{admissibility}
\State \Return $\textsc{Repair}_\rho(D')$ \Comment{agreement}
\end{algorithmic}
\end{algorithm}

A lexicon of closed lists per attribute and register is matched against the document (Figure~\ref{fig:lexicons} prints all patterns). For each match and each protected subject, a binder reads the dependency parse and returns one of five statuses: the cue describes the subject directly; it is a relational term whose possessor is the subject; its referent cannot be resolved; it describes another entity; or it is a non-person use (e.g., a role noun heading an organisation). \emph{Closure} edits every cue except the last two, so unresolved cues are edited rather than left unchanged; a short keep-list, shared with every editor we compare against, is never edited. \emph{Admissibility} then rewrites each selected cue only to a counterpart from a closed table whose outputs state no value of any protected attribute, and otherwise redacts it. A repair step re-inflects verbs affected by the rewrites. It has three settings, chosen per corpus (Appendix~\ref{app:final}): no repair; \emph{light}, which rewrites a cue only when its sentence isolates one table entry and otherwise leaves it redacted; and \emph{full}, which rewrites and repairs nearly every selected cue. Appendix~\ref{app:impl} gives the lexicons, binder rules, keep-list and rewrite table.

Closure is the main design choice that requires justification, because de-identification removes antecedents a binder would otherwise need: 41\% of cues edited on MedQA and 90\% on TAB have no resolved referent (Table~\ref{tab:binder}). Reading the \emph{source} text, a judge assigns 1{,}618 of the 1{,}659 unresolved TAB edit sites (97.5\%) to a protected person (Table~\ref{tab:binderllm}), so editing them is correct in almost all cases. The agreement setting is the method’s only tuned parameter, chosen per corpus by five-fold cross-fitting on the development adversary (Appendix~\ref{app:final}).

\section{Extended Editor Comparisons}
\label{app:probe}

\paragraph{Privacy--utility readings.}
Figure~\ref{fig:frontier} plots the fixed-protocol panel of Table~\ref{tab:main}. Pairwise contrasts use the entry-matching conventions in Appendix~\ref{app:protocol}.

\begin{figure}[t]
\centering
\safegraphic{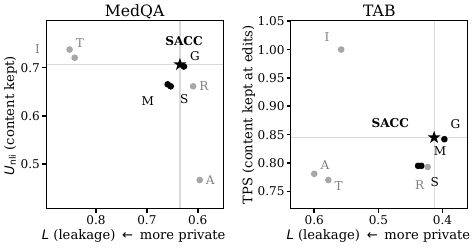}
\caption{\textbf{Privacy--utility comparison.} $L$ versus $U_{\mathrm{nli}}$ on MedQA and TPS on TAB, using the scores from Table~\ref{tab:main}. I: input; M: global mask; G: global ladder; S: SACC-mask; T: INTACT; A: Adaptive; R: LLM rewrite. \sacc{} is starred. Panel x-axes report different utility measures and scales.}
\label{fig:frontier}
\end{figure}

\paragraph{Against the global editors.}
Under the development adversary, neither global editor is demonstrably more private than \sacc{}. Global ladder is closest (MedQA $[-.021,+.003]$) and its 98\% intervals narrowly meet the $\pm.02$ equivalence criterion on all three MedQA privacy measures (lowest bound $-.0194$, on $L$; Table~\ref{tab:mde}); TAB cannot resolve differences below .04--.09. INTACT and Adaptive mainly target identifiers; \sacc{} is more private than INTACT on both corpora and Adaptive on TAB, and reduces leakage when applied after either (Table~\ref{tab:upstream}). Against same-input LLM rewrite, it is more private on MedQA \attr{Sex} under the development adversary but not separably under the held-out adversaries.

\paragraph{What was built on these corpora.}
The lexicons, keep-list and binder rules were written while reading these three corpora and the agreement setting was
cross-fitted on them, so the benchmark results characterise \sacc{} in the domains it was built in. Two measurements bound
its effect: MedQA rebuilt with TAB's lexicon changes five documents and the input reduction by .002, and the
documents nobody read while writing the selection rule gain as much as those that were read (Appendices~\ref{app:final}
and~\ref{app:protocol}). Generalisation is what the frozen run on two outside corpora tests (Appendix~\ref{app:unseen}):
there \sacc{} lowers its input's leakage by .35 and .11 while editing the fewest characters, and the ladder reaches
lower leakage by editing cues outside the protected set, the privacy--scope trade-off the task definition anticipates.
The ordering also survives an adversary given no candidates at all on MedQA, where it answers too rarely for the
reading to be strong; TAB fails that check's validity gate (Appendix~\ref{app:freeform}).

\paragraph{Which readings are themselves protocol-dependent.}
Two of the editor readings themselves depend on the interface. Rewriting's privacy gain was read
under all three adversary families and is mixed (Table~\ref{tab:increment}); closure's was read under the development
adversary alone, so its robustness to another adversary is untested. An adversary told each editor's rules reads least
from the global ladder, which the rule-unaware reading does not show (Appendix~\ref{app:ruleaware}).

\section{Every Pair of Texts}
\label{app:pairs}
Table~\ref{tab:multiverse2} gives the reading of Section~\ref{sec:multiverse} over every pair of the eight texts,
beside the perturbations that should not matter and the band width.

\begin{table}[htbp]
\centering
\small
\setlength{\tabcolsep}{3pt}
\begin{adjustbox}{max width=\linewidth}
\begin{tabular}{l ccc}
\toprule
& MedQA & TAB & SynthPAI \\
\midrule
Credit for a wrong belief & 5/28 & 6/28 & 0/28 \\
Candidate order & 4/84 & 5/84 & 4/84 \\
\midrule
Seed 1--5: mean / max rate & .007 / .012 & .012 / .024 & .024 / .036 \\
\midrule
$\delta=.02$ (\textsc{Disc}, \textsc{Leak}) & 3/56 & 9/56 & 2/56 \\
$\delta=.10$ (\textsc{Disc}, \textsc{Leak}) & 5/56 & 11/56 & 5/56 \\
\midrule
Re-read from the cache & 0/84 & 0/84 & 0/84 \\
\midrule
$\tau_b$: clip / order / re-read & .71 / 1.00 / 1.00 & 1.00 / 1.00 / 1.00 & .95 / .95 / .95 \\
\bottomrule
\end{tabular}
\end{adjustbox}
\caption{Every pair of the eight texts ($L$, \textsc{Disc}, \textsc{Leak}; unclipped $d$ for the clip): verdict changes under two choices, under two perturbations that should not matter (seed, re-read) and under the band $\delta$. Cells: changed/compared; seed: mean / maximum rate over seeds 1--5. Order and re-read use the \attr{Sex} entries the request cache rebuilds; $\tau_b$ over the seven texts sharing a cohort, where all three readings score them.}
\label{tab:multiverse2}
\end{table}

\paragraph{Magnitude and dependence (post hoc).}
For each protocol choice, we also summarise the absolute change in the paired point estimate, over the same
all-pairs contrasts Figure~\ref{fig:multiverse} counts. In units of $L$ the median shifts are .0236 for adversary
family ($N=504$), .0084 for clipping ($N=84$), .0059 for abstention ($N=75$) and .0003 for subject reference
($N=45$, both readings on one text version); candidate order's per-pair effect sizes are not stored, and subject position is read on AUC, so its shift
is not in units of $L$. Read against \sacc{} alone the ordering differs: abstention leads at .0330 on only 9 contrasts (subject position's .0388
there is likewise an AUC shift), as expected of a count over a handful of comparisons. The
summaries are descriptive and use different cohorts across choices.

\paragraph{Bootstrap computational check (post hoc).}
Re-reading cached requests changed no all-pairs verdict; reseeding 4{,}000 bootstrap draws changed 1--4\% of them. Exploratory one-sided comparisons of the all-pairs changed-verdict rates of Figure~\ref{fig:multiverse} against this \emph{Monte Carlo} rate gave unadjusted $p$, averaged over seeds, of .0000 for clipping and adversary family, .0001 for candidate order and abstention, .0003 for subject position and .019 for subject reference (.073 at the least favourable seed). These seed-dependent comparisons are not population-level null tests and do not support the main causal or inferential claims; paired effects and their intervals are reported instead.

\paragraph{Joint intervals for clipping (post hoc).}
For this choice, the readings share a paired cohort and bootstrap draw matrix, permitting simultaneous 99\% max-$t$
intervals across its system contrasts. On MedQA, 4 of 6 comparisons remain separable (versus 5 of 6 marginally);
TAB retains 3 of 7 and SynthPAI 2 of 6. MedQA clipping shifts the contrast toward \sacc{} by $+.0253$
$[+.0128,+.0382]$ against LLM rewrite, $+.0310$ $[+.0189,+.0440]$ against global mask, and $+.0314$
$[+.0191,+.0437]$ against SACC-mask. The other choices change the cohort between readings, so no joint interval is formed.

\section{SWAP: Construction Record}
\label{app:x1}

Section~\ref{sec:swap} reports the construction and its results; this appendix records the design, the pre-specified
predictions and the cost. SWAP is built from the scored SACC-mask text: within every region that SACC-mask replaced, each cue that the binder trace marks as
\attr{Sex} is rewritten to its opposite-sex form (\emph{he}/\emph{she}; \emph{her} as \emph{his} or \emph{him}, depending on its
dependency role; \emph{husband}/\emph{wife}; and the rest of the 19 sex forms the editor selects), and every other cue is written as
\texttt{[REDACTED]}. Thus SWAP and SACC-mask differ only on sex cues; written with \texttt{[REDACTED]} instead, the same
construction reproduces SACC-mask on 99.7\% of MedQA and all TAB documents. It was scored by the development adversary
using the same protocol as every other system (0.82 USD). Of four pre-specified predictions, three were confirmed; the fourth---that the
clipped leakage score $L$ would also credit SWAP on TAB---was not: there \leakE{} does and $L$ does not, so the credit the clip gives misdirection does not transfer across corpora.

\paragraph{The rule-aware reading, in full (post hoc).}
SWAP's flip is a fixed, invertible map, so an adversary informed of it should recover sex almost perfectly, grounding the clip
critique in quantitative information flow rather than in one constructed example alone. We described the map to
the development adversary (0.19 USD; the 1{,}513 MedQA \attr{Sex} entries). It
answers far more informatively, but not almost perfectly: mean posterior on the gold value rises from .252 to .769
(.802 of entries now favour the gold value over its complement), and the clipped $L$ rises with it,
from .240 to .654 ($+.414$ \,{\scriptsize[+.374,+.454]}, separable), the clearest demonstration that the clip's
floor at zero is a property of the scorer, not of what SWAP's text states. \discE{} barely moves ($-.021$
\,{\scriptsize[$-$.039,$-$.005]}, separable): it already charged SWAP's misled entries as disclosure, and telling
the adversary the map moves most of them from \misB{} to \leakB{}, both of which \discE{} counts; only a
stratum-balanced .021 move into the band instead. The rule-aware test confirms that SWAP's misdirection preserves recoverable attribute evidence; \discE{} flags
the belief shift that the clipped score ignores.

\paragraph{The same predictions under other scores.}
In a re-reading pre-specified after the main results (six of six predictions held), we read SWAP and
SACC-mask with the measures of Section~\ref{sec:related} on the 1{,}473 MedQA \attr{Sex} entries that the eight releases of
Table~\ref{tab:x1} share (Table~\ref{tab:metricfamily}). Recomputed first, $L$, \leakE{} and \discE{} of the eight releases
reproduce Table~\ref{tab:x1} exactly on MedQA and to .001 on TAB. Every reading that scores the adversary's answer about
the gold value alone---$L$, the signed residual, \leakE{}, attack accuracy, the Brier score and log loss---rates SWAP the
most private release on MedQA, and all but $L$ do so on TAB as well, where $L$ ranks it fourth: each credits a confident
false belief at least as much as no belief. The two that do not are two-sided displacement, which charges a shift in
either direction, and $|\mathrm{AUC}-.5|$, which exchanging the labels leaves unchanged. Read as a protection rate (one
minus accuracy, a single-attribute analogue of CPR and IPR, here balanced over the two sexes; \citealp{oh-etal-2026-subject}), SWAP scores .742 on MedQA
\attr{Sex} and SACC-mask .324, so an accuracy-based protection rate puts SWAP further ahead than $L$
does. CPR and IPR over all attributes are not computable here: both need the adversary's answer for every protected attribute of a subject, and the stored probability of the gold
value, renormalised over the candidates, does not recover the top answer to an eight-option \attr{Occupation} question.

\paragraph{How the readings are computed.}
Every reading uses the stored gold-value probability $\post$ of Section~\ref{sec:eval}: the mean over the eight seeded
orders of a probability renormalised over the candidates, in which a letter missing from the API's top 20 receives no
mass. Attack accuracy scores $\mathbf{1}[\post>.5]$, the Brier score $(1-\post)^2$, which lies between 0 and 1, and log
loss $-\ln\max(\post,\epsilon)$ in nats, with $\epsilon=10^{-12}$ applied to the averaged $\post$. No entry has $\post=0$,
but the floor binds on 20 of SWAP's 1{,}473 MedQA entries and 6 of its 244 TAB entries, and on 3 and 9 of Adaptive's.
Every reading except $|\mathrm{AUC}-.5|$ is aggregated exactly as $L$ is (Appendix~\ref{app:protocol}, \emph{Computing
the measures}): averaged within strata, which for \attr{Sex} are the two gold values, and then across them, so the 788
male and 685 female MedQA entries (187 and 57 on TAB) weigh equally and an accuracy of .258 is not a raw share. Each
document weighs one in the point estimates, and the 99\% intervals use the same 4{,}000 document draws as $L$.
$|\mathrm{AUC}-.5|$ is instead a rank statistic over the whole cohort, with document-clustered intervals.

\begin{table}[htbp]
\centering
\small
\setlength{\tabcolsep}{3pt}
\begin{tabular}{@{}l c cc c@{}}
\toprule
Reading & Prior & SWAP & SACC-mask & Rank M/T \\
\midrule
$L$ $\downarrow$ & yes & .241 & .495 & 1 / 4 \\
Signed residual $\downarrow$ & yes & $-$.608 & .285 & 1 / 1 \\
\leakE{} $\downarrow$ & yes & .252 & .626 & 1 / 1 \\
Attack accuracy $\downarrow$ & no & .258 & .676 & 1 / 1 \\
Brier score $\uparrow$ & no & .723 & .242 & 1 / 1 \\
Log loss $\uparrow$ & no & 14.51 & 1.33 & 1 / 1 \\
\midrule
\discE{} $\downarrow$ & yes & .980 & .897 & 6 / 5 \\
$|\mathrm{AUC}-.5|$ $\downarrow$ & no & .320 & .237 & 6 / 5$^*$ \\
\bottomrule
\end{tabular}
\caption{SWAP and SACC-mask under eight readings: the 1{,}473 MedQA \attr{Sex} entries the releases of
Table~\ref{tab:x1} share, seeded orders, development adversary. Prior: whether the reading uses the document-free prior. Arrows: the direction that would read as more private if the score were used as a privacy proxy; each keeps its usual meaning as a measure of prediction. Rank:
SWAP's place among the eight releases, most private first, on MedQA / TAB. Every MedQA difference is separable (99\%);
its paired differences and AUCs match Table~\ref{tab:x1} to .002. Log loss is in nats with a floor of $10^{-12}$. All rows but the last are balanced over the two sexes as $L$ is; the last is a rank statistic over the cohort. $^*$TAB first-named subjects; the aggregate TAB AUC mixes them with
subjects read through another person (Appendix~\ref{app:tabref}).}
\label{tab:metricfamily}
\end{table}

Told the rule, on the 1{,}513 entries of the rule-aware pass rather than the 1{,}473 of Table~\ref{tab:metricfamily} (so the untold baselines are .259 for accuracy and 14.45 for log loss), every gold-value reading reverses: accuracy rises from .259 to .805, the Brier score
falls from .723 to .146 and log loss from 14.45 to 0.58, all separable. The label-invariant reading does not weaken:
$|\mathrm{AUC}-.5|$ moves from .320 to .382 and stays .143 $[.110,.178]$ above SACC-mask's. Three qualifications were
recorded post hoc. On TAB the accuracy difference from SACC-mask ($-.113$) is not separable, while those on the Brier
score ($+.235$) and log loss ($+8.05$) are; with the sexes weighted equally, that accuracy difference leans on a stratum of
57 female entries weighed as heavily as the 187 male ones. The Brier score separates the ordinary editors by at most .002 on MedQA but
gives SWAP $+.482$, so nearly all of its sensitivity is to misdirection and it is no drop-in replacement. Finally, .688 of
SWAP's MedQA entries give the gold value a probability below .01, against .048 of \sacc{}'s, which is what drives the size
of its log loss. That size also depends on the floor, though the ranking does not: SWAP's MedQA log loss is 12.87 nats at
$\epsilon=10^{-9}$ and 14.51 at both $10^{-12}$ and $10^{-15}$ (SACC-mask 1.30, 1.33, 1.33; on TAB SWAP 9.64, 10.76,
10.77). The Brier score, bounded and floor-free, makes the same point and is the safer single reading.

\section{Prior Sensitivity}
\label{app:x2}

The document-free prior asks the question with a letter handle (``Subject A'') and no text, so for a two-valued attribute it encodes the
adversary's default answer (Section~\ref{sec:eval}). Table~\ref{tab:x2} replaces it, for MedQA \attr{Sex}, by the
adversary's answer on the document with every lexicon match masked. That document is the global mask's released text,
already scored with the same reference rule, orders and model, so the prior needs no new request; the entries are those of
Table~\ref{tab:goldvalue}, so only the prior inside $d$ changes. The comparison was pre-specified with three
predictions, two of which failed: the gap between female and male entries was predicted to halve and instead changed sign, and at most
two verdicts were predicted to move and three did, all of them SACC-mask against \sacc{}, all sign reversals. The
design did not anticipate what makes them uninformative: this prior is the global mask's output, and
on MedQA SACC-mask's output coincides with it on most documents, so those reversals are a property of how the prior was
built, not evidence that rewriting leaks more. What survives is the 39\% of entries on which the adversary, reading the masked document, still gives the true sex a high
probability: evidence that masking every lexicon match leaves in place (Appendix~\ref{app:pertag}, \emph{What still leaks}).

\begin{table}[htbp]
\centering
\small
\setlength{\tabcolsep}{2.6pt}
\begin{adjustbox}{max width=\linewidth}
\begin{tabular}{ll cccc}
\toprule
Gold & Prior & Input & Gl.\ ladder & SACC-mask & \sacc{} \\
\midrule
female (700) & document-free & .808 / .181 & .439 / .304 & .485 / .339 & .444 / .301 \\
female (700) & masked document & .785 / .114 & .265 / .206 & .223 / .034 & .274 / .213 \\
male (813) & document-free & .879 / .097 & .416 / .178 & .503 / .199 & .423 / .173 \\
male (813) & masked document & .855 / .044 & .123 / .321 & .059 / .034 & .132 / .326 \\
\bottomrule
\end{tabular}
\end{adjustbox}

\vspace{4pt}
\begin{adjustbox}{max width=\linewidth}
\begin{tabular}{l ccc ccc}
\toprule
& \multicolumn{3}{c}{Document-free prior} & \multicolumn{3}{c}{Masked-document prior} \\
\cmidrule(lr){2-4} \cmidrule(lr){5-7}
& $L$ & \textsc{Disc} & \textsc{Leak} & $L$ & \textsc{Disc} & \textsc{Leak} \\
\midrule
Input $-$ \sacc{} & \textbf{+.410} & \textbf{+.179} & \textbf{+.277} & \textbf{+.617} & \textbf{+.119} & \textbf{+.309} \\
Gl.\ ladder $-$ \sacc{} & $-$.006 & +.003 & $-$.001 & $-$.009 & $-$.007 & $-$.001 \\
SACC-mask $-$ \sacc{} & \textbf{+.061} & \textbf{+.085} & \textbf{+.053} & \textbf{$-$.062} & \textbf{$-$.382} & \textbf{$-$.147} \\
\bottomrule
\end{tabular}
\end{adjustbox}
\caption{MedQA \attr{Sex} (1,513 entries) against the document-free prior and a text-conditioned one: the same question asked of the document with every lexicon match masked. Top: $L$ / share misled by gold value. Bottom: system $-$ \sacc{}, 99\% intervals, bold excludes zero. This prior is the global mask's own output, identical to SACC-mask's on 1,009 of 1,249 documents and within $\delta$ of it on 92\% of entries, so the bottom block cannot rank masking against rewriting.}
\label{tab:x2}
\end{table}

A prior free of that confound needs a document no compared editor produces. Deleting every lexicon match, rather than
masking or rewriting it, gives one: the global mask's matches with the token and its space removed and nothing else
changed, scored once by the development adversary (0.10 USD). Masking and rewriting editors sit about equally close to
it (Table~\ref{tab:x3}), which a pre-specified neutrality check required before any comparison between them could be
reported. Against it \sacc{} still lowers its input's leakage and stays more private than SACC-mask on $L$ and
\leakE{}; \discE{} no longer separates the two, the one verdict of nine that changes, and none reverses. The prediction
that the $L$ difference would vanish failed. The value dependence persists under this prior too, in the direction the
masked-document prior gave it: male entries are now the more often misled under every editor.
For \sacc{}, 33\% of male and 19\% of female entries are misled;
the cue-deletion prior exceeds $.9$ on 41\% of entries.

\begin{table}[t]
\centering
\small
\setlength{\tabcolsep}{5pt}
\renewcommand{\arraystretch}{1.12}

\begin{tabular*}{\linewidth}{
  @{}l
  @{\extracolsep{\fill}}
  rrr@{}
}
\toprule
\textbf{System $-$ SACC}
& $\Delta L$
& $\Delta$\textsc{Disc}
& $\Delta$\textsc{Leak} \\
\midrule

\multicolumn{4}{@{}l}{
  \textit{Document-free prior}
} \\
\addlinespace[2pt]

Anonymised input
& \textbf{+.410}
& \textbf{+.179}
& \textbf{+.277} \\

Global ladder
& $-$.006
& +.003
& $-$.001 \\

SACC-mask
& \textbf{+.061}
& \textbf{+.085}
& \textbf{+.053} \\

\addlinespace[5pt]
\multicolumn{4}{@{}l}{
  \textit{Cue-deletion prior}
} \\
\addlinespace[2pt]

Anonymised input
& \textbf{+.625}
& \textbf{+.136}
& \textbf{+.314} \\

Global ladder
& $-$.010
& $-$.005
& $-$.005 \\

SACC-mask
& \textbf{+.054}
& $-$.032
& \textbf{+.033} \\

\bottomrule
\end{tabular*}

\vspace{5pt}
\parbox{\linewidth}{
  \footnotesize\raggedright
  \textit{Within $\delta$ of the cue-deletion prior:}
  Global mask .597;
  Global ladder .556;
  SACC-mask .584;
  SACC .552.
  For comparison, SACC-mask reaches .924
  under the masked-document prior.
}

\caption{\textbf{Cue-deletion prior sensitivity on MedQA
\attr{Sex}} (1,513 entries).
Values are system minus \sacc{}; bold indicates 99\%
intervals excluding zero. The prior does not reuse
a compared editor's output (Table~\ref{tab:x2}).}
\label{tab:x3}
\end{table}

\paragraph{No-document prior with a matched subject handle (post hoc).}
Unlike the text-conditioned checks above, this control uses no document in either prior. The canonical prior uses ``Subject A'', whereas the posterior uses the released text's surviving, cue-free reference. As a check on this question mismatch, we recompute the document-free prior using the posterior's subject handle. The reference itself can convey information even without the document and can vary by editor; this sensitivity analysis measures how subject-handle wording affects the document-free baseline.

\begin{table}[!htbp]
\centering
\small
\setlength{\tabcolsep}{2.1pt}
\renewcommand{\arraystretch}{1.12}
\begin{tabular}{@{}lcc@{}}
\toprule
& \textbf{MedQA} & \textbf{TAB} \\
\midrule
Mean prior $\prior$
& $.4916\to .5430$ & $.3874\to .3879$ \\
\sacc{} $L$
& $.4269\to .3867$ & $.2410\to .2407$ \\
Input $-$ \sacc{}
& \shortstack{$+.2711$ sep\\$\to +.2859$ sep}
& \shortstack{$+.2561$ sep\\$\to +.2563$ sep} \\
SACC-mask $-$ \sacc{}
& \shortstack{$+.0231$ ns\\$\to +.0290$ sep}
& \shortstack{$+.0741$ ns\\$\to +.0741$ ns} \\
\bottomrule
\end{tabular}
\caption{\textbf{Subject-handle sensitivity of the document-free prior.}
Each cell shows the canonical letter-handle prior $\to$ a document-free prior using the posterior's subject reference (post hoc). ``sep'' and ``ns'' denote separable and non-separable paired contrasts under the privacy-interval convention. All output texts and gold values are unchanged.}
\label{tab:matchedprior}
\end{table}

MedQA's prior rises by $.0514$ and \sacc{}'s $L$ falls by $.0402$; the corresponding changes on TAB are $+.0005$ and $-.0003$. The input--\sacc{} difference remains separable on both corpora, with a slightly larger point estimate. Of these reported comparisons, only SACC-mask versus \sacc{} on MedQA changes separability: $+.0231$ (non-separable) becomes $+.0290$ (separable), in \sacc{}'s favour. For the MedQA SACC-mask comparison, the letter-handle prior yields the smaller estimated advantage for \sacc{} under the development adversary. The MedQA--TAB difference indicates that a corpus's subject-reference wording can affect the baseline and hence absolute $L$; across-corpus levels should not be read as an editor-only effect, and we compare editors within a corpus.

\section{Further Related Work}
\label{app:related}

\paragraph{Anonymisation by spans.}
Span-based de-identification masks or generalises identifiers
\citep{lison-etal-2021-anonymisation,pilan-etal-2022-text,liu-etal-2023-deid} and is increasingly evaluated by
re-identification risk \citep{manzanares-salor-etal-2024-evaluating}; consistent pseudonyms keep the coreference a
binder could use \citep{yermilov-etal-2023-privacy}.

\paragraph{Anonymisation by rewriting.}
Language-model anonymisers rewrite under adversarial feedback \citep{staab-etal-2025-anonymizers}, replace detected
spans truthfully \citep{pilan-etal-2025-truthful}, optimise prompts \citep{loiseau-etal-2026-adaptive} or evaluators
\citep{yang-etal-2025-robust}, randomise the protected attribute \citep{frikha-etal-2025-incognitext}, abstract
self-disclosures \citep{dou-etal-2024-reducing} or distil anonymisation into small models \citep{kim-etal-2025-self};
earlier work obfuscated authorship and demographic style
\citep{shetty-etal-2018-a4nt,mahmood-etal-2019-girl,bevendorff-etal-2019-heuristic,faustini-etal-2025-idt}, and
differentially private rewriting trades fidelity for a formal guarantee
\citep{feyisetan-etal-2020-privacy,igamberdiev-habernal-2023-dp,mattern-etal-2022-limits}. \sacc{} calls no language
model, leaves upstream placeholders untouched and can run after such an anonymiser (Appendix~\ref{app:robust}).

\paragraph{Gender-neutral rewriting.}
Rewriting gendered English into neutral forms around singular \emph{they} is a task of its own
\citep{sun-etal-2021-they,vanmassenhove-etal-2021-neutral,amrhein-etal-2023-exploiting}; swapping gendered words is a
debiasing augmentation \citep{zhao-etal-2018-gender}, sensitive spans can be regenerated neutrally
\citep{he-etal-2021-detect}, and removing gendered language from clinical notes costs classification little
\citep{minot-etal-2022-interpretable}. These rewriters neutralise one attribute everywhere; \sacc{} decides which cues to
neutralise and admits only counterparts that state no protected value, and such a rewriter can serve as its rewrite
step (Appendix~\ref{app:final}).

% appendix float packing: more floats per page, smaller minimum text share, fuller float pages
\setcounter{topnumber}{4}\setcounter{bottomnumber}{3}\setcounter{totalnumber}{7}\setcounter{dbltopnumber}{4}
\renewcommand{\topfraction}{.92}\renewcommand{\bottomfraction}{.85}\renewcommand{\textfraction}{.06}
\renewcommand{\floatpagefraction}{.8}\renewcommand{\dbltopfraction}{.92}\renewcommand{\dblfloatpagefraction}{.8}

\section{Implementation Details}
\label{app:impl}

\paragraph{Counting conventions.}
The same corpus is counted two ways in this paper, and the difference is a convention, not a disagreement. A
\emph{cue} is one lexicon match under one attribute; an \emph{edit site} is a span the editor replaces once, so a span
matched under two attributes is one site. On TAB 70 spans are matched by both the kinship and the sex lexicon, and this
accounts for the whole difference: the binder trace's 3{,}510 TAB rows cover 3{,}440 spans, of which the keep-listed
\emph{the court} is 669, so TAB's 2{,}841 candidate cues are 2{,}771 sites and the 1{,}874 cues \sacc{} edits are
1{,}817 sites. Counts read from the binder trace (Table~\ref{tab:binder}) are per cue and before the keep-list, while
audited and judged populations are per site and after it: hence 1{,}678 unresolved TAB cues in the trace against
1{,}659 unresolved sites in the judged population, and 227 cues set aside as another person's against 214 sites. The two recall ranges, .84 to .90 from the binder
audit (Table~\ref{tab:scope}) and .82 to .88 in the placeholder-linking comparison below, are estimates of the same
quantity on different judged samples; we have not reconciled them further.

% >>> GENERATED scopemain
\begin{table}[t]
\centering
\small
\setlength{\tabcolsep}{3.0pt}
\begin{adjustbox}{max width=\linewidth}
\begin{tabular}{l r cc}
\toprule
Editor on TAB & Cues edited & Precision & Recall \\
\midrule
Global (\sacc{}'s lexicon) & 2,841 & .66--.73 & 1 \\
Global, person nouns only & 2,060 & .87--.96 & .90--.96 \\
Global + non-person rule & 2,101 & .85--.93 & .90--.96 \\
\rowcolor{oursrow}\sacc{} & 1,874 & .89--.96 & .84--.90 \\
\bottomrule
\end{tabular}
\end{adjustbox}
\caption{Scope on TAB, judged against the source text by three LLM judges on 40 sampled cues per binder decision (range over judges; Table~\ref{tab:scope}: per judge, and MedQA). Precision: share of edits on a protected person's cue; recall: share of their cues edited. The middle editors need no subject list. A span matched under two attributes is one edit site, so Section~\ref{sec:method} counts fewer sites (1,817 for \sacc{}, 2,771 for global editing; Appendix~\ref{app:impl}).}
\label{tab:scopemain}
\end{table}
% <<< GENERATED scopemain

\begin{figure*}[t]
\centering
\safegraphic{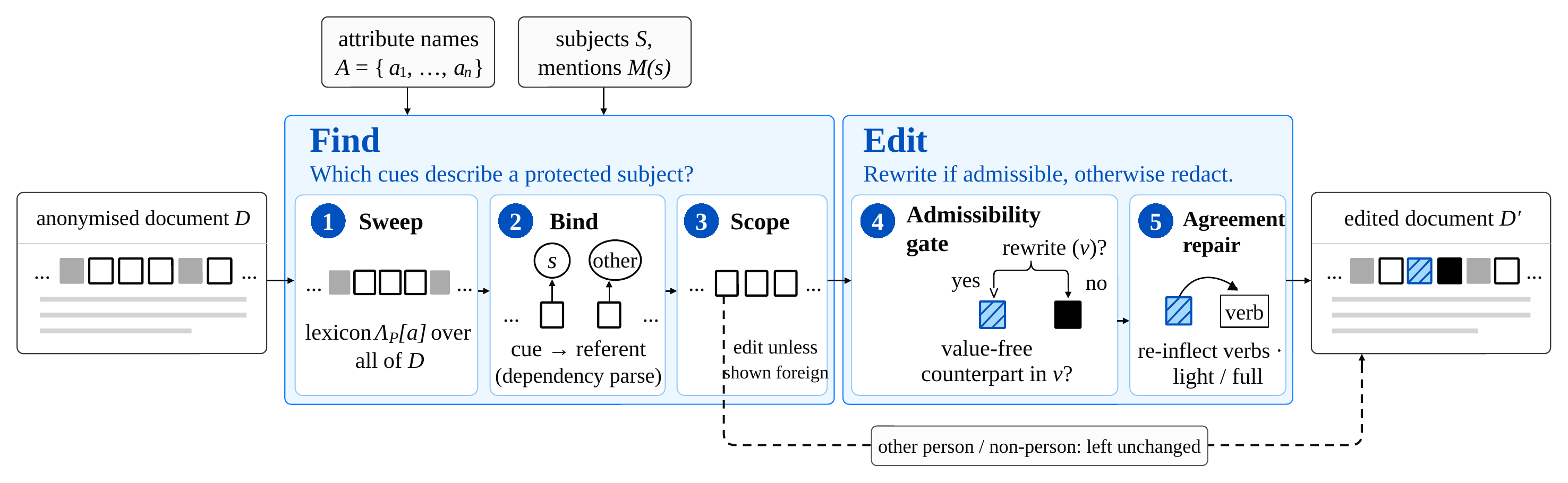}
\caption{\sacc{}: closure selects every cue the binder does not assign elsewhere; admissibility rewrites or redacts it. Grey:
placeholder; outlined: attribute cue; hatched: rewritten by $\nu$; black: redacted.}
\label{fig:pipeline}
\end{figure*}

% >>> GENERATED factorial
\begin{table*}[t]
\centering
\small
\setlength{\tabcolsep}{5pt}
\begin{adjustbox}{max width=\linewidth}
\begin{tabular}{ll cc cc}
\toprule
& & \multicolumn{2}{c}{MedQA} & \multicolumn{2}{c}{TAB} \\
\cmidrule(lr){3-4} \cmidrule(lr){5-6}
Change & at & $\Delta L$ & $\Delta$$U_{\mathrm{nli}}$ & $\Delta L$ & $\Delta$$U_{\mathrm{nli}}$ \\
\midrule
Closure over binding only & rewrite & \textbf{+.079}\,{\scriptsize[+.067,+.092]} & +.002\,{\scriptsize[$-$.002,+.007]} & \textbf{+.122}\,{\scriptsize[+.079,+.168]} & +.002\,{\scriptsize[$-$.005,+.009]} \\
Closure selection & mask & $-$.006\,{\scriptsize[$-$.019,+.003]} & \textbf{+.004}\,{\scriptsize[+.001,+.007]} & $-$.002\,{\scriptsize[$-$.051,+.023]} & \textbf{+.028}\,{\scriptsize[+.017,+.040]} \\
\quad non-person rule & mask & $-$.006\,{\scriptsize[$-$.018,+.001]} & \textbf{+.004}\,{\scriptsize[+.002,+.007]} & $-$.011\,{\scriptsize[$-$.061,+.015]} & \textbf{+.026}\,{\scriptsize[+.015,+.037]} \\
\quad other-person rule & mask & \phantom{+}.000\,{\scriptsize[$-$.004,+.005]} & $-$.001\,{\scriptsize[$-$.002,+.001]} & +.009\,{\scriptsize[$-$.001,+.021]} & +.002\,{\scriptsize[$-$.003,+.007]} \\
Closure selection & rewrite & \textbf{$-$.010}\,{\scriptsize[$-$.024,$-$.001]} & \phantom{+}.000\,{\scriptsize[$-$.002,+.003]} & +.001\,{\scriptsize[$-$.052,+.028]} & \textbf{+.026}\,{\scriptsize[+.014,+.039]} \\
Admissible rewriting & global & \textbf{+.027}\,{\scriptsize[+.016,+.040]} & \textbf{+.044}\,{\scriptsize[+.036,+.053]} & \textbf{+.020}\,{\scriptsize[+.004,+.036]} & +.002\,{\scriptsize[$-$.004,+.008]} \\
Admissible rewriting & closure & \textbf{+.024}\,{\scriptsize[+.012,+.035]} & \textbf{+.041}\,{\scriptsize[+.033,+.050]} & \textbf{+.022}\,{\scriptsize[+.006,+.039]} & \phantom{+}.000\,{\scriptsize[$-$.007,+.007]} \\
\midrule
NeuTral Rewriter & closure & $-$.003\,{\scriptsize[$-$.011,+.006]} & \textbf{$-$.004}\,{\scriptsize[$-$.009,$-$.001]} & $-$.001\,{\scriptsize[$-$.008,+.007]} & +.003\,{\scriptsize[$-$.001,+.007]} \\
Person-noun filter & selection & \textbf{+.004}\,{\scriptsize[+.001,+.009]} & \phantom{+}.000\,{\scriptsize[$-$.001,+.001]} & $-$.006\,{\scriptsize[$-$.016,+.005]} & $-$.001\,{\scriptsize[$-$.004,+.003]} \\
\bottomrule
\end{tabular}
\end{adjustbox}
\caption{What each principle contributes. $\Delta L$: privacy gain of the change (99\%); $\Delta$$U_{\mathrm{nli}}$: utility gain (95\%); bold excludes zero. Indented: the selection's two rules. Bottom: NeuTral Rewriter \citep{vanmassenhove-etal-2021-neutral} as rewrite step, the person-noun filter as selection; first and global-rewrite rows from a later pass (Appendix~\ref{app:impl}).}
\label{tab:factorial}
\end{table*}
% <<< GENERATED factorial

\paragraph{Lexicons.}
\attr{Sex}: personal pronouns (\emph{he, she, him, her, his, hers, himself, herself}), forms of address (\emph{Mr, Mrs,
Ms, Miss}), gendered kinship nouns (\emph{husband, wife, widow(er), father, mother, son, daughter, brother, sister}); the
clinical register adds sex nouns (\emph{man, woman, boy, girl, male, female, gentleman, lady}). \attr{Relationship}:
kinship and marital terms. \attr{Occupation}: role, workplace and employment-status terms. The first-person register
matches self-descriptions (\emph{as a guy}, \emph{my wife}, \emph{I work as}). Figure~\ref{fig:lexicons} gives every pattern verbatim.
Within an attribute, duplicate matches are collapsed longest-first
(e.g., \emph{mother} and \emph{father} occur in two MedQA \attr{Sex}
patterns). The frozen \texttt{retire[dment]*} expression is a character
class rather than the intended \texttt{retire(?:d|ment)?}; in these
corpora it matched only \emph{retired} and \emph{retirement}.
These patterns are retained for reproducibility; resolving duplicates
or removing keep-listed tokens would not change an applied edit.

% >>> GENERATED lexicons
\begin{figure*}[t]
\centering
\fbox{\parbox{0.97\textwidth}{\scriptsize\ttfamily\raggedright
\textrm{\textbf{MedQA (clinical, third person)}}\\
\textrm{\emph{Sex: pronouns and sex nouns}}: \textbackslash{}b(he|\allowbreak she|\allowbreak him|\allowbreak her|\allowbreak hers|\allowbreak his|\allowbreak himself|\allowbreak herself|\allowbreak man|\allowbreak woman|\allowbreak men|\allowbreak women|\allowbreak male|\allowbreak female|\allowbreak boy|\allowbreak girl|\allowbreak gentleman|\allowbreak lady|\allowbreak mother|\allowbreak father)\textbackslash{}b\\
\textrm{\emph{Sex: forms of address}}: \textbackslash{}b(Mr|\allowbreak Mrs|\allowbreak Ms|\allowbreak Miss)\textbackslash{}.?(?=\textbackslash{}s)\\
\textrm{\emph{Sex: gendered kinship nouns}}: \textbackslash{}b(husband|\allowbreak wife|\allowbreak widow(?:er)?|\allowbreak father|\allowbreak mother|\allowbreak son|\allowbreak daughter|\allowbreak brother|\allowbreak sister)\textbackslash{}b\\
\textrm{\emph{Relationship}}: \textbackslash{}b(wife|\allowbreak husband|\allowbreak spouse|\allowbreak widow(?:er)?|\allowbreak married|\allowbreak marriage|\allowbreak divorced?|\allowbreak partner|\allowbreak fianc[e\'{e}]e?|\allowbreak the couple|\allowbreak wedding)\textbackslash{}b\\
\textrm{\emph{Occupation}}: \textbackslash{}b(lawyer|\allowbreak solicitor|\allowbreak barrister|\allowbreak attorney|\allowbreak judge|\allowbreak prosecutor|\allowbreak notary|\allowbreak magistrate|\allowbreak law firm|\allowbreak the court|\allowbreak chambers|\allowbreak legal (?:practice|\allowbreak profession|\allowbreak adviser)|\allowbreak doctor|\allowbreak physician|\allowbreak nurse|\allowbreak surgeon|\allowbreak hospital|\allowbreak clinic|\allowbreak patient[s]? of|\allowbreak teacher|\allowbreak professor|\allowbreak school|\allowbreak university|\allowbreak engineer|\allowbreak company|\allowbreak business(?:man|\allowbreak woman)?|\allowbreak farm(?:er)?|\allowbreak unemploy(?:ed|\allowbreak ment)|\allowbreak pension(?:er)?|\allowbreak retire[dment]*|\allowbreak police|\allowbreak military|\allowbreak army|\allowbreak officer|\allowbreak minister|\allowbreak government|\allowbreak diplomat|\allowbreak politician|\allowbreak journalist|\allowbreak represent(?:ed|\allowbreak ing) (?:the|\allowbreak his|\allowbreak her|\allowbreak a)|\allowbreak patient|\allowbreak physician|\allowbreak resident|\allowbreak attending|\allowbreak intern|\allowbreak pharmacist|\allowbreak technician|\allowbreak therapist|\allowbreak paramedic|\allowbreak surgeon|\allowbreak midwife)\textbackslash{}b\\[0.5em]
\textrm{\textbf{TAB (third person)}}\\
\textrm{\emph{Sex: pronouns}}: \textbackslash{}b(he|\allowbreak she|\allowbreak him|\allowbreak her|\allowbreak hers|\allowbreak his|\allowbreak himself|\allowbreak herself)\textbackslash{}b\\
\textrm{\emph{Sex: forms of address}}: \textbackslash{}b(Mr|\allowbreak Mrs|\allowbreak Ms|\allowbreak Miss)\textbackslash{}.?(?=\textbackslash{}s)\\
\textrm{\emph{Sex: gendered kinship nouns}}: \textbackslash{}b(husband|\allowbreak wife|\allowbreak widow(?:er)?|\allowbreak father|\allowbreak mother|\allowbreak son|\allowbreak daughter|\allowbreak brother|\allowbreak sister)\textbackslash{}b\\
\textrm{\emph{Relationship}}: \textbackslash{}b(wife|\allowbreak husband|\allowbreak spouse|\allowbreak widow(?:er)?|\allowbreak married|\allowbreak marriage|\allowbreak divorced?|\allowbreak partner|\allowbreak fianc[e\'{e}]e?|\allowbreak the couple|\allowbreak wedding)\textbackslash{}b\\
\textrm{\emph{Occupation}}: \textbackslash{}b(lawyer|\allowbreak solicitor|\allowbreak barrister|\allowbreak attorney|\allowbreak judge|\allowbreak prosecutor|\allowbreak notary|\allowbreak magistrate|\allowbreak law firm|\allowbreak the court|\allowbreak chambers|\allowbreak legal (?:practice|\allowbreak profession|\allowbreak adviser)|\allowbreak doctor|\allowbreak physician|\allowbreak nurse|\allowbreak surgeon|\allowbreak hospital|\allowbreak clinic|\allowbreak patient[s]? of|\allowbreak teacher|\allowbreak professor|\allowbreak school|\allowbreak university|\allowbreak engineer|\allowbreak company|\allowbreak business(?:man|\allowbreak woman)?|\allowbreak farm(?:er)?|\allowbreak unemploy(?:ed|\allowbreak ment)|\allowbreak pension(?:er)?|\allowbreak retire[dment]*|\allowbreak police|\allowbreak military|\allowbreak army|\allowbreak officer|\allowbreak minister|\allowbreak government|\allowbreak diplomat|\allowbreak politician|\allowbreak journalist|\allowbreak represent(?:ed|\allowbreak ing) (?:the|\allowbreak his|\allowbreak her|\allowbreak a))\textbackslash{}b\\[0.5em]
\textrm{\textbf{SynthPAI (first person)}}\\
\textrm{\emph{Sex: self-descriptions}}: \textbackslash{}b(?:as a |\allowbreak i'?m a |\allowbreak being a )(guy|\allowbreak girl|\allowbreak man|\allowbreak woman|\allowbreak dude|\allowbreak lady|\allowbreak mom|\allowbreak mum|\allowbreak dad|\allowbreak father|\allowbreak mother|\allowbreak wife|\allowbreak husband|\allowbreak bloke)\textbackslash{}b\\
\textrm{\emph{Sex: forms of address}}: (none)\\
\textrm{\emph{Sex: gendered kinship nouns}}: \textbackslash{}bmy (husband|\allowbreak wife|\allowbreak boyfriend|\allowbreak girlfriend|\allowbreak fianc[e\'{e}]e?)\textbackslash{}b\\
\textrm{\emph{Relationship}}: \textbackslash{}b(my (?:husband|\allowbreak wife|\allowbreak partner|\allowbreak boyfriend|\allowbreak girlfriend|\allowbreak fianc[e\'{e}]e?|\allowbreak ex|\allowbreak spouse)|\allowbreak single|\allowbreak divorced?|\allowbreak engaged|\allowbreak married|\allowbreak marriage|\allowbreak widow(?:ed|\allowbreak er)?|\allowbreak in a relationship|\allowbreak we(?:'ve|\allowbreak  have) been together|\allowbreak my kids|\allowbreak my children)\textbackslash{}b\\
\textrm{\emph{Occupation}}: \textbackslash{}b(i work (?:as|\allowbreak at|\allowbreak in)|\allowbreak my (?:job|\allowbreak career|\allowbreak clients|\allowbreak patients|\allowbreak students|\allowbreak colleagues|\allowbreak boss|\allowbreak office|\allowbreak shift|\allowbreak company)|\allowbreak as a [a-z-]+(?:er|\allowbreak ist|\allowbreak or|\allowbreak ian|\allowbreak ant|\allowbreak ent)\textbackslash{}b|\allowbreak at work|\allowbreak day job|\allowbreak freelanc\textbackslash{}w+|\allowbreak self-employed|\allowbreak lawyer|\allowbreak nurse|\allowbreak doctor|\allowbreak teacher|\allowbreak engineer|\allowbreak designer|\allowbreak developer|\allowbreak analyst|\allowbreak chef|\allowbreak barista|\allowbreak architect|\allowbreak librarian|\allowbreak curator|\allowbreak therapist|\allowbreak technician|\allowbreak consultant|\allowbreak manager|\allowbreak artist|\allowbreak writer|\allowbreak researcher|\allowbreak scientist|\allowbreak professor|\allowbreak lecturer|\allowbreak student|\allowbreak retired|\allowbreak unemployed)\textbackslash{}b}}
\caption{\sacc{}'s frozen, case-insensitive cue lexicons
by corpus and attribute. Within-attribute overlaps are
resolved longest-first; keep-listed matches are not edited.}
\label{fig:lexicons}
\end{figure*}
% <<< GENERATED lexicons

\paragraph{Binder rules.}
In order: a non-person use if the cue is not relational and its head is an organisation, place, facility, event or law
entity or one of 26 institution nouns; otherwise the referent is the possessor of the head, the subject of a copular
complement, the head of an apposition, or the cue itself in argument position; a pronoun is direct when the document has
at most one subject, or when exactly one person or first-person mention occurs between 400 characters before the start
of the pronoun's sentence and the pronoun itself and that mention matches $M(s)$, and unresolved otherwise. The
400-character window was set by hand during development and its sensitivity was not measured; since an unresolved
pronoun is edited anyway, a narrower window can only move a pronoun from direct to unresolved, which leaves the edit in
place. The occupation pattern \texttt{represent(?:ed|ing) (?:the|his|her|a)} matched twice (\emph{representing the}, TAB),
never a pronoun, so overlap resolution never handed a possessive pronoun to the occupation branch. The 26
institution nouns are \emph{court, tribunal, hospital, clinic, ward, school, university, college, company, firm, chambers,
practice, government, ministry, department, office, police, army, military, bank, agency, council, prison, embassy,
institution, workplace}. The keep-list $\mathcal{K}$ holds the occupation expressions for which the generalisation
ladder has no truthful broader term, setting nouns and first-person phrases that state employment without naming a job:
\emph{the court, court, patient(s), at work, day job, I work as/at/in, as a laser} and \emph{my} followed by \emph{job,
career, clients, patients, students, colleagues, boss, office, shift} or \emph{company}. It exempts \emph{patient} 1{,}576
times on MedQA, \emph{the court} 669 times on TAB and 44 phrases on SynthPAI (\emph{at work} 34, \emph{day job} 4,
\emph{my job} 3, \emph{my career} 2, \emph{as a laser} 1), 42 of them the protected author's own; those 42 state that
the author is employed and stay in every editor's output. Mention matching is case-insensitive containment or equality with a token of
more than three characters. The lexicons were written on the three evaluation corpora and validated only by the leave-one-domain-out
run of Appendix~\ref{app:final}; their false matches surfaced case by case (\emph{as a laser}, matched by the SynthPAI
role suffix and placed on the keep-list, and the verb \emph{miss}, matched four times by the address pattern), so a
held-out estimate of their false-positive rate is not available. Table~\ref{tab:binder} counts the binder's decisions on every candidate cue.

% >>> GENERATED binder
\begin{table}[htbp]
\centering
\small
\setlength{\tabcolsep}{2.4pt}
\begin{adjustbox}{max width=\linewidth}
\begin{tabular}{ll r rrr rr}
\toprule
& & & \multicolumn{3}{c}{Edited} & \multicolumn{2}{c}{Left unchanged} \\
\cmidrule(lr){4-6} \cmidrule(lr){7-8}
Corpus & Attr. & Cues & Direct & Relat. & Unres. & Other & Non-pers. \\
\midrule
MedQA & Sex & 7,347 & 4,150 & 142 & 3,018 & 37 & 0 \\
 & Occ & 1,847 & 240 & 0 & 578 & 787 & 242 \\
 & Rel & 14 & 7 & 1 & 1 & 5 & 0 \\
\midrule
 & All & 9,208 & 4,397 & 143 & 3,597 & 829 & 242 \\
\midrule
TAB & Sex & 1,674 & 47 & 18 & 1,579 & 30 & 0 \\
 & Occ & 1,722 & 93 & 0 & 55 & 165 & 1,409 \\
 & Rel & 114 & 30 & 8 & 44 & 32 & 0 \\
\midrule
 & All & 3,510 & 170 & 26 & 1,678 & 227 & 1,409 \\
\midrule
SynthPAI & Sex & 8 & 8 & 0 & 0 & 0 & 0 \\
 & Occ & 98 & 87 & 0 & 0 & 10 & 1 \\
 & Rel & 72 & 31 & 35 & 4 & 2 & 0 \\
\midrule
 & All & 178 & 126 & 35 & 4 & 12 & 1 \\
\bottomrule
\end{tabular}
\end{adjustbox}
\caption{\textbf{Binder decisions by corpus and attribute},
before the keep-list and overlap resolution.
Unresolved cues are edited; cues assigned to another
person or a non-person use are left unchanged.
Applied edits number 7,339 / 1,817 / 120
(MedQA / TAB / SynthPAI). After the keep-list,
unresolved cues make up 41\% of MedQA and 90\% of
TAB's edited cues.}
\label{tab:binder}
\end{table}
% <<< GENERATED binder

\paragraph{Binder audit.}
A first audit against the source text found the binder sending kinship nouns of protected relatives to another person;
the mention-form rule, which binds a noun matching a protected person's mention form to that person, repairs this. We then audited the final binder on a fresh sample of 40
cues per decision and corpus, judged by \mbox{GPT-4o-mini}, \mbox{GPT-4.1-mini} and \mbox{GPT-4.1} against the source text
(Table~\ref{tab:binderaudit}). Decisions that lead to an edit are correct on almost every cue, .99 of the edited cues on
MedQA and .88 to .95 on TAB when weighted by population (the scope estimate of Table~\ref{tab:scope}, which
leaves out the keep-list and counts a relational cue as its possessor's, gives .89 to .96 on TAB). The remaining errors are sex and relationship nouns still sent
to another person (42 cues on MedQA, 62 on TAB), mostly relatives the benchmark describes only by a role or a name; the
MedQA occupation cues sent to another person are the noun \emph{patient}, which states no occupation. Pairwise, the
judges agree moderately to well on MedQA ($\kappa$ .64 to .90) and weakly on TAB (.49 to .58), so the TAB cells are
only indicative. Where GPT-4o-mini and GPT-4.1-mini both
attribute a sex cue that the binder sent to another person to a protected person of the document, that person's entry
keeps its leakage: on the 22 such MedQA entries, \sacc{} lowers $L$ from 1.000 to .930 (99\% interval of the gain
$[.000,.277]$), while global masking reaches .626 and \sacc{} lowers $L$ by .416 on the other 1{,}491 \attr{Sex}
entries. TAB has four such entries, too few to score.

\paragraph{Scope precision and recall.}
The audit's populations and labels give the precision and recall of the edits (Table~\ref{tab:scope}) over the cues
some system edits, stratified by binder decision and by whether the cue is a person noun; the keep-list, which no system
edits, is left out. A global editor edits every candidate, so its precision is the share of candidates that are a
protected person's. On TAB a quarter to a third of the global editor's edits fall on cues the judges assign to no person
(19\% to 33\% of its edits) or to another person (2\% to 9\%), against at most one in nine of \sacc{}'s, which edits
84\% to 90\% of the protected persons' cues. Two global editors that need no subject list close most of that gap: one
that applies only the binder's non-person rule, which is SACC-mask with the other-person rule turned off, reaches
precision .85 to .93 at recall .90 to .96, and under the development adversary it is not separable from SACC-mask in
privacy ($\Delta L$ $.000$ on MedQA, $+.009$ $[-.001,+.021]$ on TAB and $-.006$ $[-.019,+.000]$ on SynthPAI, so that
TAB's interval still leaves a difference of .02 undetected and the two corpora differ in sign) or in \unli{} (at most
.002 on any corpus; Table~\ref{tab:factorial}). Both arms of that comparison redact at their edit sites, so the rule
was not ablated against the rewrite step the reported system uses; one whose lexicon drops institution and activity
nouns reaches .87 to .96 at recall .90 to .96 and shares 2{,}027 of its 2{,}060 TAB edits with the first. The rule that
sets aside other people's cues, which does need the subject list, adds three to four points of precision and costs six to
eight of recall; it applies to 227 of the 2{,}841 TAB candidates some system edits. On MedQA nearly every candidate cue is the patient's, and the precisions differ by two to
three points. An editor that edited exactly the protected persons' cues would have precision and recall 1, so the most
any subject-aware selection could add over the person-noun editor on TAB is .128, .044 and .107 of precision and .038,
.048 and .104 of recall (GPT-4o-mini, GPT-4.1-mini, GPT-4.1). Three judges read the same sample and give the same reading; one of them is the development adversary's
model and GPT-4.1 is not used as an adversary anywhere, and they agree only moderately on TAB, so these are estimates to be
checked by human annotation.

% >>> GENERATED scope
\begin{table*}[tbp]
\centering
\small
\setlength{\tabcolsep}{2.6pt}
\begin{adjustbox}{max width=\linewidth}
\begin{tabular}{ll cccc ccc cc}
\toprule
& & \multicolumn{4}{c}{Precision} & \multicolumn{3}{c}{Recall} & \multicolumn{2}{c}{Global edits on} \\
\cmidrule(lr){3-6} \cmidrule(lr){7-9} \cmidrule(lr){10-11}
Corpus & Judge & \sacc{} & Gl.\ + non-person & Gl., person nouns & Global & \sacc{} & Gl.\ + non-person & Gl., person nouns & another person & no person \\
\midrule
MedQA & GPT-4o-mini & 1.000 & 1.000\,{\scriptsize[.999,1.000]} & 1.000\,{\scriptsize[.999,1.000]} & .980\,{\scriptsize[.975,.985]} & .983 & .988 & .988 & .001 & .019 \\
 & GPT-4.1-mini & 1.000 & .999\,{\scriptsize[.998,1.000]} & .999\,{\scriptsize[.998,1.000]} & .968\,{\scriptsize[.967,.970]} & .995 & .999 & .999 & .002 & .030 \\
 & GPT-4.1 & 1.000 & .999\,{\scriptsize[.998,.999]} & .999\,{\scriptsize[.998,.999]} & .970\,{\scriptsize[.967,.974]} & .992 & .997 & .997 & .002 & .028 \\
\midrule
TAB & GPT-4o-mini & .893\,{\scriptsize[.803,.963]} & .854\,{\scriptsize[.772,.921]} & .872\,{\scriptsize[.788,.940]} & .658\,{\scriptsize[.591,.719]} & .896 & .960 & .962 & .016 & .326 \\
 & GPT-4.1-mini & .960\,{\scriptsize[.912,.985]} & .930\,{\scriptsize[.881,.959]} & .956\,{\scriptsize[.908,.985]} & .729\,{\scriptsize[.682,.780]} & .869 & .944 & .952 & .070 & .201 \\
 & GPT-4.1 & .918\,{\scriptsize[.846,.986]} & .882\,{\scriptsize[.812,.944]} & .893\,{\scriptsize[.822,.956]} & .723\,{\scriptsize[.651,.795]} & .838 & .903 & .896 & .093 & .185 \\
\bottomrule
\end{tabular}
\end{adjustbox}
\caption{\textbf{Scope precision and recall} from source-text
binder audits (three LLM judges; 40 sampled cues per
decision), excluding the keep-list.
Precision is the protected-subject share of edited cues;
recall is the share of protected-subject cues edited
(relational cues count toward their possessor).
Gl.\ + non-person excludes non-person uses;
Gl., person nouns excludes institution and activity nouns.
Neither uses a subject list. ``Global edits on'' partitions
edits assigned to another person or to no person.
Brackets: 95\% intervals.}
\label{tab:scope}
\end{table*}
% <<< GENERATED scope

\paragraph{Closure and admissibility, ablated.}
Two arms, pre-specified after internal review and scored in one pass with the input, SACC-mask, \sacc{}, the global
mask and the global ladder, isolate the two principles (Table~\ref{tab:factorial}). \emph{Binding-only selection} is
\sacc{} with unresolved cues left unedited, everything else unchanged; it edits 4{,}304 of SACC-mask's 7{,}339 sites on
MedQA and 158 of 1{,}817 on TAB. It leaks more than \sacc{} on $L$ by .079 $[.067,.092]$ on MedQA and .122 $[.079,.168]$
on TAB, and on \discE{} by .048 and .064 (separable); the loss sits on entries whose subject has no bound cue (.208 and
.145) more than on the others (.020 and .081). On TAB it is barely more private than its input ($L$ $+.012$), and it keeps
no more content than \sacc{} on either corpus (\unli{} $-.002$ on both, not separable). \emph{Global with $\nu$} passes the global
mask through \sacc{}'s fixed rewrite step, so it differs from the global ladder only in how a selected cue is replaced.
It is not separably less private than the ladder on $L$ ($-.002$ on MedQA, $+.016$ on TAB) and keeps more on MedQA
(\unli{} $+.003$ $[.001,.006]$) but not on TAB ($-.003$, not separable); on TAB's aggregate \attr{Sex} measure the ladder
is more private ($+.028$, separable).
Against it, \sacc{}'s selection is .010 less private on MedQA ($[-.024,-.001]$; \leakE{} $-.013$) and not separable on
TAB, where it keeps more ($+.026$). The pre-specified predictions that binding-only selection leaks more on $L$ and
\discE{} on both corpora, and that global with $\nu$ is not separably less private than the ladder, were confirmed; those that
binding-only selection would keep more content, and global with $\nu$ more than the ladder on both corpora, were not. The
MedQA \unli{} of this pass ran on a different GPU model from the earlier passes and differs from their values on 3 to 6
of 1{,}247 documents per system (mean difference at most .0003); every contrast is read within the pass.

\paragraph{Richer subject knowledge.}
Three variants give the binder more subject knowledge than the rule binder has; none lets the subject list change the measures. Linking
every TAB \texttt{[PERSON]} placeholder to the person whose mention it replaced (375 decisions change; MedQA and SynthPAI
have no such placeholder) sends 285 pronouns to another person, the only one named nearby, and the three judges find all
but at most two of them to be protected persons' pronouns; leakage rises ($L$ $+.072$ $[.036,.114]$, \discE{} $+.041$)
and recall falls from .82--.88 to .69--.72. Replacing TAB's placeholders by consistent surname pseudonyms (849) doubles the
directly bound cues (325 against 170) but leaves 82\% unresolved, and with the other-person rule turned off \sacc{} is
equally private ($L$ $-.001$ $[-.011,+.007]$) and keeps as much (\unli{} $+.003$). Protecting only each judgment's
first-named subject, with GPT-4.1-mini labelling every candidate span from the source text with the audit's prompt
(Table~\ref{tab:primary}), finds that subject's cues with .86 to .95 precision on 1{,}450 sites but leaves the subject
more exposed than the global ladder ($L$ $+.101$), consistent with an adversary that reads other people's cues as the
subject's (Table~\ref{tab:subjectpos}). The pre-specified predictions that these variants would make the subject list matter
failed.

% >>> GENERATED primary
\begin{table*}[tbp]
\centering
\small
\setlength{\tabcolsep}{3.0pt}
\begin{adjustbox}{max width=\linewidth}
\begin{tabular}{l r cc cc cc}
\toprule
& & \multicolumn{2}{c}{Scope for the first-named subject} & \multicolumn{3}{c}{Privacy of the first-named subject} & \\
\cmidrule(lr){3-4} \cmidrule(lr){5-7}
Editor on TAB & Sites & Precision & Recall & $\Delta L$ & $\Delta$\textsc{Leak} & AUC & $U_{\mathrm{nli}}$ \\
\midrule
Anonymised input & 0 & --- & --- & \textbf{+.197}\,{\scriptsize[+.113,+.292]} & \textbf{+.098}\,{\scriptsize[+.019,+.194]} & .978 & --- \\
Global ladder & 2,771 & .39 / .49 & 1 / 1 & \textbf{$-$.101}\,{\scriptsize[$-$.250,$-$.013]} & $-$.083\,{\scriptsize[$-$.228,\phantom{+}.000]} & .589 & .296 \\
Global, person nouns & 1,990 & .55 / .61 & .99 / .88 & $-$.042\,{\scriptsize[$-$.120,+.029]} & $-$.033\,{\scriptsize[$-$.098,+.005]} & .619 & .318 \\
\sacc{} (all subjects, rule binder) & 1,817 & .59 / .66 & .99 / .89 & $-$.048\,{\scriptsize[$-$.129,+.004]} & \textbf{$-$.049}\,{\scriptsize[$-$.124,$-$.003]} & .617 & .319 \\
\sacc{}, first-named only, rule binder & 1,680 & .65 / .72 & .99 / .89 & $-$.004\,{\scriptsize[$-$.020,+.014]} & $-$.004\,{\scriptsize[$-$.020,\phantom{+}.000]} & .712 & .321 \\
\sacc{}, first-named only, LLM binder & 1,450 & .86 / .95 & .97 / .92 &  &  & .717 & .316 \\
\sacc{}, all subjects, LLM binder & 2,011 & .55 / .64 & .99 / .95 & $-$.042\,{\scriptsize[$-$.110,+.014]} & $-$.036\,{\scriptsize[$-$.099,+.001]} & .619 & .315 \\
\bottomrule
\end{tabular}
\end{adjustbox}
\caption{\textbf{Protecting TAB's first-named subject.}
Scope is precision/recall for the first-named subject (GPT-4o-mini /
GPT-4.1 judges); the LLM binder uses GPT-4.1-mini on
source text. Privacy covers 157 first-named entries.
$\Delta$ is system minus first-named-only \sacc{} with
the LLM binder (99\% intervals; bold excludes zero).
$U_{\mathrm{nli}}$ is computed over all documents and
cannot credit content retained about other subjects.}
\label{tab:primary}
\end{table*}
% <<< GENERATED primary

\paragraph{Privacy by binding decision.}
Grouping each entry by the binder's decisions on its subject's cues shows where the gain comes from. Under the
development adversary, \sacc{} lowers its input's $L$ by .209 on MedQA entries with a direct or relational binding and by
.214 on entries whose subject has no bound cue but some unresolved one (99\% intervals $[.187,.231]$ and $[.186,.244]$; 1{,}009 and 722 entries);
on TAB by .102 and .168 ($[.027,.186]$, $[.113,.232]$; 91 and 259 entries). Entries with no selected cue gain nothing on
either corpus. The conservative rule that edits unresolved cues therefore carries as much of the gain as positive
binding does. The selection depends little on complete mention forms: dropping each at random with probability .5 never
adds an edit and removes 2.1\% of them on MedQA and 3.0\% on TAB, the kinship and role nouns bound through a mention form
(a pre-specified prediction of under 2\% on each corpus failed). One seed was drawn, and the reduced inputs were not
scored for privacy.
Nor does it depend much on a complete subject list: removing each protected subject from $S$ at random with probability
.5 (882 of 1{,}793 on MedQA, 266 of 586 on TAB) never adds an edit and removes 2.1\% of them on MedQA and 2.5\% on TAB,
because the cues of a subject missing from the list mostly become unresolved and are still edited. An attribute missing
from $A$ is not edited at all.

\paragraph{Runtime.}
On one core of an Intel Xeon Gold 6230R, our implementation takes 0.17\,s per MedQA document (120 words on average) and
0.54\,s per TAB document (573 words), excluding the upstream de-identifier. Most of it is dependency parsing, which this
implementation repeats seven times per document because it writes several ablation variants in one pass; binding takes 2 and
74\,ms per document and the rewrite step 1 and 9\,ms. No step calls a model or a GPU.

% >>> GENERATED binderaudit
\begin{table}[htbp]
\centering
\small
\setlength{\tabcolsep}{2.0pt}
\begin{adjustbox}{max width=\linewidth}
\begin{tabular}{ll rr cccc}
\toprule
Corpus & Binder decision & Cues & $n$ & 4o-mini & 4.1-mini & 4.1 & Agree \\
\midrule
MedQA & Direct & 4,397 & 40 & .975 & .975 & .975 & .975 (40) \\
 & Relational & 143 & 40 & 1.000 & 1.000 & 1.000 & 1.000 (39) \\
 & Unresolved & 3,597 & 40 & 1.000 & 1.000 & 1.000 & 1.000 (40) \\
 & Other person, sex/rel. & 42 & 40 & .075 & .200 & .225 & .061 (33) \\
 & Other person, occ. & 787 & 40 & .025 & .025 & .025 & .000 (39) \\
 & Non-person & 242 & 40 & .625 & .975 & .900 & .955 (22) \\
\midrule
TAB & Direct & 170 & 40 & .925 & .925 & .950 & .973 (37) \\
 & Relational$^\dagger$ & 26 & 26 & 1.000 & 1.000 & 1.000 & 1.000 (25) \\
 & Unresolved & 1,678 & 40 & .875 & .950 & .900 & .946 (37) \\
 & Other person, sex/rel. & 62 & 40 & .275 & .150 & .150 & .000 (23) \\
 & Other person, occ. & 165 & 40 & .550 & .350 & .550 & .000 (8) \\
 & Non-person & 1,409 & 40 & .900 & .900 & .850 & .935 (31) \\
\bottomrule
\end{tabular}
\end{adjustbox}
\caption{Fresh audit of the final binder: share of sampled cues whose decision is right, judged against the source text by GPT-4o-mini, GPT-4.1-mini and GPT-4.1. Cues: the decision's population, keep-list included; Agree: the share on which all three judges give the same label (their number in brackets). Right: direct, the bound subject; relational, a person; unresolved, a protected person; left unchanged, not a protected person. $^\dagger$TAB's 26 relational cues are read from the first audit's labels; GPT-4.1 judged them separately. Cohen's $\kappa$ (4o-mini/4.1-mini, 4o-mini/4.1, 4.1-mini/4.1): MedQA .670, .637, .899; TAB .510, .493, .576.}
\label{tab:binderaudit}
\end{table}
% <<< GENERATED binderaudit

\paragraph{What the unresolved cues are.}
Closure edits a cue whose referent the binder cannot resolve, and on TAB those are 90\% of what it edits
(Table~\ref{tab:binder}), so what they describe determines whether default-deny is appropriate. GPT-4.1-mini,
given the source document and the list of protected persons and allowed to answer \textsc{unclear}, labelled all
2{,}771 TAB candidate edit sites and answered \textsc{unclear} on none (Table~\ref{tab:binderllm}). Of the 1{,}659 the rule
binder leaves unresolved it assigns 1{,}618, or 97.5\%, to a protected person, 8 to a person outside the set and 33 to
no person: editing them is correct in almost all cases, and the binding-only arm of Table~\ref{tab:factorial} skips 1{,}618
sites that do describe someone the release protects. The same comparison bounds the other direction: of the 214 sites the
binder sets aside as another person's, the judge assigns 145 to a protected person, the over-exclusion the two unseen
corpora measure as leakage (Appendix~\ref{app:unseen}). This judge reads the source text, with the names
de-identification removed, so it is not a binder the editor could use; what it shows is that the unresolved cues are
unresolved for the released text rather than undecidable in principle, and that a binder able to resolve them would
leave closure nothing to do on this corpus.

% >>> GENERATED binderllm
\begin{table}[htbp]
\centering
\small
\setlength{\tabcolsep}{2.6pt}
\begin{adjustbox}{max width=\linewidth}
\begin{tabular}{l rrr rc}
\toprule
& \multicolumn{3}{c}{Judge reading the source says} & & \\
\cmidrule(lr){2-4}
\sacc{}'s binder says & protected & not protected & no person & All & Share prot. \\
\midrule
Resolved to a protected subject & 152 & 1 & 5 & 158 & .962 \\
\rowcolor{oursrow}Unresolved & 1,618 & 8 & 33 & 1,659 & .975 \\
Another person & 145 & 38 & 31 & 214 & .678 \\
Non-person use & 95 & 76 & 569 & 740 & .128 \\
\bottomrule
\end{tabular}
\end{adjustbox}
\caption{\sacc{}'s rule binder against GPT-4.1-mini reading the \emph{source} document and the protected persons, on all 2,771 TAB candidate edit sites: the LLM binder of Table~\ref{tab:primary}, here with every annotated person protected, which could answer \textsc{unclear} and never did. Share prot.: the row's share the judge assigns to a protected person, not ground truth.}
\label{tab:binderllm}
\end{table}
% <<< GENERATED binderllm

\paragraph{Rewrite table.}
Table~\ref{tab:nu} lists $\nu$ in full. Its 85 surface forms map to 30 outputs: epicene pronouns, \emph{the person} for
forms of address, unmarked person and kinship nouns, and the unmarked counterparts of a few sex-marked role nouns. Both
agreement settings use the same table. The admissibility check is implemented as this table plus a block on outputs that state a
relationship status (\emph{spouse}, \emph{surviving spouse}, \emph{partner}); the table is not re-checked against every
attribute in $A$ at run time, and one entry escapes it: an unmarked role noun such as \emph{businessperson} still states
an occupation. In the reported texts this happens
once (one MedQA \emph{businessman}, an \attr{Occupation} cue, became \emph{businessperson}; TAB has none), so the
admissibility of Section~\ref{sec:method} holds for all other applied rewrites.

% >>> GENERATED neutral
\begin{table}[htbp]
\centering
\small
\setlength{\tabcolsep}{3pt}
\begin{adjustbox}{max width=\linewidth}
\begin{tabular}{l p{0.68\linewidth}}
\toprule
Output & Surface forms rewritten to it \\
\midrule
\emph{they} & he, she \\
\emph{them} & him, her \\
\emph{their} & his \\
\emph{themselves} & himself, herself \\
\emph{the person} & mr, ms, mr., ms., mrs, mrs., sir, madam \\
\emph{person} & gentleman, lady, man, woman, male, female, guy, gal \\
\emph{people} & gentlemen, ladies, men, women, males, females \\
\emph{child} & boy, girl, son, daughter \\
\emph{children} & boys, girls, sons, daughters \\
\emph{parent} & father, mother, dad, mom \\
\emph{sibling} & brother, sister \\
\emph{siblings} & brothers, sisters \\
\emph{spouse} & husband, wife \\
\emph{relative} & uncle, aunt, nephew, niece \\
\emph{grandparent} & grandfather, grandmother \\
\emph{grandchild} & grandson, granddaughter \\
\emph{surviving spouse} & widower, widow \\
\emph{monarch} & king, queen \\
\emph{royal} & prince, princess \\
\emph{server} & waiter, waitress \\
\emph{performer} & actor, actress \\
\emph{businessperson} & businessman, businesswoman \\
\emph{chair} & chairman, chairwoman \\
\emph{spokesperson} & spokesman, spokeswoman \\
\emph{partner} & boyfriend, girlfriend, fiance, fiancee \\
\emph{godparent} & godfather, godmother \\
\emph{stepparent} & stepfather, stepmother \\
\emph{stepchild} & stepson, stepdaughter \\
\emph{youth} & lad, lass \\
\emph{parental} & paternal, maternal \\
\bottomrule
\end{tabular}
\end{adjustbox}
\caption{The rewrite table $\nu$: 85 surface forms, 30 outputs. Case follows the source; \emph{her} becomes \emph{their} or \emph{them}, and \emph{his} \emph{their} or \emph{theirs}, by whether a noun phrase follows (a word, a placeholder or a number). \emph{Spouse}, \emph{surviving spouse} and \emph{partner} state a relationship status, so they are not admissible while \attr{Relationship} is protected and those cues stay redacted. Forms not listed stay redacted.}
\label{tab:nu}
\end{table}
% <<< GENERATED neutral

\paragraph{Value asymmetry of the edits.}
Table~\ref{tab:sidechannel} splits \sacc{}'s edits of sex cues by slot and by the gender of the original form. On
MedQA, where the full setting rewrites almost everything, the forms are treated alike, with two exceptions: object
\emph{her} became \emph{their} 29 times, an error \emph{him} cannot make, and the address pattern matched the verb
\emph{miss} four times, each redacted. On TAB the light setting leaves more feminine than masculine pronouns
redacted (25\% against 16\% of edited pronouns), because object and possessive \emph{her} share a form: a redacted
pronoun raises the odds of a feminine referent by a likelihood ratio of about 1.6. The light setting rewrites a
redaction only when it can isolate one table entry, which a sentence with an ambiguous \emph{her} prevents; this also
explains the subject pronouns, since \emph{she} stays redacted in 26\% of sentences that also hold \emph{her} and in 12\%
of the others, against 13\% for \emph{he}. \emph{Miss} and \emph{hers} are not in $\nu$ and stay redacted;
the title never occurs among the edited cues and \emph{hers} once. The pattern also matched one \emph{ms} meaning milliseconds,
which the binder left unedited, and no upper-case \emph{MS}.

% >>> GENERATED sidechannel
\begin{table}[htbp]
\centering
\small
\setlength{\tabcolsep}{2.6pt}
\begin{adjustbox}{max width=\linewidth}
\begin{tabular}{ll rc rc}
\toprule
& & \multicolumn{2}{c}{Feminine form} & \multicolumn{2}{c}{Masculine form} \\
\cmidrule(lr){3-4} \cmidrule(lr){5-6}
Corpus & Slot & Edited & Redacted & Edited & Redacted \\
\midrule
MedQA & subject pronoun & 1,619 & .000 & 2,031 & .005 \\
 & object pronoun & 160 & .000 & 71 & .000 \\
 & possessive determiner & 1,306 & .000 & 1,754 & .001 \\
 & possessive pronoun & 1 & 1.000 & 8 & .000 \\
 & reflexive & 8 & .000 & 18 & .000 \\
 & address & 4 & 1.000 & 0 & --- \\
 & noun & 218 & .000 & 112 & .000 \\
\midrule
TAB & subject pronoun & 113 & .195 & 559 & .131 \\
 & object pronoun & 23 & .348 & 168 & .149 \\
 & possessive determiner & 129 & .271 & 551 & .191 \\
 & possessive pronoun & 0 & --- & 5 & .000 \\
 & reflexive & 6 & .500 & 5 & .200 \\
 & noun & 57 & .825 & 28 & .857 \\
\bottomrule
\end{tabular}
\end{adjustbox}
\caption{\sacc{}'s edits of sex cues by grammatical slot and gender of the original form: edits, and the share left redacted rather than rewritten. Slots from the parse of the input; MedQA's feminine address forms are the verb \emph{miss}, matched by the address pattern.}
\label{tab:sidechannel}
\end{table}
% <<< GENERATED sidechannel

\paragraph{Applied edits.}
\label{sec:applicability}How often rewriting applies is known before any adversary is run, and it orders the corpora as the privacy gain of
rewriting does: first-person posts have no epicene counterpart for their cues.
Over the corpora, \sacc{} applies 7{,}339 / 1{,}817 / 120 edits (MedQA / TAB / SynthPAI); the rewrite share is .994 on
MedQA, .716 on TAB with the light setting and .000 on SynthPAI. The binder's mention-form rule adds 157 edits on MedQA and
42 on TAB and lowers the TAB share from .751; before that, the fix of Appendix~\ref{app:history} had returned 21
marital rewrites to redactions, lowering it from .763 (.901 with the full setting).

\section{Baselines}
\label{app:baselines}

\paragraph{Global mask and global ladder.}
Both run \sacc{}'s sweep over the whole document and skip the bind and scope steps, so every lexicon match is edited,
whoever it describes, except the expressions on \sacc{}'s keep-list $\mathcal{K}$, which these editors leave in place as \sacc{}
does. Global masking writes a redaction over each match. The global ladder applies a fixed generalisation
ladder: sex cues are rewritten to epicene pronouns and unmarked nouns, with
the verb agreement the pronoun needs, and forms of address are dropped; occupation and relationship cues are replaced by
the first rung above them, a sector term for a role noun or a generic kinship or partnership term for a relational noun,
and are redacted when the ladder has no rung or the cue is negated. A short list of occupational expressions for which
no truthful more general term exists is left in place. The ladder's sex rewrites are therefore close to \sacc{}'s, while its occupation
and relationship rewrites still name a value of the attribute, which \sacc{}'s admissibility rule does not allow, so \sacc{} redacts those cues.

\paragraph{INTACT.}
INTACT \citep{pilan-etal-2025-truthful} replaces each span of the detector with a truthful, less specific description
written by a language model. We run the released implementation and prompts with \mbox{Llama-3.1-8B-Instruct} on the same
detector spans that produce $D$, so its edit sites relative to $D$ are the detector's placeholders, which it fills with
text derived from the source.

\paragraph{Adaptive.}
We use the published outputs of Adaptive \citep{loiseau-etal-2026-adaptive} for \mbox{Qwen3-30B-A3B-Instruct-2507} with its
released prompts. They are produced from the source text by Adaptive's own pipeline, with its own placeholders, not from
$D$. Its MedQA outputs were scored by the development adversary in a separate pass; the two systems scored
in both passes agree to a mean absolute posterior difference below .0001.

\paragraph{LLM rewrite.}
The LLM rewrite receives what \sacc{} receives and nothing else: the anonymised document, each protected subject's mention
forms and the names of its protected attributes. The prompt (Figure~\ref{fig:llmprompt}) was revised twice on a
development set of 173 documents and frozen before any scored call. The scored documents are those outside that
set: 91 of TAB, 221 of SynthPAI, 17 of which have no protected subject and pass through unchanged, and all 1{,}189 of
MedQA, 19 of which likewise pass through. A call is retried up to three times, and a document fails when the reply is
not valid JSON, does not end normally, or is shorter than half or longer than one and a half times the input; no
document failed. The model changed at least one upstream placeholder in 10.4\%, 16.5\% and 3.4\% of the documents it
was called on (MedQA, TAB, SynthPAI). The TAB and SynthPAI samples and a first sample of 200 MedQA documents cost 0.14 USD,
at a median of 1.9 seconds per MedQA document, and the remaining MedQA documents 0.19 USD.

% >>> GENERATED prompt
\begin{figure*}[t]
\centering
\fbox{\parbox{0.97\textwidth}{\scriptsize\ttfamily\raggedright
\textrm{\textbf{System.}}\\
You edit anonymized documents. You change only the expressions you are asked to change and copy every other character exactly.\\[0.5em]
\textrm{\textbf{User.}}\\
This document has already been anonymized: bracketed placeholders stand for spans that were removed earlier.\\
\rule{0pt}{0.6em}\\
Protected people (their mention forms) and the attributes to hide for each:\\
\{people\}\\
\rule{0pt}{0.6em}\\
Placeholders in this document, with the number of times each occurs: \{placeholders\}\\
\rule{0pt}{0.6em}\\
Edit the document so that it no longer reveals any listed attribute of any listed person:\\
- SEX: words that reveal the person's sex or gender, such as pronouns, titles and gendered nouns.\\
- OCCUPATION: words that reveal the person's job or profession.\\
- RELATIONSHIP: words that reveal the person's relationship or marital status.\\
\rule{0pt}{0.6em}\\
Rules:\\
1. Change only words that reveal a listed attribute of a listed person. Rewrite those words with neutral wording (for example "the person", "they", "their work") or replace those words with [REDACTED].\\
2. Placeholders are not words you may change. Every placeholder listed above must appear in your output exactly as written, the same number of times and in the same place. Never replace a placeholder with [REDACTED], never put a placeholder inside the words you replace, and never join it with the words next to it: if a revealing word stands next to a placeholder, change only that word.\\
3. Copy everything else unchanged: all other words, their order, punctuation and line breaks. Do not add information and do not summarise.\\
\rule{0pt}{0.6em}\\
Return JSON: \{"text": "<the whole edited document>"\}\\
\rule{0pt}{0.6em}\\
DOCUMENT:\\
\{document\}}}
\caption{The LLM-rewrite prompt (GPT-4o-mini, temperature 0, seed 0, JSON output). Braced names are filled per document: the protected people with their mention forms and attributes, the placeholders with their counts, and the anonymised document.}
\label{fig:llmprompt}
\end{figure*}
% <<< GENERATED prompt

\section{Protocol Details}
\label{app:protocol}

\paragraph{Panel and scorer provenance.}
The TAB subset uses the human subject--attribute annotations of the SPIA benchmark
\citep{oh-etal-2026-subject}; SynthPAI's original human labels were remapped for this study,
and MedQA's subject--attribute labels were constructed using two fixed GPT-4o-mini
annotation prompts, not the editor's cue lexicon. Frozen entry registries admit one nonempty
gold value per subject--attribute entry with annotation certainty at least 3:
3{,}134 MedQA, 819 TAB, and 461 SynthPAI entries. A corpus-specific prior threshold
excludes 820, 226, and 107 entries, respectively, leaving 2{,}314, 593, and 354
eligible before the matched cohorts used in the tables. The registry and evaluation
scripts are our implementation; they are not the CPR/IPR scorer in SPIA.
The candidate composition is held fixed while its order varies. With $0\leq\post\leq1$ and $\prior<1$, the upper clipping bound is inactive. The lower floor changes the score; $\delta=.05$ is the fixed operational band. A post hoc no-document
prior using the posterior's surviving subject reference is tested in Appendix~\ref{app:x2}
(Table~\ref{tab:matchedprior}).

\begin{table*}[tbp]
\centering
\small
\begin{tabular}{p{0.26\linewidth} p{0.30\linewidth} p{0.26\linewidth} l}
\toprule
Decision & Reason & Effect & Where \\
\midrule
Score $d$ against a cue-free prior; drop entries with $\prior\geq\tau$ & credit the text only with what it adds to the
adversary's own guess & $L$ is read on entries the adversary cannot answer unaided & \S\ref{sec:eval} \\
Clip $d$ at zero; read \leakE{} and \discE{} beside $L$ & the clip credits an adversary pushed to a wrong value &
\discE{} charges misdirection as leakage & Fig.~\ref{fig:states} \\
Balance candidate orders for \attr{Sex} & with two options the prior follows a position preference & balanced prior
.50; seeded reading kept & Tab.~\ref{tab:mainseeded} \\
Report \attr{Sex} AUC & a measure no position preference can move & MedQA .89 to .73 under \sacc{} &
Tab.~\ref{tab:orders} \\
Match the subject reference across editor types & a paraphraser deletes mentions and weakens the reference & rewriting
systems compared on same-reference entries & Tab.~\ref{tab:reftypes} \\
Split TAB \attr{Sex} by subject position & the adversary reads other subjects' cues as the first-named one's & TAB
\attr{Sex} read on first-named subjects beside the aggregate & Tab.~\ref{tab:subjectpos} \\
Balance $L$ over strata and attributes; \attr{Relationship} on SynthPAI only & strata are unbalanced; the
\attr{Relationship} prior is too uncertain on MedQA and TAB & two attributes on MedQA and TAB, three on SynthPAI &
\S\ref{sec:eval} \\
Read MedQA through \attr{Sex} & the upstream tagger's \texttt{[OCCUPATION]} placeholders mostly name the clinician or no
occupation & MedQA \attr{Occupation} not interpreted & \S\ref{sec:setup} \\
Pair contrasts within one adversary; bootstrap by document; report detectable differences & documents are the sampling
unit, and TAB has 144 & TAB nulls read as statements about power & Tab.~\ref{tab:mde} \\
Fix the configuration on one development adversary; read two families after freezing & keep the held-out adversaries out
of the tuning & held-out families as a reference & Tab.~\ref{tab:transfer} \\
\unli{} on sentences free of protected cues by a detector older than the binder; TPS at the edit site & measure
collateral loss independently of the selection, and the edit site separately & content read away from and at the edits
& \S\ref{sec:eval} \\
\bottomrule
\end{tabular}
\caption{Evaluation decisions: what each choice of the protocol is for and what it changes.}
\label{tab:decisions}
\end{table*}
% >>> GENERATED cohorts
\paragraph{Cohorts and conventions.}
Privacy is scored on the entries every system shares. Under the order-balanced reading of Tables~\ref{tab:main} and~\ref{tab:contrasts} these are 2,149 on MedQA, 501 on TAB and 351 on SynthPAI; under the seeded reading of Tables~\ref{tab:mainseeded} and~\ref{tab:contrastsseeded}, 2,093 and 505 (1,165 and 140 documents). The seeded cohort also requires the published Adaptive output, which misses 63 MedQA entries; the balanced one compares Adaptive on its own same-reference entries and drops the \attr{Sex} entries whose eight orders all list the gold value on one side (7 on MedQA, 4 on TAB). Two systems are scored on other sets. The LLM rewrite is run on every MedQA document outside its prompt's development set and on a fixed sample of 91 TAB documents (1,999 and 298 scored entries), on which \sacc{} scores $L$ .635 and .431 and \unli{} .704 and .322. Adaptive is compared on the 1,008 and 487 entries on which it and \sacc{} give the adversary the same subject reference (see Rewriting systems and the subject reference, below), where \sacc{} scores $L$ .597 and .417; its MedQA \unli{} is measured on the 1,207 documents of the published output, where \sacc{} scores .703. Every other system shares \sacc{}'s reference on at least 99.9\% of entries, and the LLM rewrite on 89\% and 91\%. Paired differences are always taken on the entries both systems share. TPS is computed over the documents a system edits, on every document it was run on, and is read beside the share of characters edited; the input edits nothing, so its TPS is 1 by definition, and INTACT's edit sites relative to $D$ are the detector's placeholders, which it fills with text from the source. Table~\ref{tab:meanthree} averages each paired difference over MedQA, TAB and SynthPAI with equal weight; since it mixes MedQA's \attr{Occupation} and TAB's aggregate \attr{Sex} measures, we read it as a summary only, and every comparison we draw rests on per-corpus paired differences.
% <<< GENERATED cohorts

% Cohort and release-version audit, compiled from the frozen scoring panels and Appendix~\ref{app:history}.
\begin{table*}[tbp]
\centering
\small
\setlength{\tabcolsep}{3.2pt}
\renewcommand{\arraystretch}{1.10}
\begin{tabular*}{\textwidth}{@{\extracolsep{\fill}}p{0.19\textwidth}p{0.23\textwidth}p{0.19\textwidth}p{0.17\textwidth}p{0.15\textwidth}@{}}
\toprule
\textbf{Reading} & \textbf{Entry cohort} & \textbf{Text version} & \textbf{Adversary} & \textbf{Results} \\
\midrule
Main, balanced \attr{Sex} orders & MedQA 2,149; TAB 501; SynthPAI 351 & After both fixes & GPT-4o-mini & Tables~\ref{tab:main}, \ref{tab:contrasts} \\
Seeded answer orders & MedQA 2,093; TAB 505 & After both fixes & GPT-4o-mini & Tables~\ref{tab:mainseeded}, \ref{tab:contrastsseeded} \\
Adaptive, matched reference & MedQA 1,008; TAB 487 & After both fixes & GPT-4o-mini & Table~\ref{tab:contrasts} \\
LLM rewrite, development adversary & MedQA 1,999; TAB 298 & After both fixes & GPT-4o-mini & Table~\ref{tab:contrasts} \\
Subject-reference sensitivity & Comparison-specific; Figure~\ref{fig:multiverse}: 45 contrasts & Before binder fix & GPT-4o-mini; Llama; Qwen & Table~\ref{tab:fixedref}, Fig.~\ref{fig:multiverse} \\
Held-out family readings & Family-specific eligible entries & After both fixes & Llama; Qwen & Tables~\ref{tab:transfer}, \ref{tab:heldout} \\
Upstream and held-out rewrite checks & Comparison-specific & Before binder fix & As specified per test & Tables~\ref{tab:upstream}, \ref{tab:llmheld} \\
Configuration and reference counts & Comparison-specific & Before both fixes & As specified per test & Tables~\ref{tab:config}, \ref{tab:reftypes} \\
\bottomrule
\end{tabular*}
\caption{\textbf{Cohort and text-version provenance for the reported comparisons.} Counts denote scored subject--attribute entries, except the Figure~\ref{fig:multiverse} contrast count. Paired effects use the intersection of each system pair's entries; Adaptive and LLM rewrite use their own eligible sets. The source of the version exceptions is Appendix~\ref{app:history}.}
\label{tab:cohortprovenance}
\end{table*}

\paragraph{The MedQA input.}
MedQA has no span annotation, so its input comes from a rule-based tagger we wrote: spaCy person entities, sex and
relationship word lists, and WordNet occupation nouns \citep{honnibal-etal-2020-spacy,miller-1995-wordnet}. It replaces 1{,}096 words with \texttt{[OCCUPATION]}. Most name the clinician (\emph{physician} 507 times)
and at least a quarter no occupation at all (\emph{head}, \emph{hand}, \emph{seconds}, \emph{tender}, \emph{Tanner},
\emph{para}). \emph{Tanner} stage and \emph{para} are sex-specific clinical terms, and the tagger also masks sex nouns,
so the MedQA input has already lost some sex evidence before any editor runs. Every system receives the same $D$, so paired comparisons are unaffected, but the adversary can read the
placeholder, which bears on MedQA's \attr{Occupation} measure.

\paragraph{Registration.}
Before each run we wrote its question, design, numbered predictions and kill condition (the result that would make us drop the claim) to a dated file in the
repository, and appended the outcome afterwards, failed predictions included, without editing the predictions. These
are internal working files, not a public preregistration: the supplementary package ships each file with the date it
records, and we claim no third-party timestamp. We therefore write ``pre-specified'' rather than ``registered'' throughout, to keep the claim to what it is: the question, its numbered predictions and its kill condition were written down before the run, and the outcome was appended afterwards without editing them.

\paragraph{Adversaries.}
Llama is \nolinkurl{meta-llama/Llama-3.1-8B-Instruct} and Qwen is \nolinkurl{Qwen/Qwen3-30B-A3B-Instruct-2507}, both served
locally with vLLM \citep{kwon-etal-2023-efficient}; \mbox{GPT-4o-mini} \citep{openai-2024-gpt4omini} is the snapshot
\nolinkurl{gpt-4o-mini-2024-07-18} reported by the OpenAI chat-completions service. Every request asks for a single-letter
completion at temperature 0 with the top 20 log-probabilities at that position, renormalised over the candidate set.

\paragraph{Prompts.}
{\small\raggedright The prior prompt is \texttt{Consider "\{handle\}", a person involved in \{domain\}. No further
information is available. Question: What is their \{attr\}?} followed by the lettered options and \texttt{Answer with a
single letter. Answer:}. The posterior prompt is \texttt{Text: \{text\}} followed by \texttt{Question: Considering the
person referred to as "\{handle\}" in the text, what is their \{attr\}?}, the lettered options and the same closing
lines. The attribute is phrased as \emph{sex}, \emph{relationship status} or \emph{occupation}.\par}

\paragraph{Reading the answer.}
The distribution is the probability mass on the option letters, renormalised. For the local families, when an option
letter is missing from the top 20, the probability of every option letter is computed directly (teacher forcing);
through the API a missing letter receives no mass. Absolute scores therefore differ between the local models and the API,
and every comparison is paired within one adversary.

\begin{figure}[htbp]
\centering
\safegraphic{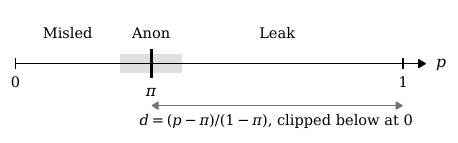}
\caption{One entry's posterior $\post$ against the adversary's prior $\prior$; the shaded band is $\prior\pm\delta$. $d$ is 0 for any posterior at or
below the prior, so $L$ scores \textsc{Misled} as it scores no change; \leakE{} counts \textsc{Leak}, \discE{} counts
\textsc{Leak} and \textsc{Misled}.}
\label{fig:states}
\end{figure}

\begin{figure*}[tbp]
\centering
\safegraphic{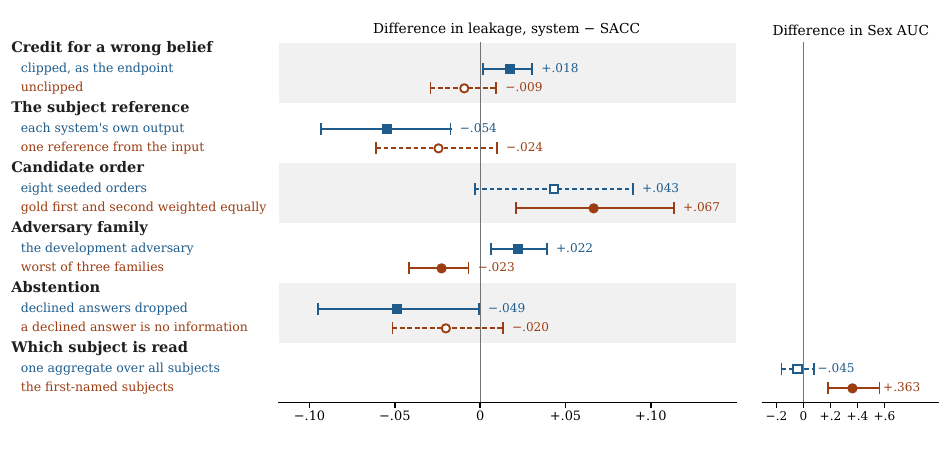}
\caption{\textbf{Illustrative effects of six protocol choices.}
Each panel compares two readings of one contrast; filled markers and
solid 99\% intervals indicate separation from zero. Examples were selected
post hoc, not sampled to estimate frequency; content proxies have no paired
verdict contrast. Sources: Tables~\ref{tab:signed} (clipped vs.\ signed),
\ref{tab:contrasts} and~\ref{tab:fixedref} (reference, adversary),
\ref{tab:contrastsseeded} (order), \ref{tab:tabref} (subject) and
\ref{tab:reasoning} (abstention). The second row of the reference panel and both adversary rows are read with
one shared subject reference, where per-adversary intervals are reported.}
\label{fig:choices}
\end{figure*}

\paragraph{Candidates, orders and eligibility.}
A check fixed in advance found the \attr{Relationship} prior too uncertain to serve as a baseline on MedQA and TAB, so
\attr{Relationship} enters the endpoint on SynthPAI only.
\attr{Sex} has two candidates, \attr{Occupation} eight drawn in equal numbers from the registry's occupation strata with
the gold value included, and \attr{Relationship} at least five. Every entry is scored under eight seeded candidate
orders, the same for every system. Entries with $\prior\geq\tau$ ($\tau=.8$, $.6$ and $.5$ for MedQA, TAB and
SynthPAI) are excluded, which removes 26.2\%, 27.6\% and 23.2\% of each corpus's protected entries. The thresholds were
frozen in the evaluation's first phase, before any editor was scored, after inspecting the eligible share at
$\tau\in\{.5,\dots,.9\}$. For \attr{Occupation} and \attr{Relationship} they are mostly entries whose gold value is
the adversary's default guess: 74\% of the MedQA occupation entries whose value is \emph{unemployed} and every SynthPAI
relationship entry whose value is \emph{single}. For \attr{Sex} the letter label carries no evidence, and the
two-option prior follows the adversary's preference for the option listed first: it is close to the share of the eight
orders that list the gold value first (.50 at four of eight, .62 at five, .87 at seven), so which \attr{Sex} entries
are excluded is set by the seeded orders, not by the value (on TAB 37\% of male and 35\% of female entries; on MedQA,
with $\tau=.8$, only entries with seven or eight such orders). The orders are the same for every system, so paired
contrasts are unaffected, but $\prior$ for \attr{Sex} measures position preference rather than a belief about sex.
Deciding eligibility from the balanced \attr{Sex} prior instead would admit every currently excluded entry on both
corpora, since that prior depends only on the letter-handle template and the position split, not on any entry's own
content, and sits within .02 of .50 for every entry we checked (post hoc; 3.2\% and 36.3\% of MedQA's and TAB's
checked entries flip, none the other way). This says nothing about those entries specifically; it says the balanced
prior rarely clears any reasonable $\tau$ for a two-valued attribute, by construction. A dedicated pass, scoring the
newly admitted entries for the five systems Table~\ref{tab:contrasts} compares (0.90 USD; not read from the stored
score files, which hold only the entries already eligible under the seeded prior), finds that adding them changes
no verdict there, so the eligibility filter decides no comparison we report.

\paragraph{A bias-invariant check, and the band.}
We also read the adversary's AUC for \attr{Sex}, the probability that it gives a male subject a higher
$\Pr(\textit{male})$ than a female one, which no shift of its bias changes. On MedQA \sacc{} lowers it as far as the
global ladder and further than global masking and SACC-mask (Table~\ref{tab:orders}). On TAB the aggregate AUC mixes two
readings (Table~\ref{tab:subjectpos}): .98 for each judgment's first-named subject and .32 for the others, whose predicted
sex follows the first-named subject's even when the adversary receives a surviving mention. Aggregated over both, the
adversary's \attr{Sex} AUC is .532 on the input and .578 under \sacc{}, so the aggregate does not show \sacc{} reducing
discrimination: raising an AUC of .32 towards chance on the subjects the adversary reads through someone else counts as
an increase. The interpretable reading is that of the first-named subjects, where \sacc{} lowers its input's $L$ by .246
$[+.161,+.333]$, \discE{} by .079, \leakE{} by .147 and the AUC from .98 to .62, a paired difference of +.363
(99\% interval $[+.182,+.563]$). The \discE{} and \leakE{}
readings do not depend on the band: at $\delta=.02$ and $.10$ every contrast of Table~\ref{tab:contrasts} that separates
at $.05$ still separates.

\paragraph{References shared within a judgment.}\label{app:tabref}
The rule-built reference names a subject by the longest cue-free mention that survives, and on TAB that mention is
often a role. In 73 of the 244 TAB \attr{Sex} entries, in 33 judgments, another subject of the same judgment receives
the same reference, 40 times ``the applicant'', so the question cannot single out the person asked about; 22 of these
entries are female, against 57 of the 244. Leaving them out raises \sacc{}'s reduction of its input's \attr{Sex} $L$
from .239 to .280 on the stored scores (seeded orders) and from .262 to .304 on the order-balanced reading; the \discE{} reduction
stays separable, and every comparison with the global editors keeps its sign (Table~\ref{tab:tabref}). The shared
references dilute the measured reduction; they do not produce it.

% >>> GENERATED tabref
\begin{table}[htbp]
\centering
\small
\setlength{\tabcolsep}{2.0pt}
\begin{adjustbox}{max width=\linewidth}
\begin{tabular}{l cc cc}
\toprule
& \multicolumn{2}{c}{All entries} & \multicolumn{2}{c}{Without shared references} \\
\cmidrule(lr){2-3} \cmidrule(lr){4-5}
System & $\Delta L$ & $\Delta$\textsc{Disc} & $\Delta L$ & $\Delta$\textsc{Disc} \\
\midrule
\multicolumn{5}{l}{\emph{Stored scores} (244 / 171 entries)} \\
Anonymised input & \textbf{+.239}\,{\scriptsize[+.155,+.330]} & \textbf{+.137}\,{\scriptsize[+.052,+.244]} & \textbf{+.280}\,{\scriptsize[+.172,+.398]} & \textbf{+.148}\,{\scriptsize[+.049,+.267]} \\
Global mask & \textbf{+.064}\,{\scriptsize[+.035,+.093]} & \textbf{+.100}\,{\scriptsize[+.016,+.197]} & \textbf{+.057}\,{\scriptsize[+.022,+.089]} & +.112\,{\scriptsize[$-$.010,+.240]} \\
Global ladder & $-$.009\,{\scriptsize[$-$.038,+.021]} & +.037\,{\scriptsize[$-$.026,+.107]} & $-$.020\,{\scriptsize[$-$.050,+.009]} & +.043\,{\scriptsize[$-$.033,+.125]} \\
SACC-mask & \textbf{+.044}\,{\scriptsize[+.018,+.075]} & +.061\,{\scriptsize[$-$.016,+.150]} & \textbf{+.040}\,{\scriptsize[+.013,+.068]} & +.061\,{\scriptsize[$-$.038,+.167]} \\
\midrule
\multicolumn{5}{l}{\emph{Order-balanced reading} (195 / 169 entries)} \\
Anonymised input & \textbf{+.262}\,{\scriptsize[+.173,+.365]} & \textbf{+.147}\,{\scriptsize[+.048,+.257]} & \textbf{+.304}\,{\scriptsize[+.196,+.429]} & \textbf{+.131}\,{\scriptsize[+.025,+.244]} \\
Global mask & \textbf{+.061}\,{\scriptsize[+.031,+.094]} & \textbf{+.101}\,{\scriptsize[+.018,+.192]} & \textbf{+.058}\,{\scriptsize[+.027,+.090]} & \textbf{+.109}\,{\scriptsize[+.008,+.215]} \\
Global ladder & $-$.017\,{\scriptsize[$-$.046,+.012]} & +.031\,{\scriptsize[$-$.041,+.107]} & $-$.022\,{\scriptsize[$-$.054,+.006]} & +.011\,{\scriptsize[$-$.065,+.090]} \\
SACC-mask & \textbf{+.052}\,{\scriptsize[+.023,+.087]} & +.059\,{\scriptsize[$-$.004,+.133]} & \textbf{+.047}\,{\scriptsize[+.021,+.076]} & +.058\,{\scriptsize[$-$.008,+.150]} \\
\bottomrule
\end{tabular}
\end{adjustbox}
\caption{\textbf{TAB \attr{Sex} with and without shared subject references.}
The 73 of 244 ambiguous entries are excluded using
the input's reference, identically across systems.
Stored and order-balanced cohorts are shown separately.
$\Delta$: system minus \sacc{}; positive means more
leakage. Brackets: 99\% intervals; bold excludes zero.
The split is post hoc.}
\label{tab:tabref}
\end{table}
% <<< GENERATED tabref

\paragraph{The order-balanced reading.}
The seeded orders are not balanced by position, and the eligibility threshold keeps entries whose gold value is listed
first in few of them, so the retained \attr{Sex} entries carry a position preference that depends on the gold value. The
main tables therefore re-read every order's probability from the request cache and weight the orders listing the gold
value first and second equally, for prior and posterior (the balanced prior is .50 for both values on every corpus).
Table~\ref{tab:orders} gives the \attr{Sex}-only measures on the entries reproduced for every system; the seeded reading
changes no verdict of the main tables but TAB's \leakE{} (Tables~\ref{tab:mainseeded} and~\ref{tab:contrastsseeded}).
Re-read scores reproduce the stored ones for 98\% to 99.5\% of MedQA entries and 90\% to 96\% of TAB's; the rest are
identical requests whose cached reply a later pass replaced, as the service's log-probabilities drift between sessions
(by more than .01 on at most 13 entries per system and corpus). A pre-specified prediction that balancing alone would make
TAB's aggregate \attr{Sex} measure informative failed; the subject-position split, made afterwards, explains why.

% >>> GENERATED orders
\begin{table*}[tbp]
\centering
\small
\setlength{\tabcolsep}{2.4pt}
\begin{adjustbox}{max width=\linewidth}
\begin{tabular}{l ccccc ccccc}
\toprule
& \multicolumn{5}{c}{MedQA (1,478)} & \multicolumn{5}{c}{TAB (193)} \\
\cmidrule(lr){2-6} \cmidrule(lr){7-11}
System & AUC & $\Delta$AUC & $\Delta L$ & $\Delta$\textsc{Disc} & $\Delta$\textsc{Leak} & AUC & $\Delta$AUC & $\Delta L$ & $\Delta$\textsc{Disc} & $\Delta$\textsc{Leak} \\
\midrule
Anonymised input & .891 & \textbf{+.158}\,{\scriptsize[+.130,+.185]} & \textbf{+.413}\,{\scriptsize[+.383,+.444]} & \textbf{+.174}\,{\scriptsize[+.145,+.203]} & \textbf{+.275}\,{\scriptsize[+.242,+.308]} & .532 & $-$.045\,{\scriptsize[$-$.163,+.076]} & \textbf{+.253}\,{\scriptsize[+.164,+.362]} & \textbf{+.150}\,{\scriptsize[+.043,+.269]} & \textbf{+.103}\,{\scriptsize[+.003,+.216]} \\
Global mask & .741 & +.007\,{\scriptsize[$-$.008,+.022]} & \textbf{+.057}\,{\scriptsize[+.042,+.072]} & \textbf{+.082}\,{\scriptsize[+.055,+.110]} & \textbf{+.045}\,{\scriptsize[+.022,+.071]} & .577 & $-$.001\,{\scriptsize[$-$.070,+.055]} & \textbf{+.062}\,{\scriptsize[+.032,+.096]} & \textbf{+.103}\,{\scriptsize[+.016,+.203]} & \textbf{+.052}\,{\scriptsize[+.012,+.102]} \\
Global ladder & .734 & \phantom{+}.000\,{\scriptsize[$-$.009,+.008]} & $-$.004\,{\scriptsize[$-$.015,+.005]} & $-$.001\,{\scriptsize[$-$.018,+.016]} & $-$.004\,{\scriptsize[$-$.017,+.010]} & .557 & $-$.021\,{\scriptsize[$-$.079,+.043]} & $-$.016\,{\scriptsize[$-$.047,+.013]} & +.032\,{\scriptsize[$-$.040,+.116]} & +.007\,{\scriptsize[$-$.034,+.060]} \\
SACC-mask & .743 & +.009\,{\scriptsize[$-$.005,+.023]} & \textbf{+.061}\,{\scriptsize[+.047,+.075]} & \textbf{+.085}\,{\scriptsize[+.059,+.113]} & \textbf{+.049}\,{\scriptsize[+.027,+.072]} & .604 & +.026\,{\scriptsize[$-$.025,+.089]} & \textbf{+.053}\,{\scriptsize[+.023,+.091]} & +.060\,{\scriptsize[$-$.002,+.143]} & \textbf{+.057}\,{\scriptsize[+.003,+.124]} \\
Global, person nouns & .730 & $-$.004\,{\scriptsize[$-$.009,\phantom{+}.000]} & \textbf{$-$.008}\,{\scriptsize[$-$.017,$-$.001]} & $-$.004\,{\scriptsize[$-$.012,+.002]} & \textbf{$-$.006}\,{\scriptsize[$-$.014,$-$.001]} & .543 &  &  &  &  \\
\rowcolor{oursrow}\sacc{} & .734 &  &  &  &  & .578 &  &  &  &  \\
\bottomrule
\end{tabular}
\end{adjustbox}
\caption{\textbf{Balanced-order \attr{Sex} evaluation}
under the development adversary. Eight candidate orders
are reweighted equally by gold-first versus gold-second
position for prior and posterior. AUC measures \attr{Sex}
discrimination; $\Delta$ is system minus \sacc{}.
Brackets: 99\% intervals; bold excludes zero
(positive $\Delta$AUC means greater discrimination).
Only entries with reproducible cached scores and both
gold-first and gold-second orders are used. The person-noun editor's TAB
paired contrasts were not computed.}
\label{tab:orders}
\end{table*}
% <<< GENERATED orders

On the main tables' cohort the re-read prior reproduces the stored one on 1{,}498 of 1{,}506 MedQA, 236 of 240 TAB and
53 of 54 SynthPAI \attr{Sex} entries; 7 and 4 entries with all orders on one side leave the \attr{Sex} cohort.

% >>> GENERATED mainseeded
\begin{table*}[tbp]
\centering
\small
\setlength{\tabcolsep}{3.4pt}
\begin{adjustbox}{max width=\linewidth}
\begin{tabular}{l cccccc cccccc}
\toprule
& \multicolumn{6}{c}{MedQA} & \multicolumn{6}{c}{TAB} \\
\cmidrule(lr){2-7} \cmidrule(lr){8-13}
System & $L\downarrow$ & \textsc{Disc}$\downarrow$ & \textsc{Leak}$\downarrow$ & $U_{\mathrm{nli}}$$\uparrow$ & TPS & Edit\% & $L\downarrow$ & \textsc{Disc}$\downarrow$ & \textsc{Leak}$\downarrow$ & $U_{\mathrm{nli}}$$\uparrow$ & TPS & Edit\% \\
\midrule
Anonymised input & .842 & .990 & .871 & \underline{.720} & 1.000 & 0.0 & .556 & .867 & .642 & .313 & 1.000 & 0.0 \\
\midrule
Global mask & .653 & .931 & .741 & .661 & .795 & 2.4 & .444 & .848 & .634 & .291 & .803 & 2.9 \\
Global ladder & \underline{.628} & \underline{.898} & \underline{.723} & .702 & .842 & 1.4 & \underline{.408} & \underline{.800} & \underline{.603} & .296 & .801 & 1.9 \\
\midrule
INTACT \citep{pilan-etal-2025-truthful} & .853 & .991 & .879 & \underline{.737} & .806 & 2.7 & .576 & .885 & .661 & .293 & .770 & 5.0 \\
Adaptive \citep{loiseau-etal-2026-adaptive}$^\dagger$ & .595 & .908 & .696 & .467 & .772 & 51.7 & .598 & .886 & .660 & .287 & .781 & 5.5 \\
\midrule
SACC-mask & .659 & .937 & .749 & .665 & .794 & 2.2 & .446 & .817 & .620 & \underline{.319} & .795 & 1.2 \\
\rowcolor{oursrow}SACC & \underline{.636} & \underline{.901} & \underline{.728} & .706 & .846 & 1.3 & \underline{.423} & \underline{.792} & \underline{.599} & \underline{.319} & .845 & 0.8 \\
\midrule
\textcolor{black!45}{LLM rewrite$^\dagger$} & \textcolor{black!45}{.609} & \textcolor{black!45}{.910} & \textcolor{black!45}{.721} & \textcolor{black!45}{.661} & \textcolor{black!45}{.802} & \textcolor{black!45}{2.8} & \textcolor{black!45}{.436} & \textcolor{black!45}{.803} & \textcolor{black!45}{.620} & \textcolor{black!45}{.297} & \textcolor{black!45}{.793} & \textcolor{black!45}{2.1} \\
\bottomrule
\end{tabular}
\end{adjustbox}
\caption{Table~\ref{tab:main} with \attr{Sex} read under the eight seeded candidate orders, as first pre-specified. Privacy is scored on the 2,093, 505 and 351 entries every unmarked system shares; $^\dagger$: scored on other entries, Adaptive on those where it gives the adversary \sacc{}'s subject reference and the LLM rewrite on its own sample (Appendix~\ref{app:protocol}). Underlined: the two best per privacy and \unli{} column among unmarked systems (TPS and Edit\% are levels); grey: the LLM rewrite, which calls a language model on every document it edits.}
\label{tab:mainseeded}
\end{table*}
% <<< GENERATED mainseeded

% >>> GENERATED contrastsseeded
\begin{table*}[tbp]
\centering
\small
\setlength{\tabcolsep}{3.6pt}
\begin{adjustbox}{max width=\linewidth}
\begin{tabular}{l ccc ccc}
\toprule
& \multicolumn{3}{c}{MedQA} & \multicolumn{3}{c}{TAB} \\
\cmidrule(lr){2-4} \cmidrule(lr){5-7}
System $-$ SACC & $\Delta L$ & $\Delta$\textsc{Disc} & $\Delta$\textsc{Leak} & $\Delta L$ & $\Delta$\textsc{Disc} & $\Delta$\textsc{Leak} \\
\midrule
Anonymised input & \textbf{+.207}\,{\scriptsize[+.192,+.222]} & \textbf{+.092}\,{\scriptsize[+.076,+.108]} & \textbf{+.144}\,{\scriptsize[+.125,+.164]} & \textbf{+.133}\,{\scriptsize[+.090,+.179]} & \textbf{+.075}\,{\scriptsize[+.033,+.122]} & +.043\,{\scriptsize[$-$.003,+.089]} \\
\midrule
Global mask & \textbf{+.018}\,{\scriptsize[+.002,+.030]} & \textbf{+.033}\,{\scriptsize[+.013,+.053]} & +.016\,{\scriptsize[$-$.002,+.032]} & +.021\,{\scriptsize[$-$.031,+.051]} & +.057\,{\scriptsize[$-$.014,+.123]} & +.036\,{\scriptsize[$-$.030,+.098]} \\
Global ladder & $-$.008\,{\scriptsize[$-$.021,+.003]} & $-$.003\,{\scriptsize[$-$.016,+.008]} & $-$.005\,{\scriptsize[$-$.018,+.005]} & $-$.015\,{\scriptsize[$-$.070,+.016]} & +.009\,{\scriptsize[$-$.051,+.054]} & +.004\,{\scriptsize[$-$.059,+.047]} \\
\midrule
INTACT \citep{pilan-etal-2025-truthful} & \textbf{+.218}\,{\scriptsize[+.197,+.238]} & \textbf{+.091}\,{\scriptsize[+.072,+.109]} & \textbf{+.151}\,{\scriptsize[+.128,+.173]} & \textbf{+.153}\,{\scriptsize[+.107,+.202]} & \textbf{+.094}\,{\scriptsize[+.038,+.164]} & \textbf{+.062}\,{\scriptsize[+.005,+.127]} \\
Adaptive \citep{loiseau-etal-2026-adaptive} & $-$.002\,{\scriptsize[$-$.047,+.044]} & +.004\,{\scriptsize[$-$.033,+.035]} & +.018\,{\scriptsize[$-$.034,+.070]} & \textbf{+.181}\,{\scriptsize[+.128,+.248]} & \textbf{+.092}\,{\scriptsize[+.038,+.160]} & \textbf{+.062}\,{\scriptsize[+.006,+.130]} \\
\midrule
SACC-mask & \textbf{+.024}\,{\scriptsize[+.012,+.035]} & \textbf{+.039}\,{\scriptsize[+.022,+.055]} & \textbf{+.024}\,{\scriptsize[+.010,+.038]} & \textbf{+.022}\,{\scriptsize[+.006,+.039]} & +.025\,{\scriptsize[$-$.014,+.072]} & +.021\,{\scriptsize[$-$.017,+.067]} \\
\midrule
\textcolor{black!45}{LLM rewrite} & \textcolor{black!45}{\textbf{$-$.026}\,{\scriptsize[$-$.052,$-$.003]}} & \textcolor{black!45}{+.013\,{\scriptsize[$-$.012,+.035]}} & \textcolor{black!45}{$-$.004\,{\scriptsize[$-$.028,+.021]}} & \textcolor{black!45}{+.005\,{\scriptsize[$-$.061,+.072]}} & \textcolor{black!45}{+.014\,{\scriptsize[$-$.064,+.085]}} & \textcolor{black!45}{+.010\,{\scriptsize[$-$.067,+.084]}} \\
\bottomrule
\end{tabular}
\end{adjustbox}
\caption{Table~\ref{tab:contrasts} with \attr{Sex} read under the eight seeded candidate orders, as first pre-specified. 99\% intervals; bold excludes zero.}
\label{tab:contrastsseeded}
\end{table*}
% <<< GENERATED contrastsseeded

% >>> GENERATED subjectpos
\begin{table*}[tbp]
\centering
\small
\setlength{\tabcolsep}{3.0pt}
\begin{adjustbox}{max width=\linewidth}
\begin{tabular}{l cccc ccc}
\toprule
& \multicolumn{4}{c}{First-named subject (157 entries, 98 \attr{Sex})} & \multicolumn{3}{c}{Other subjects (348 entries, 146 \attr{Sex})} \\
\cmidrule(lr){2-5} \cmidrule(lr){6-8}
System & AUC & $\Delta L$, \attr{Sex} & $\Delta L$ & $\Delta$\textsc{Leak} & AUC & AUC$_A$ & $\Delta L$ \\
\midrule
Anonymised input & .979\,{\scriptsize[.928,1.00]} & \textbf{+.480}\,{\scriptsize[+.305,+.651]} & \textbf{+.246}\,{\scriptsize[+.161,+.333]} & \textbf{+.147}\,{\scriptsize[+.057,+.241]} & .320 & .988 & \textbf{+.082}\,{\scriptsize[+.040,+.125]} \\
Global mask & .668\,{\scriptsize[.435,.869]} & \textbf{+.059}\,{\scriptsize[+.004,+.121]} & $-$.020\,{\scriptsize[$-$.148,+.052]} & $-$.033\,{\scriptsize[$-$.155,+.072]} & .425 & .738 & \textbf{+.037}\,{\scriptsize[+.010,+.066]} \\
Global ladder & .592\,{\scriptsize[.354,.810]} & $-$.015\,{\scriptsize[$-$.058,+.039]} & $-$.051\,{\scriptsize[$-$.182,+.015]} & $-$.034\,{\scriptsize[$-$.174,+.062]} & .375 & .707 & $-$.005\,{\scriptsize[$-$.032,+.018]} \\
SACC-mask & .666\,{\scriptsize[.475,.839]} & +.032\,{\scriptsize[$-$.006,+.068]} & +.013\,{\scriptsize[$-$.013,+.034]} & +.016\,{\scriptsize[$-$.022,+.084]} & .425 & .777 & +.020\,{\scriptsize[$-$.004,+.042]} \\
\rowcolor{oursrow}\sacc{} & .616\,{\scriptsize[.398,.812]} &  &  &  & .378 & .744 &  \\
\bottomrule
\end{tabular}
\end{adjustbox}
\caption{\textbf{TAB by subject position} under the development
adversary (stored scores, seeded orders; post hoc).
AUC measures discrimination for the queried subject;
AUC$_A$ scores other subjects' predictions against the
first-named subject's sex (61\% share that sex).
Only 14 of the 98 first-named \attr{Sex} entries are female.
$\Delta$ is system minus \sacc{} on \attr{Sex} or
the full endpoint; brackets are 99\% intervals,
and bold excludes zero.}
\label{tab:subjectpos}
\end{table*}
% <<< GENERATED subjectpos

\paragraph{Stricter thresholds.}
Score files hold only eligible entries, so a looser threshold cannot be read from them; stricter ones can, and they leave the comparisons of Table~\ref{tab:contrasts} in place (Table~\ref{tab:tausens}): \sacc{}
lowers its input's $L$ separably at every threshold, no global editor becomes separably more private, and the LLM
rewrite's MedQA lead keeps its sign, separable only from $\tau=.7$ up.

% >>> GENERATED tausens
\begin{table}[htbp]
\centering
\small
\setlength{\tabcolsep}{2.4pt}
\begin{adjustbox}{max width=\linewidth}
\begin{tabular}{ll r ccccc}
\toprule
Corpus & $\tau$ & Entries & Input & Gl. mask & Gl. ladder & SACC-mask & LLM \\
\midrule
MedQA & .5 & 1,097 & \textbf{+.207} & \textbf{+.027} & $-$.005 & \textbf{+.031} & $-$.019 \\
MedQA & .6 & 1,435 & \textbf{+.203} & \textbf{+.022} & $-$.006 & \textbf{+.027} & $-$.025 \\
MedQA & .7 & 1,871 & \textbf{+.204} & \textbf{+.018} & $-$.007 & \textbf{+.025} & \textbf{$-$.027} \\
MedQA & .8* & 2,156 & \textbf{+.207} & \textbf{+.018} & $-$.008 & \textbf{+.024} & \textbf{$-$.026} \\
TAB & .5 & 421 & \textbf{+.141} & +.021 & $-$.017 & \textbf{+.021} & +.001 \\
TAB & .6* & 505 & \textbf{+.133} & +.021 & $-$.015 & \textbf{+.022} & +.005 \\
\bottomrule
\end{tabular}
\end{adjustbox}
\caption{$\Delta L$, system minus \sacc{}, keeping only entries whose development-adversary prior is below $\tau$; * the frozen threshold. Entries: \sacc{}'s eligible entries at $\tau$, each contrast on those the system shares (at the frozen threshold every unmarked system shares 2{,}093 and 505 entries under the seeded reading and 2,149 and 501 under the order-balanced reading of Table~\ref{tab:main}). Bold estimates exclude zero (99\%). The LLM rewrite is scored on its own documents.}
\label{tab:tausens}
\end{table}

% >>> GENERATED contrasts
\begin{table*}[tbp]
\centering
\small
\setlength{\tabcolsep}{3.6pt}
\begin{adjustbox}{max width=\linewidth}
\begin{tabular}{l ccc ccc}
\toprule
& \multicolumn{3}{c}{MedQA} & \multicolumn{3}{c}{TAB} \\
\cmidrule(lr){2-4} \cmidrule(lr){5-7}
System $-$ SACC & $\Delta L$ & $\Delta$\textsc{Disc} & $\Delta$\textsc{Leak} & $\Delta L$ & $\Delta$\textsc{Disc} & $\Delta$\textsc{Leak} \\
\midrule
Anonymised input & \textbf{+.207}\,{\scriptsize[+.191,+.223]} & \textbf{+.091}\,{\scriptsize[+.075,+.107]} & \textbf{+.142}\,{\scriptsize[+.124,+.161]} & \textbf{+.145}\,{\scriptsize[+.100,+.191]} & \textbf{+.077}\,{\scriptsize[+.032,+.128]} & \textbf{+.067}\,{\scriptsize[+.021,+.113]} \\
\midrule
Global mask & \textbf{+.018}\,{\scriptsize[+.002,+.031]} & \textbf{+.035}\,{\scriptsize[+.012,+.053]} & +.017\,{\scriptsize[$-$.001,+.033]} & +.020\,{\scriptsize[$-$.033,+.050]} & +.056\,{\scriptsize[$-$.021,+.120]} & +.025\,{\scriptsize[$-$.039,+.077]} \\
Global ladder & $-$.007\,{\scriptsize[$-$.021,+.003]} & $-$.005\,{\scriptsize[$-$.020,+.007]} & $-$.006\,{\scriptsize[$-$.019,+.004]} & $-$.016\,{\scriptsize[$-$.072,+.016]} & +.010\,{\scriptsize[$-$.056,+.059]} & \phantom{+}.000\,{\scriptsize[$-$.064,+.046]} \\
\midrule
INTACT \citep{pilan-etal-2025-truthful} & \textbf{+.217}\,{\scriptsize[+.196,+.238]} & \textbf{+.089}\,{\scriptsize[+.069,+.107]} & \textbf{+.149}\,{\scriptsize[+.126,+.172]} & \textbf{+.165}\,{\scriptsize[+.117,+.214]} & \textbf{+.100}\,{\scriptsize[+.048,+.162]} & \textbf{+.086}\,{\scriptsize[+.027,+.148]} \\
Adaptive \citep{loiseau-etal-2026-adaptive} & $-$.001\,{\scriptsize[$-$.046,+.045]} & +.009\,{\scriptsize[$-$.025,+.042]} & +.020\,{\scriptsize[$-$.031,+.072]} & \textbf{+.193}\,{\scriptsize[+.139,+.260]} & \textbf{+.092}\,{\scriptsize[+.038,+.157]} & \textbf{+.084}\,{\scriptsize[+.027,+.151]} \\
\midrule
SACC-mask & \textbf{+.024}\,{\scriptsize[+.011,+.035]} & \textbf{+.040}\,{\scriptsize[+.021,+.057]} & \textbf{+.022}\,{\scriptsize[+.009,+.036]} & \textbf{+.024}\,{\scriptsize[+.007,+.042]} & +.022\,{\scriptsize[$-$.008,+.060]} & +.021\,{\scriptsize[$-$.005,+.049]} \\
\midrule
\textcolor{black!45}{LLM rewrite} & \textcolor{black!45}{\textbf{$-$.026}\,{\scriptsize[$-$.052,$-$.003]}} & \textcolor{black!45}{+.018\,{\scriptsize[$-$.007,+.041]}} & \textcolor{black!45}{+.001\,{\scriptsize[$-$.025,+.025]}} & \textcolor{black!45}{+.001\,{\scriptsize[$-$.064,+.063]}} & \textcolor{black!45}{$-$.022\,{\scriptsize[$-$.093,+.039]}} & \textcolor{black!45}{$-$.003\,{\scriptsize[$-$.073,+.066]}} \\
\bottomrule
\end{tabular}
\end{adjustbox}
\caption{System minus SACC on the entries both share (Adaptive: same-reference entries, a different cohort from the levels of Table~\ref{tab:main}, which therefore need not subtract to these differences), \attr{Sex} read under balanced orders; positive: the system leaks more. 99\% intervals; bold excludes zero; intervals covering zero can hide differences up to about .02 on MedQA, .07 for the two systems scored on their own entries, and .04 to .09 on TAB (Table~\ref{tab:mde}).}
\label{tab:contrasts}
\end{table*}
% <<< GENERATED contrasts

\paragraph{Equivalence, power and the mean of three corpora.}
A contrast whose 99\% interval covers zero is not evidence that two systems are equal. For each such contrast of
Table~\ref{tab:contrasts}, Table~\ref{tab:mde} gives the smallest difference the comparison would detect with 80\%
power and whether its 98\% interval (two one-sided tests at .01) lies within $\pm.02$. On MedQA the global ladder is
narrowly equivalent to \sacc{} on all three measures; no other null comparison is. On TAB the detectable
differences are .04 to .09, so its null comparisons are statements about power. Table~\ref{tab:meanthree} gives the
mean-of-three columns of Table~\ref{tab:main} as paired differences with intervals: averaged over the corpora, \sacc{}
is more private than global masking on \discE{} only, is not separable from the global ladder on any privacy measure,
and keeps more content than both.

\paragraph{Documents the selection rule was designed on.}
The selection rule was designed while reading a regression subset of 60 MedQA, 53 TAB and 60 SynthPAI documents and
first confirmed, in a pre-specified run, on the remaining documents. Later rounds (the rewrite table, the agreement setting
and both rounds of fixes) used every document, so the rest is not a held-out split, but nobody wrote the selection rule while
reading it. Under the development adversary, \sacc{} lowers its input's $L$ on the rest by .208 on MedQA, .114 on TAB and
.123 on SynthPAI (99\% intervals $[.190,.225]$, $[.064,.172]$, $[.038,.208]$), against .205, .161 and .105 on the
regression subset; no difference between the two is separable, the largest being TAB's $-.047$ ($[-.140,+.042]$).

% >>> GENERATED mde
\begin{table}[htbp]
\centering
\small
\setlength{\tabcolsep}{2.6pt}
\begin{adjustbox}{max width=\linewidth}
\begin{tabular}{l ccc ccc}
\toprule
& \multicolumn{3}{c}{MedQA} & \multicolumn{3}{c}{TAB} \\
\cmidrule(lr){2-4} \cmidrule(lr){5-7}
System $-$ SACC & $L$ & \textsc{Disc} & \textsc{Leak} & $L$ & \textsc{Disc} & \textsc{Leak} \\
\midrule
Global mask & sep. & sep. & +.022 & +.053 & +.093 & +.075 \\
Global ladder & \textbf{+.016} & \textbf{+.017} & \textbf{+.016} & +.056 & +.074 & +.067 \\
Adaptive$^\dagger$ & +.061 & +.045 & +.069 & sep. & sep. & sep. \\
SACC-mask & sep. & sep. & sep. & sep. & +.045 & +.036 \\
LLM rewrite$^\dagger$ & sep. & +.032 & +.033 & +.081 & +.086 & +.091 \\
\bottomrule
\end{tabular}
\end{adjustbox}
\caption{Minimum detectable difference (80\% power, two-sided $\alpha=.01$, from the bootstrap standard error) of each contrast of Table~\ref{tab:contrasts} whose interval covers zero; sep.: the interval excludes zero. Bold: shown equivalent within $\pm.02$ (98\% interval, two one-sided tests at .01).}
\label{tab:mde}
\end{table}
% <<< GENERATED mde

% >>> GENERATED meanthree
\begin{table}[htbp]
\centering
\small
\setlength{\tabcolsep}{2.4pt}
\begin{adjustbox}{max width=\linewidth}
\begin{tabular}{l cccc}
\toprule
System $-$ SACC & $\Delta L$ & $\Delta$\textsc{Disc} & $\Delta$\textsc{Leak} & $\Delta$$U_{\mathrm{nli}}$ \\
\midrule
Anonymised input & \textbf{+.156}\,{\scriptsize[+.130,+.184]} & \textbf{+.085}\,{\scriptsize[+.061,+.111]} & \textbf{+.103}\,{\scriptsize[+.077,+.130]} & +.003\,{\scriptsize[$-$.002,+.007]} \\
Global mask & +.010\,{\scriptsize[$-$.009,+.022]} & +.026\,{\scriptsize[$-$.003,+.050]} & +.010\,{\scriptsize[$-$.016,+.030]} & \textbf{$-$.024}\,{\scriptsize[$-$.029,$-$.019]} \\
Global ladder & +.003\,{\scriptsize[$-$.022,+.026]} & +.018\,{\scriptsize[$-$.012,+.045]} & +.011\,{\scriptsize[$-$.019,+.038]} & \textbf{$-$.008}\,{\scriptsize[$-$.013,$-$.004]} \\
INTACT & \textbf{+.164}\,{\scriptsize[+.136,+.193]} & \textbf{+.094}\,{\scriptsize[+.067,+.124]} & \textbf{+.112}\,{\scriptsize[+.082,+.143]} & $-$.002\,{\scriptsize[$-$.010,+.006]} \\
Adaptive$^\dagger$ & \textbf{+.092}\,{\scriptsize[+.055,+.132]} & \textbf{+.057}\,{\scriptsize[+.023,+.092]} & \textbf{+.061}\,{\scriptsize[+.024,+.098]} & --- \\
SACC-mask & \textbf{+.016}\,{\scriptsize[+.009,+.023]} & \textbf{+.021}\,{\scriptsize[+.008,+.035]} & \textbf{+.014}\,{\scriptsize[+.004,+.025]} & \textbf{$-$.014}\,{\scriptsize[$-$.017,$-$.010]} \\
LLM rewrite$^\dagger$ & $-$.018\,{\scriptsize[$-$.063,+.025]} & +.015\,{\scriptsize[$-$.028,+.055]} & $-$.011\,{\scriptsize[$-$.062,+.036]} & \textbf{$-$.070}\,{\scriptsize[$-$.083,$-$.056]} \\
\bottomrule
\end{tabular}
\end{adjustbox}
\caption{Mean over MedQA, TAB and SynthPAI of each paired difference, system minus \sacc{}, under the development adversary, \attr{Sex} read under balanced orders; the three corpora's bootstrap draws are averaged draw by draw. Privacy: positive means the system leaks more (99\% intervals); \unli{}: positive means it keeps more (95\%). $^\dagger$: on its own entries as in Table~\ref{tab:contrasts}; Adaptive's \unli{} is not paired on MedQA. Bold estimates exclude zero.}
\label{tab:meanthree}
\end{table}
% <<< GENERATED meanthree

\paragraph{Computing the measures.}
For a per-entry score $f$, which is $d$ for $L$, $\mathbf{1}[\post>\prior+\delta]$ for \leakE{} and
$\mathbf{1}[|\post-\prior|>\delta]$ for \discE{}, and a corpus whose measures cover the attributes $A_c$,
\begin{equation*}
  \bar f=\frac{1}{|A_c|}\sum_{a\in A_c}\frac{1}{|\Sigma_a|}\sum_{\sigma\in\Sigma_a}\frac{1}{|E_{a\sigma}|}
  \sum_{e\in E_{a\sigma}} f(e),
\end{equation*}
where $\Sigma_a$ are the panel's strata of attribute $a$ and $E_{a\sigma}$ the eligible entries in stratum $\sigma$.
Every stratum of an attribute weighs the same, whatever its size.

\paragraph{Intervals and non-inferiority.}
A paired difference between two systems is computed on the entries both are scored on. Each bootstrap draw resamples
documents with replacement, gives every entry the weight of its document, recomputes $\bar f$ for both systems and takes
the difference; a stratum absent from a draw drops out of that draw's mean. We use 4{,}000 draws with seed 0 and report
the percentile interval. \sacc{} is non-inferior to a system at margin $m$ when the upper end of the two-sided 99\%
interval of $L(\sacc{})-L(\text{system})$ lies below $m$; the margin is the pre-specified minimum meaningful effect,
$m=.02$.

% >>> GENERATED multiplicity
\paragraph{Multiple comparisons.}
Table~\ref{tab:contrasts} holds 42 paired comparisons. Treated as one family, with two-sided p-values from the normal approximation to each bootstrap distribution and Holm's step-down correction, 17 of the 21 differences whose 99\% interval excludes zero remain significant at the family-wise .01 level and 18 at .05. The differences that do not survive at .05 are those of the LLM rewrite on MedQA ($L$), INTACT on TAB (\leakE{}) and Adaptive on TAB (\leakE{}); at .01, those of global masking on MedQA ($L$) are also lost.
% <<< GENERATED multiplicity

\paragraph{Rewriting systems and the subject reference.}
\label{sec:paraphrase}
Adaptive paraphrases the whole document and deletes the patient's mention in most MedQA vignettes, so its adversary
receives a generic reference on about seven entries in ten, against two in ten under \sacc{}; such references lower the
posterior by .18 to .30 within each system (Table~\ref{tab:reftypes}). Its lower clipped leakage score $L$ over all MedQA
entries ($\Delta L=-.054$) is not separable once every system is scored with one reference built from the input
(Appendix~\ref{app:robust}), whereas its content score, a third below \sacc{}'s, does not depend on the reference at all.

\paragraph{Subject reference.}
The reference is built from the released text in this order: the longest gold mention of more than three characters
that survives verbatim and carries no cue, rendered as \emph{the person referred to as ``\ldots''}; otherwise a
neutralised descriptor whose content word survives, rendered as \emph{the person the text introduces as ``\ldots''};
otherwise, in a single-subject document, \emph{the person this text is mainly about}; otherwise a generic description
of a person the text cannot name without revealing a protected attribute, which is still scored. On SynthPAI the
reference is always the author of the post. Table~\ref{tab:reftypes} shows how often each reference type reaches the
adversary on MedQA, and how much a fallback reference lowers the posterior within each system; the numbers come from the
pass that scored the published Adaptive.

% >>> GENERATED reftypes
\begin{table}[htbp]
\centering
\small
\setlength{\tabcolsep}{3pt}
\begin{adjustbox}{max width=\linewidth}
\begin{tabular}{l rrr cc}
\toprule
System & Mention & Generic & Main subj. & Fallback & $\Delta\post$ \\
\midrule
SACC-mask & 1,885 & 527 & 5 & .220 & $-$.303 \\
\rowcolor{oursrow}SACC & 1,883 & 529 & 5 & .221 & $-$.291 \\
Adaptive \citep{loiseau-etal-2026-adaptive} & 714 & 1,126 & 577 & .705 & $-$.177 \\
\bottomrule
\end{tabular}
\end{adjustbox}
\caption{Subject reference handed to the adversary on MedQA, by type, over all 2,417 scored entries. $\Delta\post$: mean posterior under a fallback reference minus under a mention. From the pass that scored the published Adaptive, on the texts before both rounds (Appendix~\ref{app:history}).}
\label{tab:reftypes}
\end{table}
% <<< GENERATED reftypes

\paragraph{Band.}
Re-scoring byte-identical texts on 20{,}208 entry pairs moves the posterior by .0012 on average (95th percentile at
most .014 per corpus and family, maximum .126) and by more than $\delta$ on at most 1.8\% of a corpus's pairs, and changes
the state of 0.25\% of entries, so $\delta=.05$ lies above the scoring noise for nearly every entry.

\paragraph{Utility measures.}
For \unli{}, the allowed sentences are the source sentences of at least 40 characters that contain no protected cue
according to a fixed cue detector that predates \sacc{}'s binder (the subject's mentions and the attribute lexicon, matched in the
source sentences about the subject);
each is the hypothesis, and the premise is a 1{,}200-character window of the released text centred on the median
position of the sentence's rarest content words. The cross-encoder is DeBERTa-v3-large fine-tuned for NLI
\citep{he-etal-2023-debertav3}, and a sentence counts as retained when entailment is the argmax. For TPS, edit regions are
the character-level differences between $D$ and the released text; the information content of an original span is its
negative log-probability under BERT \citep{devlin-etal-2019-bert} with the span masked and 600 characters of context on
each side, and similarity is the clipped cosine of E5 embeddings \citep{wang-etal-2022-e5}, which replace the original
metric's sentence encoder. TPS is averaged over the documents each system edits, on every document of each corpus.
With the original encoder, paraphrase-albert-small-v2, on every MedQA document, \sacc{}'s TPS ties the global ladder's
(difference $-.001$, 99\% interval $[-.004,+.003]$) and stays above every other editor's by .10 to .16; with E5 it is
above the ladder's by .004 ($[+.002,+.005]$). A prediction pre-specified before that pass, that \sacc{} would score
highest with either encoder, therefore failed with the original one.

\paragraph{Compute.}
\sacc{} runs on a CPU with a spaCy parser \citep{honnibal-etal-2020-spacy} and makes no model call. Llama and Qwen ran on
local NVIDIA RTX A6000 cards (48\,GB), which also computed \unli{} and TPS; \mbox{GPT-4o-mini} was queried through the API, whose outputs can drift between sessions.

\section{SynthPAI}
\label{app:synthpai}
The primary measure on SynthPAI covers \attr{Sex}, \attr{Occupation} and \attr{Relationship}. \sacc{} rewrites
nothing on this corpus, so its output equals SACC-mask; the LLM rewrite is scored on its own document sample
(Table~\ref{tab:synthpai}).

% >>> GENERATED synthpai
\begin{table}[htbp]
\centering
\small
\setlength{\tabcolsep}{2.4pt}
\begin{adjustbox}{max width=\linewidth}
\begin{tabular}{l cccc ccc}
\toprule
System & $L$ & \textsc{Disc} & \textsc{Leak} & $U_{\mathrm{nli}}$ & $\Delta L$ & $\Delta$\textsc{Disc} & $\Delta$\textsc{Leak} \\
\midrule
Anonymised input & .816 & .923 & .885 & .777 & \textbf{+.112} & \textbf{+.089} & \textbf{+.101} \\
\midrule
Global mask & .699 & .835 & .783 & .777 & $-$.006 & \phantom{+}.000 & $-$.002 \\
Global ladder & .735 & .884 & .835 & .779 & +.030 & +.050 & +.050 \\
\midrule
INTACT \citep{pilan-etal-2025-truthful} & .813 & .916 & .885 & .766 & \textbf{+.108} & \textbf{+.081} & \textbf{+.100} \\
Adaptive \citep{loiseau-etal-2026-adaptive} & .790 & .905 & .864 & .713 & +.085 & +.070 & +.079 \\
\midrule
SACC-mask & .705 & .835 & .785 & .777 & \phantom{+}.000 & \phantom{+}.000 & \phantom{+}.000 \\
\rowcolor{oursrow}SACC & .705 & .835 & .785 & .777 &  &  &  \\
\midrule
\textcolor{black!45}{LLM rewrite} & \textcolor{black!45}{.659} & \textcolor{black!45}{.822} & \textcolor{black!45}{.748} & \textcolor{black!45}{.627} & \textcolor{black!45}{$-$.029} & \textcolor{black!45}{+.001} & \textcolor{black!45}{$-$.008} \\
\bottomrule
\end{tabular}
\end{adjustbox}
\caption{SynthPAI under the development adversary. $\Delta$ is system minus SACC; bold $\Delta$ estimates exclude zero; SACC is shaded.}
\label{tab:synthpai}
\end{table}
% <<< GENERATED synthpai

\section{Held-out Adversaries}
\label{app:transfer}
The frozen configuration was also read under the held-out \mbox{Llama-3.1-8B} and \mbox{Qwen3-30B-A3B}
(Table~\ref{tab:transfer}), with \sacc{} in the light agreement setting on MedQA.
\sacc{} and SACC-mask lower $L$ relative to the input under Qwen on every corpus and under Llama on MedQA and SynthPAI;
under Llama on TAB both reductions are positive but not separable, and the rewritten text raises that family's \leakE{} rate. The gain of
rewriting over masking is separable under Qwen on both corpora, absent under Llama on MedQA and reversed under Llama on
TAB (Table~\ref{tab:increment}), and global masking, scored in a separate pass, reaches \sacc{}'s leakage on most
measures (Table~\ref{tab:heldout}). We therefore report rewriting as a configuration the development adversary
selects, not one the held-out families confirm.

Table~\ref{tab:transfer} gives the reduction relative to the input, input minus system, each
paired within one adversary's scoring pass; under the held-out families \sacc{} uses the light agreement setting on
MedQA. The held-out families were first scored before the full setting was chosen on MedQA, and every later held-out
pass keeps the setting they used, so that \sacc{}'s held-out numbers stay comparable across passes. On MedQA, \sacc{} moves Llama's entries mostly into the anonymised
state and Qwen's mostly into the misled one, though \discE{} improves under both. The held-out columns of Table~\ref{tab:increment} pair SACC-mask and \sacc{} on the reported texts within one
pass of each held-out family, the pass that also scores SEL+NTR, \sacc{}'s selection with an existing gender-neutral rewriter (Appendix~\ref{app:final}).

% >>> GENERATED transfer
\begin{table}[htbp]
\centering
\small
\setlength{\tabcolsep}{2.2pt}
\begin{adjustbox}{max width=\linewidth}
\begin{tabular}{ll ccc ccc}
\toprule
& & \multicolumn{3}{c}{SACC-mask} & \multicolumn{3}{c}{SACC} \\
\cmidrule(lr){3-5} \cmidrule(lr){6-8}
Corpus & Adversary & $L$ & \textsc{Disc} & \textsc{Leak} & $L$ & \textsc{Disc} & \textsc{Leak} \\
\midrule
\multirow{3}{*}{MedQA} & GPT-4o-mini & \textbf{+.183} & \textbf{+.053} & \textbf{+.120} & \textbf{+.207} & \textbf{+.092} & \textbf{+.144} \\
 & Llama-3.1-8B & \textbf{+.160} & \textbf{+.056} & \textbf{+.071} & \textbf{+.160} & \textbf{+.052} & \textbf{+.079} \\
 & Qwen3-30B & \textbf{+.218} & \textbf{+.037} & \textbf{+.140} & \textbf{+.227} & \textbf{+.035} & \textbf{+.148} \\
\midrule
\multirow{3}{*}{TAB} & GPT-4o-mini & \textbf{+.111} & \textbf{+.050} & +.022 & \textbf{+.133} & \textbf{+.075} & +.043 \\
 & Llama-3.1-8B & +.049 & \textbf{+.075} & $-$.092 & +.028 & +.003 & \textbf{$-$.150} \\
 & Qwen3-30B & \textbf{+.064} & \textbf{+.045} & +.013 & \textbf{+.104} & \textbf{+.041} & +.019 \\
\midrule
\multirow{3}{*}{SynthPAI} & GPT-4o-mini & \textbf{+.112} & \textbf{+.089} & \textbf{+.101} & \textbf{+.112} & \textbf{+.089} & \textbf{+.101} \\
 & Llama-3.1-8B & \textbf{+.109} & \textbf{+.095} & \textbf{+.093} & \textbf{+.109} & \textbf{+.086} & \textbf{+.084} \\
 & Qwen3-30B & \textbf{+.098} & \textbf{+.086} & \textbf{+.086} & \textbf{+.098} & \textbf{+.086} & \textbf{+.086} \\
\bottomrule
\end{tabular}
\end{adjustbox}
\caption{Reduction in leakage relative to the anonymised input (input $-$ system) under three adversaries; 99\%
intervals are omitted for space, and bold excludes zero.}
\label{tab:transfer}
\end{table}

\begin{table}[htbp]
\centering
\small
\begin{adjustbox}{max width=\linewidth}
\begin{tabular}{l ccc}
\toprule
Corpus & GPT-4o-mini & Llama-3.1-8B & Qwen3-30B \\
\midrule
MedQA & \textbf{+.024} & \phantom{+}.000 & \textbf{+.009} \\
TAB & \textbf{+.022} & \textbf{$-$.023} & \textbf{+.042} \\
\bottomrule
\end{tabular}
\end{adjustbox}
\caption{Gain of rewriting over masking in $L$ (SACC-mask $-$ \sacc{}) under each adversary, paired within one scoring pass on the reported texts (light agreement setting on MedQA under the held-out families). Bold estimates exclude zero.}
\label{tab:increment}
\end{table}
% <<< GENERATED transfer

Table~\ref{tab:heldout} gives the level of every system scored on the shared entries under the two held-out adversaries,
with \sacc{} defined as in Table~\ref{tab:transfer}. The global editors were scored in a different pass of each
adversary from the other systems. Both passes use identical priors, and SACC-mask, scored in both, agrees across them to a
mean absolute posterior difference of at most .005, so the levels are shown side by side, but no paired contrast is formed
across the passes. Under the held-out adversaries \sacc{} does not keep the position it holds under the development
adversary. Global masking is at or below \sacc{} on nearly every column; the global ladder is below it on $L$ and
\leakE{} on MedQA under both families and above it on most SynthPAI columns; and SACC-mask is below it on every TAB
column under Llama but above it on $L$ and \leakE{} on MedQA under Qwen, the same dependence of the rewriting gain on the
adversary as in Table~\ref{tab:increment}. Under Llama on TAB, the input and INTACT have the
lowest \leakE{} rate of all systems: every editor raises that family's leaking rate on that corpus.

% >>> GENERATED heldout
\begin{table*}[tbp]
\centering
\small
\setlength{\tabcolsep}{3.4pt}
\begin{adjustbox}{max width=\linewidth}
\begin{tabular}{ll ccc ccc ccc}
\toprule
& & \multicolumn{3}{c}{MedQA} & \multicolumn{3}{c}{TAB} & \multicolumn{3}{c}{SynthPAI} \\
\cmidrule(lr){3-5} \cmidrule(lr){6-8} \cmidrule(lr){9-11}
Adversary & System & $L$ & \textsc{Disc} & \textsc{Leak} & $L$ & \textsc{Disc} & \textsc{Leak} & $L$ & \textsc{Disc} & \textsc{Leak} \\
\midrule
\multirow{6}{*}{Llama-3.1-8B} & Anonymised input & .665 & .979 & .753 & .362 & .867 & \underline{.522} & .769 & .943 & .941 \\
 & Global mask & \underline{.496} & \underline{.915} & \underline{.673} & \underline{.303} & \underline{.815} & .640 & \underline{.653} & \underline{.848} & \underline{.846} \\
 & Global ladder & \underline{.503} & .927 & \underline{.670} & .328 & .839 & .684 & .676 & .920 & .917 \\
 & INTACT & .687 & .968 & .769 & .363 & .899 & \underline{.529} & .762 & .933 & .933 \\
 & SACC-mask & .506 & \underline{.922} & .682 & \underline{.313} & \underline{.792} & .614 & \underline{.660} & \underline{.848} & \underline{.848} \\
\rowcolor{oursrow} & SACC & .506 & .927 & .674 & .335 & .864 & .672 & \underline{.660} & \underline{.848} & \underline{.848} \\
\midrule
\multirow{6}{*}{Qwen3-30B} & Anonymised input & .771 & .961 & .835 & .518 & .894 & .639 & .724 & .937 & .854 \\
 & Global mask & .545 & \underline{.920} & .688 & .421 & \underline{.842} & \underline{.588} & \underline{.625} & \underline{.851} & \underline{.768} \\
 & Global ladder & \underline{.531} & \underline{.924} & \underline{.677} & \underline{.380} & \underline{.825} & \underline{.571} & \underline{.616} & .868 & .784 \\
 & INTACT & .817 & .967 & .860 & .517 & .900 & .653 & .713 & .937 & .854 \\
 & SACC-mask & .553 & .925 & .696 & .454 & .848 & .625 & .626 & \underline{.851} & \underline{.768} \\
\rowcolor{oursrow} & SACC & \underline{.544} & .926 & \underline{.687} & \underline{.414} & .853 & .619 & .626 & \underline{.851} & \underline{.768} \\
\bottomrule
\end{tabular}
\end{adjustbox}
\caption{Leakage of every system scored on the shared entries under the two held-out adversaries (lower is less leakage). Underlined: the two best in each column within an adversary. Entries per corpus: 1,719 / 1,714, 478 / 529, 350 / 333 (MedQA, TAB, SynthPAI; Llama / Qwen).}
\label{tab:heldout}
\end{table*}
% <<< GENERATED heldout

Table~\ref{tab:llmheld} scores the same-input LLM rewrite under the two held-out adversaries, on its first document
samples (on MedQA the 200-document one; Table~\ref{tab:llmfull} gives the full set) and paired with \sacc{} on the
entries both share. These are the comparisons in which the LLM rewrite is not read
by its own model family. Over all entries it is not separable from \sacc{} on $L$ or \discE{} on any corpus under
either family. Under Qwen on MedQA it is more private on \leakE{} ($-.044$, 99\% interval $[-.103,-.003]$; not
tabulated) and, on the
entries where both systems give the adversary the same reference, on $L$, with an interval that only just excludes
zero; under Qwen on TAB, \sacc{} is non-inferior to it at .02 on those entries. Elsewhere the samples are too small to
decide non-inferiority either way.

% >>> GENERATED llmheld
\begin{table*}[tbp]
\centering
\small
\setlength{\tabcolsep}{3.4pt}
\begin{adjustbox}{max width=\linewidth}
\begin{tabular}{ll ccc c}
\toprule
Adversary & Corpus & $\Delta L$ & $\Delta$\textsc{Disc} & $\Delta L$, same ref. & Non-inferior at .02 \\
\midrule
\multirow{3}{*}{Llama-3.1-8B} & MedQA & $-$.020\,{\scriptsize[$-$.074,+.017]} & $-$.005\,{\scriptsize[$-$.060,+.037]} & $-$.026\,{\scriptsize[$-$.093,+.015]} & no / no \\
 & TAB & +.019\,{\scriptsize[$-$.058,+.097]} & +.008\,{\scriptsize[$-$.073,+.088]} & +.048\,{\scriptsize[$-$.032,+.128]} & no / no \\
 & SynthPAI & $-$.011\,{\scriptsize[$-$.119,+.079]} & +.065\,{\scriptsize[$-$.048,+.156]} & $-$.011\,{\scriptsize[$-$.119,+.079]} & no / no \\
\midrule
\multirow{3}{*}{Qwen3-30B} & MedQA & $-$.030\,{\scriptsize[$-$.088,+.005]} & $-$.003\,{\scriptsize[$-$.068,+.062]} & \textbf{$-$.040}\,{\scriptsize[$-$.114,\phantom{+}.000]} & no / no \\
 & TAB & +.022\,{\scriptsize[$-$.030,+.077]} & $-$.017\,{\scriptsize[$-$.067,+.036]} & +.046\,{\scriptsize[$-$.005,+.103]} & no / yes \\
 & SynthPAI & $-$.007\,{\scriptsize[$-$.113,+.079]} & +.077\,{\scriptsize[$-$.005,+.149]} & $-$.007\,{\scriptsize[$-$.113,+.079]} & no / no \\
\bottomrule
\end{tabular}
\end{adjustbox}
\caption{The same-input LLM rewrite against \sacc{} under the held-out adversaries, LLM rewrite minus \sacc{} (negative: the LLM rewrite leaks less), on its document sample. Non-inferior: whether \sacc{}'s deficit is shown to be below .02 on all entries / on same-reference entries. Brackets are 99\% intervals. Scored on the texts before the binder fix (Appendix~\ref{app:history}).}
\label{tab:llmheld}
\end{table*}
% <<< GENERATED llmheld

\paragraph{GPT-4.1-mini without chain-of-thought, post hoc.}
After the configuration was frozen, we also read \mbox{GPT-4.1-mini} as a plain (no chain-of-thought) adversary, on
the same 360-entries-per-corpus sample the reasoning adversary draws (Appendix~\ref{app:robust}),
all 8 seeded orders, five systems (anonymised input, global mask, global ladder, SACC-mask, \sacc{}). This is the
same model as that reasoning adversary, read a second way, not an additional family. On MedQA it reproduces the
other families: input $-$ \sacc{} $+.142$ \,{\scriptsize[+.098,+.189]}, separable. On TAB it does not: input $-$
\sacc{} $-.042$ \,{\scriptsize[$-$.135,+.042]}, not separable, matching Llama rather than Qwen; and SACC-mask $-$
\sacc{} $-.101$ \,{\scriptsize[$-$.165,$-$.042]}, separable and negative, reproducing Llama's rewriting-vs-masking
reversal, under a model never served locally. Two of the readings therefore show the TAB reversal, one of them
through an API; Section~\ref{sec:multiverse} and Limitations report both readings.

\section{Adversaries That Know the Rules}
\label{app:ruleaware}
\sacc{} is deterministic and its lexicon, replacement table and scope rule are published, so an adversary can know
them. Two pre-specified tests give the rules to an adversary. The first gives them to classifiers: 17 counts of traces that only the
editors' rules leave, alone and with the development adversary's score, and a TF-IDF classifier trained on the same
system's released texts, all cross-fitted by document and stacked (Table~\ref{tab:ruleaware}). The second gives them
to the development adversary: each text is preceded by a description of the de-identifier and of the editor that
produced it (Table~\ref{tab:ruletold}).

On MedQA the rules give an adversary no advantage. The rule features alone cannot tell the sexes apart in any edited text
(AUC .48 to .55) and add nothing to the development adversary on \sacc{} or SACC-mask; the classifier trained on each
system's outputs infers less than the development adversary (.62 against .72 to .74). The rewrite table leaves one trace a
rule-aware adversary could use, \emph{them} before a noun where \emph{her} was possessive: the global ladder leaves it 7.2
times per 100 female single-subject vignettes against 0.8 per 100 male, \sacc{}'s rewrite step 0.3 against 0.8, and
neither yields a measurable gain. Under the stacked adversary, input minus
\sacc{} on AUC is +.173 $[+.143,+.205]$, whereas \sacc{} minus
the global ladder is +.001 $[-.008,+.009]$.
Told how each text was edited, the development adversary gives less probability to the true sex
in every edited text and the same in the input. \sacc{}'s reduction of its input grows ($L$ .410 to .513), and it stays
ahead of global masking and SACC-mask. The global ladder, whose rules reach every person, is read least and leads
\sacc{} in the editor-aware reading. Post hoc, \sacc{} minus
the ladder is +.033 $[+.005,+.060]$ in AUC and +.075
$[+.058,+.093]$ in $L$; the AUC result is the one
pre-specified MedQA prediction of the two tests that failed.
On $L$, \sacc{} minus global mask and SACC-mask is $-$.111
$[-.133,-.089]$ and $-$.109 $[-.129,-.088]$, respectively
(post hoc).

TAB cannot answer either question: even on its input, the rule features and the development adversary rank sex below
the pre-specified validity gates, since the adversary reads sex reliably only for each judgment's first-named subject
(Section~\ref{sec:multiverse}; Appendix~\ref{app:tabref}).
The TAB input's AUC is .686 for the rule-feature classifier
and .540 for the development adversary, both below the .80
gate; with the rule-aware prompt it is .494.

% >>> GENERATED ruleaware
\begin{table*}[tbp]
\centering
\small
\setlength{\tabcolsep}{4pt}
\begin{adjustbox}{max width=\linewidth}
\begin{tabular}{l ccccc c}
\toprule
System & Development & Rule features & Both & Trained on outputs & Stacked & $\Delta$, rules \\
\midrule
Anonymised input & .869 & .863 & .885 & .847 & .887 & \textbf{+.016}\,{\scriptsize[+.002,+.030]} \\
Global mask & .735 & .480 & .723 & .616 & .733 & \textbf{$-$.012}\,{\scriptsize[$-$.021,$-$.004]} \\
Global ladder & .717 & .548 & .711 & .624 & .713 & $-$.007\,{\scriptsize[$-$.020,+.007]} \\
SACC-mask & .737 & .491 & .724 & .616 & .731 & \textbf{$-$.012}\,{\scriptsize[$-$.021,$-$.004]} \\
\rowcolor{oursrow}\sacc{} & .718 & .538 & .718 & .619 & .714 & \phantom{+}.000\,{\scriptsize[$-$.013,+.012]} \\
\bottomrule
\end{tabular}
\end{adjustbox}
\caption{\textbf{Rule-aware adversaries on MedQA \attr{Sex}}
(1,513 entries; AUC).
Development: standard adversary; Rule features: logistic
regression on 17 edit-rule counts; Both: their combination;
Trained on outputs: TF-IDF word 1--2-grams;
Stacked: all three. Fitted scores use five
out-of-fold, document-level splits.
$\Delta$, rules: Both minus Development (99\% intervals;
bold excludes zero).}
\label{tab:ruleaware}
\end{table*}

\begin{table*}[tbp]
\centering
\small
\setlength{\tabcolsep}{4pt}
\begin{adjustbox}{max width=\linewidth}
\begin{tabular}{l ccc ccc}
\toprule
& \multicolumn{3}{c}{\attr{Sex} AUC} & \multicolumn{3}{c}{\attr{Sex} $L$} \\
\cmidrule(lr){2-4} \cmidrule(lr){5-7}
System & Standard & Told & $\Delta$ & Standard & Told & $\Delta$ \\
\midrule
Anonymised input & .889 & .885 & $-$.004\,{\scriptsize[$-$.014,+.006]} & .844 & .850 & +.006\,{\scriptsize[$-$.001,+.014]} \\
Global mask & .737 & .725 & $-$.012\,{\scriptsize[$-$.037,+.016]} & .489 & .448 & \textbf{$-$.041}\,{\scriptsize[$-$.065,$-$.015]} \\
Global ladder & .719 & .667 & \textbf{$-$.052}\,{\scriptsize[$-$.084,$-$.020]} & .428 & .262 & \textbf{$-$.166}\,{\scriptsize[$-$.189,$-$.144]} \\
SACC-mask & .739 & .738 & \phantom{+}.000\,{\scriptsize[$-$.027,+.025]} & .494 & .446 & \textbf{$-$.048}\,{\scriptsize[$-$.074,$-$.021]} \\
\rowcolor{oursrow}\sacc{} & .719 & .700 & $-$.019\,{\scriptsize[$-$.053,+.014]} & .433 & .337 & \textbf{$-$.096}\,{\scriptsize[$-$.122,$-$.071]} \\
\bottomrule
\end{tabular}
\end{adjustbox}
\caption{\textbf{Editor-aware prompting on MedQA \attr{Sex}}
(1,513 entries). Told prefixes the frozen question with
the upstream de-identifier's and each editor's rules.
Standard and Told use matched documents, references,
orders and priors, rescored in one pass (standard AUCs
differ by at most .019 from stored scores in
Table~\ref{tab:ruleaware}).
$\Delta$: Told minus Standard (99\% intervals;
bold excludes zero).}
\label{tab:ruletold}
\end{table*}
% <<< GENERATED ruleaware

\section{An Adversary With No Candidates}
\label{app:freeform}
Every privacy number above comes from an adversary that chooses among candidates, so we pre-specified an open-ended
reading. Per entry the adversary receives the released text and the same subject reference as the main tables and is
asked to name the attribute in one word, or \textsc{unknown}; there are no options and no log-probabilities, and one
completion is drawn per entry and system on 400 entries per corpus sampled with a fixed seed. $L$ needs a candidate set
and is undefined here, so the measure is accuracy against the gold value under a synonym map fixed before the run,
beside the share of replies naming no value, with a no-text reading as the floor (Table~\ref{tab:freeform}).

A pre-specified validity gate asks the input to beat that floor by .15. MedQA passes and TAB does not, so only MedQA
is read, as pre-specified. There the ordering of the main tables survives an adversary with no options: both editors lower
the accuracy separably, \sacc{} is within .05 of the global ladder, and rewriting is not worse than redaction, the
three pre-specified predictions. The caveat is the size of the effect, not its direction: this adversary answers at all on
a quarter of the input's entries and on under a tenth of an edited text's, and reaches .21 accuracy on the unedited
input, so most of what the editors take away is a wrong but committed answer becoming \textsc{unknown}. It is a check
on the question format, not a second estimate of leakage, and we do not rank systems by it.

% >>> GENERATED freeform
\begin{table}[htbp]
\centering
\small
\setlength{\tabcolsep}{3pt}
\begin{adjustbox}{max width=\linewidth}
\begin{tabular}{l cc cc}
\toprule
& \multicolumn{2}{c}{MedQA} & \multicolumn{2}{c}{TAB} \\
\cmidrule(lr){2-3} \cmidrule(lr){4-5}
Reading & Accuracy & No answer & Accuracy & No answer \\
\midrule
Anonymised input & .207 & .735 & .105 & .780 \\
Global mask & .065 & .910 & .040 & .907 \\
Global ladder & .058 & .920 & .060 & .858 \\
SACC-mask & .065 & .910 & .045 & .890 \\
\rowcolor{oursrow}SACC & .062 & .915 & .055 & .870 \\
\midrule
No text & .000 & 1.000 & .000 & 1.000 \\
\bottomrule
\end{tabular}
\end{adjustbox}
\caption{An adversary given no candidates: it reads the released text and the main tables' subject reference and names the attribute in one word, or \textsc{unknown}, on 400 entries per corpus (335 and 135 documents, fixed seed). Accuracy: share matching the gold value under a synonym map fixed before the run; No answer: share naming no value. $L$ needs a candidate set and is undefined here. On MedQA, the corpus the pre-specified validity gate admits, the input exceeds \sacc{} by \textbf{+.145}\,{\scriptsize[+.098,+.195]}, \sacc{} the global ladder by +.005\,{\scriptsize[\phantom{+}.000,+.017]} and SACC-mask by $-$.003\,{\scriptsize[$-$.015,+.008]} (99\% document bootstrap).}
\label{tab:freeform}
\end{table}

\section{Two Unseen Corpora}
\label{app:unseen}
The lexicons, rules and the one tuned setting were written on the three evaluation corpora, so we ran the frozen
editor on two corpora it had not seen, each pre-specified in a dated file before its data was read: 500 encyclopaedic biographies
with self-reported profession and gender labels \citep{de-arteaga-etal-2019-bias} and 500 Wikipedia film plot
summaries whose characters carry a gender label from a knowledge base \citep{bamman-etal-2013-learning}. Both keep
\sacc{}'s released configuration unchanged: the clinical register's lexicon, the binder, the full agreement setting
and the shared keep-list, and use a generic spaCy NER as upstream de-identifier with the domain phrase
in the adversary's prompts adapted for each corpus. The summaries name their characters, so the binder
receives mention forms as the task defines them; the biographies name no one, so it does not.
Table~\ref{tab:unseen} gives the result.

The main result transfers. \sacc{} lowers its input's leakage on both corpora, and admissible rewriting stays more
private than redaction ($-.016$ and $-.037$). It edits the fewest characters of any system on both, 1.6\% against
the global ladder's 1.9\% and 2.8\%, and on the biographies, where the content measures ran, it keeps more at the
edit site than the ladder (TPS $+.013$ $[+.011,+.015]$). \attr{Occupation} does not transfer: the biographies'
profession measure moves by .004, the reach of an occupation lexicon written for two other registers. One gate
failed: the NER layer masks 25\% of the biographies' characters against 12\% of the summaries', so on that corpus
the content comparison with the input is compromised, and $U_{\mathrm{nli}}$, whose sentences come from the source,
sits at its floor for every system there.

What these corpora add is the scope trade-off of Section~\ref{sec:method}, measured where it matters. On the
three evaluation corpora nearly every annotated person is protected, so editing by scope and editing globally nearly coincide.
These corpora name many people a release would not protect, every person a biography mentions besides its subject
and every character of a summary beyond the four its metadata names, and there the global ladder reaches lower
leakage, by .030 and .089, while 14\% and 29\% of the cues it edits are ones the binder assigns to a person outside
the protected set or to no person at all. On the summaries those are 2{,}000 of 7{,}978 edited cues, almost all
other characters' kinship and person nouns (\emph{wife} 314, \emph{father} 199, \emph{mother} 145,
\emph{husband} 132, \emph{son} 131). Such a noun states a protected subject's sex by relation while describing
someone else, the channel Limitations names, and on TAB two in three of the cues the binder assigns elsewhere turn out
to be a protected person's after all, so part of the ladder's lead is an edit \sacc{} should have made and part is
content outside the release policy (next paragraph).

% >>> GENERATED unseen
\begin{table}[htbp]
\centering
\small
\setlength{\tabcolsep}{2.4pt}
\begin{adjustbox}{max width=\linewidth}
\begin{tabular}{l ccc ccc}
\toprule
& \multicolumn{3}{c}{Biographies} & \multicolumn{3}{c}{Plot summaries} \\
\cmidrule(lr){2-4} \cmidrule(lr){5-7}
System & $L$ & \textsc{Disc} & Edit\% & $L$ & \textsc{Disc} & Edit\% \\
\midrule
Anonymised input & .877 & .963 & 0.0 & .577 & .972 & 0.0 \\
Global mask & .512 & .792 & 3.3 & .481 & .923 & 5.3 \\
Global ladder & .496 & .763 & 1.9 & .382 & .797 & 2.8 \\
SACC-mask & .542 & .804 & 2.5 & .508 & .951 & 3.3 \\
\rowcolor{oursrow}\sacc{} & .526 & .782 & 1.6 & .471 & .893 & 1.6 \\
\bottomrule
\end{tabular}
\end{adjustbox}
\caption{Frozen \sacc{} on two corpora it was not built on, under the development adversary: 990 entries on 500 biographies (\attr{Sex}, \attr{Occupation}) and 1,059 on 494 plot summaries (\attr{Sex} only). Paired differences, 99\% intervals: input $-$ \sacc{} \textbf{+.351}\,{\scriptsize[+.328,+.373]} and \textbf{+.106}\,{\scriptsize[+.078,+.138]}; \sacc{} $-$ SACC-mask \textbf{$-$.016}\,{\scriptsize[$-$.029,$-$.002]} and \textbf{$-$.037}\,{\scriptsize[$-$.060,$-$.013]}; \sacc{} $-$ global ladder \textbf{+.030}\,{\scriptsize[+.015,+.048]} and \textbf{+.089}\,{\scriptsize[+.063,+.116]}. Of the ladder's edited cues, 14\% and 29\% go to a person outside the protected set or to no person, the cues \sacc{} leaves. Edit\%: share of the input's characters edited.}
\label{tab:unseen}
\end{table}
% <<< GENERATED unseen

\paragraph{What scoping keeps, and what it misses.}
Neither content measure in Section~\ref{sec:eval} captures the cues that \sacc{} leaves untouched: \unli{} excludes
all annotated subjects’ cues by construction, and TPS scores only edited spans. Table~\ref{tab:scopeutil} reports the
resulting cue population. The ``another person'' category is relatively rare in the evaluation panel (0.7\% of MedQA
lexicon matches and 8.0\% of TAB’s) but common in the plot summaries (22\%, or 2{,}000 cues). Together with non-person
uses (3.8\%, 34.0\%, and 25.5\%), this helps explain why the unseen corpora distinguish scoped from global editing more
sharply than the three evaluation corpora do.

These are the binder's classifications, not ground truth, and the two rules behind them are not equally reliable.
Judged against the source text on TAB (Table~\ref{tab:scopejudged}), the non-person rule is correct on .87 of the sites
it keeps, and within the selection it is the only rule that moves content in Table~\ref{tab:factorial}; what it moves is
$+.004$ in \unli{} on MedQA under masking, and on TAB the measure that moves is the one Section~\ref{sec:eval} reads as a
floor. The other-person rule, the only one that reads the subject list, is correct on .32: it leaves 2.1 protected
persons' sites unedited for each site it rightly keeps, which is the over-exclusion these corpora register as leakage and the
audit of Table~\ref{tab:binderaudit}
already showed on a sample. So a global editor's lower leakage here is bought partly by editing content outside the
release policy and partly by making edits \sacc{} misses, and the share of each is measured only on TAB. Turning the
other-person rule off changes no measure on the benchmarks (Table~\ref{tab:factorial}) and would remove both the
missed edits and the side channel of Limitations, at the cost of the scope the rule is for. We therefore describe it as
a release policy a deployment can turn off rather than as part of the privacy mechanism, and keep it in the reported
configuration because the task defines scope; an audit of the two unseen corpora remains to be done.

% >>> GENERATED scopeutil
\begin{table}[htbp]
\centering
\small
\setlength{\tabcolsep}{2.6pt}
\begin{adjustbox}{max width=\linewidth}
\begin{tabular}{l rrr rr}
\toprule
Corpus & Universe & Another person & Non-person & Share & Docs \\
\midrule
MedQA & 7,632 & 51 & 242 & 3.8\% & 39 \\
TAB & 2,841 & 227 & 740 & 34.0\% & 92 \\
SynthPAI & 134 & 10 & 1 & 8.2\% & 9 \\
Biographies & 1,965 & 149 & 106 & 13.0\% & 112 \\
Plot summaries & 9,124 & 2,000 & 327 & 25.5\% & 427 \\
\bottomrule
\end{tabular}
\end{adjustbox}
\caption{The cues \sacc{} leaves within the universe that a global mask edits (every lexicon match surviving the keep-list and overlap resolution), grouped by the binder's decision. These cues are not captured by the automatic content measures: \unli{} excludes every annotated subject's cues, and TPS scores only edited spans. The counts are based on the binder's classifications (not ground truth); Table~\ref{tab:scopejudged} audits them on TAB.}
\label{tab:scopeutil}
\end{table}

\begin{table}[htbp]
\centering
\small
\setlength{\tabcolsep}{3pt}
\begin{adjustbox}{max width=\linewidth}
\begin{tabular}{l rrr cc}
\toprule
TAB, sites \sacc{} leaves & Sites & Keepable & Missed & Share keep. & Missed/keep. \\
\midrule
Non-person use & 740 & 645 & 95 & .872 & 0.15 \\
Another person & 214 & 69 & 145 & .322 & 2.1 \\
\bottomrule
\end{tabular}
\end{adjustbox}
\caption{The sites of Table~\ref{tab:scopeutil} judged against the source text by GPT-4.1-mini (Table~\ref{tab:binderllm}). Keepable: the judge assigns the site to no person or to one the release does not protect, so leaving it keeps content. Missed: it assigns the site to a protected person, so leaving it is an edit not made. The biographies and plot summaries are not audited, so their scope figures rest on the binder's own labels.}
\label{tab:scopejudged}
\end{table}
% <<< GENERATED scopeutil

\section{Per-Attribute Results and the Two-Sided Reading}
\label{app:pertag}
Table~\ref{tab:pertag} gives \sacc{} against its input per attribute (input minus \sacc{}), with the
stratum-balanced share of entries in the anonymised and misled states before and after. Table~\ref{tab:pertag-sys}
compares systems with \sacc{} per attribute, Adaptive on its same-reference entries. Table~\ref{tab:states} counts
entries in each state ($\delta=.05$) for the attributes inside each corpus's primary measure. On MedQA, rewriting
changes the kind of protection: masking moves entries out of the leaking state into the misled and the anonymised states
in similar numbers, rewriting about twice as many into the anonymised state.

\paragraph{Displacement under every editor.}
As pre-specified with the readings of Appendix~\ref{app:x1}, Table~\ref{tab:displacement} gives the share of MedQA
\attr{Sex} entries in each state for every release. Every editor except INTACT leaves about a quarter of the entries
\misB{}, against .14 for the input. Of \sacc{}'s .274 reduction in the \leakE{} rate, $+.174$ $[+.157,+.190]$ moves
entries into the band and $+.101$ $[+.083,+.119]$ into the misled state, so a third of the measured reduction is a
shift towards the wrong value rather than a return to the prior. Rewriting is the better half of that trade: against SACC-mask,
\sacc{} has more \anonB{} entries ($+.080$ $[+.063,+.096]$) and fewer \misB{} ones ($-.031$ $[-.048,-.015]$). None of
this shows recoverability, a boundary fixed before the run: SWAP has a published inverse, so Section~\ref{sec:swap} can
test recovery directly, whereas an ordinary editor has no such rule, and a \misB{} share does not license that reading.

\begin{table}[htbp]
\centering
\small
\setlength{\tabcolsep}{4pt}
\begin{tabular}{@{}l ccc@{}}
\toprule
System & \textsc{Leak} & \textsc{Anon} & \textsc{Misled} \\
\midrule
Anonymised input & .851 & .009 & .140 \\
INTACT & .874 & .004 & .122 \\
Global mask & .617 & .106 & .277 \\
SACC-mask & .626 & .103 & .272 \\
Adaptive & .605 & .141 & .254 \\
LLM rewrite$^\dagger$ & .611 & .127 & .262 \\
Global ladder & .576 & .179 & .245 \\
\sacc{} & .577 & .182 & .241 \\
\midrule
SWAP (construct) & .252 & .020 & .728 \\
\bottomrule
\end{tabular}
\caption{MedQA \attr{Sex} entries by state ($\delta=.05$), stratum-balanced shares on the 1{,}473 entries the releases
share; seeded orders, development adversary. $^\dagger$: its own document sample (1{,}375 entries).}
\label{tab:displacement}
\end{table}

On TAB the misled share splits by subject position. For the 98 first-named \attr{Sex} entries the input rarely misleads
the adversary (.071), and every cue editor raises the share to .20--.26 (global mask .238, global ladder .256, SACC-mask
.202, \sacc{} .208; INTACT .071): editing has its intended effect, and about a fifth of that effect is displacement. For the 146 other
subjects the input already misleads about half the time (.491), because the adversary answers from the first-named
person's cues, and the editors lower the share to .34--.37 (INTACT .503). Pooling the two, as the aggregate TAB measure
does, averages an editing effect against a misattribution artefact (Appendix~\ref{app:tabref}).

\paragraph{\attr{Sex}-only held-out check (post hoc).}
We re-aggregate the stored per-entry panels on \attr{Sex} entries for each adversary family.
Entry cohorts differ by adversary; Table~\ref{tab:main} uses balanced shared cohorts.
\begin{table}[htbp]
\centering
\small
\setlength{\tabcolsep}{3pt}
\begin{adjustbox}{max width=\linewidth}
\begin{tabular}{@{}lcc@{}}
\toprule
Adversary & MedQA: input $-$ \sacc{} & TAB: input $-$ \sacc{} \\
\midrule
GPT-4o-mini & $+.3801$ [+.3508,+.4094] & $+.2465$ [+.1705,+.3298] \\
Llama-3.1-8B & $+.2834$ [+.2408,+.3264] & $+.0708$ [$-$.0233,+.1766] \\
Qwen3-30B-A3B & $+.4236$ [+.3954,+.4546] & $+.2140$ [+.1545,+.2791] \\
\bottomrule
\end{tabular}
\end{adjustbox}
\caption{\textbf{Interpretable \attr{Sex}-only contrasts (post hoc).}
Per-family input-minus-\sacc{} $L$ differences and paired 99\% intervals from stored panels.
Family-specific entry cohorts; Table~\ref{tab:main} uses balanced shared cohorts.}
\label{tab:heldoutsex}
\end{table}
On TAB, the rewriting-minus-masking contrast under Llama is $-.0082$ and non-separable on \attr{Sex}; the
sign reversal of the composite endpoint is concentrated in \attr{Occupation}. MedQA \attr{Sex}-only input reductions
remain separable for all three adversaries; Llama's TAB input comparison has an interval crossing zero.

\paragraph{What still leaks on MedQA.}
Of the 1{,}513 MedQA \attr{Sex} entries, 874 remain in the leaking state under \sacc{} and the development adversary.
Their texts keep an unedited lexicon cue in 4\% of cases (mostly kinship nouns of relatives the binder sent to another
person), against 2\% of anonymised entries. A sex-specific clinical term (pregnancy, menstruation, gynaecological or
urological anatomy; a list fixed before counting) appears in 32\% of them, against 13\% of anonymised and 23\% of
misled entries. The upstream placeholder \texttt{[SEX]}, which marks that a word was removed but not its value, is
present in nearly all of them (95\%, against 90\% of anonymised entries). Most of the remaining leakage is therefore
read from clinical content that no list of attribute cues covers, which a lexical editor cannot reach.

\paragraph{MedQA read through \attr{Sex} alone.}
MedQA's \attr{Occupation} measure inherits a defective upstream tagger (Appendix~\ref{app:protocol}) and the frozen
endpoint averages it with \attr{Sex}. Table~\ref{tab:medqasex} re-reads every main-table comparison on \attr{Sex}
alone. Five of the seven keep both their sign and their separability. The two that change are the rewriting systems,
both towards \sacc{}: Adaptive's lead of .054 becomes a deficit of .051, and the same-input LLM rewrite's .026 becomes
.039, each separable in both readings. Narrowing the endpoint after the fact would therefore improve \sacc{}'s
standing, which is why the definition frozen before any editor was scored is kept and both readings are reported.

% >>> GENERATED medqasex
\begin{table}[htbp]
\centering
\small
\setlength{\tabcolsep}{3pt}
\begin{adjustbox}{max width=\linewidth}
\begin{tabular}{l cc}
\toprule
System $-$ \sacc{} on MedQA & $\Delta L$, reported & $\Delta L$, \attr{Sex} only \\
\midrule
Anonymised input & \textbf{+.207}\,{\scriptsize[+.192,+.222]} & \textbf{+.410}\,{\scriptsize[+.381,+.440]} \\
Global mask & \textbf{+.018}\,{\scriptsize[+.002,+.030]} & \textbf{+.056}\,{\scriptsize[+.040,+.071]} \\
Global ladder & $-$.008\,{\scriptsize[$-$.021,+.003]} & $-$.006\,{\scriptsize[$-$.016,+.004]} \\
INTACT & \textbf{+.218}\,{\scriptsize[+.197,+.238]} & \textbf{+.438}\,{\scriptsize[+.407,+.468]} \\
Adaptive$\ast$ & \textbf{$-$.054}\,{\scriptsize[$-$.093,$-$.017]} & \textbf{+.051}\,{\scriptsize[+.025,+.078]} \\
SACC-mask & \textbf{+.024}\,{\scriptsize[+.012,+.035]} & \textbf{+.061}\,{\scriptsize[+.046,+.075]} \\
LLM rewrite$\ast$ & \textbf{$-$.026}\,{\scriptsize[$-$.052,$-$.003]} & \textbf{+.039}\,{\scriptsize[+.021,+.058]} \\
\bottomrule
\end{tabular}
\end{adjustbox}
\caption{MedQA's endpoint as reported (\attr{Sex} and \attr{Occupation}) and read as \attr{Sex} alone, the attribute its upstream tagger leaves interpretable. \sacc{} scores $L$ .635 and .433 on the two readings (2,156 and 1,513 entries). 99\% intervals, bold excludes zero; positive means the system leaks more. $\ast$: the only two comparisons whose sign changes, both rewriting systems and both towards \sacc{}, which is why the frozen endpoint is kept as it was fixed rather than narrowed after the fact.}
\label{tab:medqasex}
\end{table}
% <<< GENERATED medqasex

\paragraph{By gold value.}
The measures cover entries the adversary cannot answer from its prior, mostly values other than its default guess.
Table~\ref{tab:goldvalue} splits the \attr{Sex} entries by gold value. Every editor lowers $L$ for both values on both
corpora. After editing, female entries are misled more often than male ones on MedQA (.30 against .17 under \sacc{}):
with the cue gone, the adversary falls back on \emph{male}, which \discE{} counts as disclosure. On TAB the adversary is
already misled on two thirds of the 57 female entries of the input, and most male entries still leak after every editor.

% >>> GENERATED goldvalue
\begin{table}[htbp]
\centering
\small
\setlength{\tabcolsep}{2.4pt}
\begin{adjustbox}{max width=\linewidth}
\begin{tabular}{ll ccccc}
\toprule
Corpus & Gold value & Input & Gl.\ mask & Gl.\ ladder & SACC-mask & \sacc{} \\
\midrule
MedQA & female (700) & .808 / .181 & .482 / .343 & .439 / .304 & .485 / .339 & .444 / .301 \\
MedQA & male (813) & .879 / .097 & .496 / .204 & .416 / .178 & .503 / .199 & .423 / .173 \\
TAB & female (57) & .281 / .667 & .175 / .614 & .171 / .544 & .176 / .561 & .150 / .544 \\
TAB & male (187) & .826 / .080 & .581 / .059 & .440 / .091 & .541 / .070 & .478 / .064 \\
\bottomrule
\end{tabular}
\end{adjustbox}
\caption{\attr{Sex} entries by gold value under the development adversary: $L$ / share of entries in the misled state ($\delta=.05$), on the entries every listed system shares.}
\label{tab:goldvalue}
\end{table}
% <<< GENERATED goldvalue

% >>> GENERATED pertag
\begin{table}[htbp]
\centering
\small
\setlength{\tabcolsep}{2.0pt}
\begin{adjustbox}{max width=\linewidth}
\begin{tabular}{ll r ccc cc}
\toprule
Corpus & Attr. & $n$ & $\Delta L$ & $\Delta$\textsc{Disc} & $\Delta$\textsc{Leak} & \textsc{Anon} & \textsc{Misled} \\
\midrule
MedQA & Sex & 1473 & \textbf{+.408} & \textbf{+.174} & \textbf{+.274} & .009$\to$.182 & .140$\to$.241 \\
MedQA & Occ & 620 & +.004 & +.005 & +.011 & .011$\to$.016 & .101$\to$.107 \\
TAB & Sex & 244 & \textbf{+.239} & \textbf{+.137} & +.067 & .037$\to$.174 & .373$\to$.304 \\
TAB & Occ & 261 & \textbf{+.027} & +.013 & +.020 & .230$\to$.243 & .075$\to$.081 \\
SynthPAI & Sex & 54 & \textbf{+.050} & \phantom{+}.000 & +.033 & .013$\to$.013 & .080$\to$.113 \\
SynthPAI & Occ & 233 & \textbf{+.009} & \phantom{+}.000 & +.002 & .009$\to$.009 & .034$\to$.036 \\
SynthPAI & Rel & 64 & \textbf{+.275} & \textbf{+.266} & \textbf{+.266} & .208$\to$.474 & .000$\to$.000 \\
\bottomrule
\end{tabular}
\end{adjustbox}
\caption{\sacc{} against the anonymised input, per attribute: $\Delta$ is input minus \sacc{} on $L$,
\discE{} and \leakE{} (99\% intervals omitted for space; bold excludes zero); \textsc{Anon}/\textsc{Misled}
is the stratum-balanced share in that state ($\delta=.05$), input $\to$ \sacc{}.}
\label{tab:pertag}
\end{table}

\begin{table}[htbp]
\centering
\small
\setlength{\tabcolsep}{3pt}
\begin{adjustbox}{max width=\linewidth}
\begin{tabular}{ll cc}
\toprule
Corpus & System & \textsc{Sex} $\Delta L$ / $\Delta$\textsc{Disc} & \textsc{Occ} $\Delta L$ / $\Delta$\textsc{Disc} \\
\midrule
MedQA & Adaptive \citep{loiseau-etal-2026-adaptive} & +.028 / +.023 & $-$.033 / $-$.016 \\
MedQA & LLM rewrite & \textbf{+.039} / \textbf{+.056} & \textbf{$-$.092} / $-$.031 \\
MedQA & Global ladder & $-$.006 / +.003 & $-$.010 / $-$.009 \\
MedQA & Global mask & \textbf{+.056} / \textbf{+.081} & \textbf{$-$.020} / $-$.014 \\
TAB & Adaptive \citep{loiseau-etal-2026-adaptive} & \textbf{+.269} / \textbf{+.145} & \textbf{+.093} / +.039 \\
TAB & LLM rewrite & +.097 / +.081 & \textbf{$-$.087} / $-$.054 \\
TAB & Global ladder & $-$.009 / +.037 & $-$.021 / $-$.019 \\
TAB & Global mask & \textbf{+.064} / \textbf{+.100} & $-$.023 / +.013 \\
\bottomrule
\end{tabular}
\end{adjustbox}
\caption{System minus \sacc{} on $L$ / \discE{}, per attribute (companion to Table~\ref{tab:pertag}'s input $-$
\sacc{}); 99\% intervals are omitted for space, and bold estimates exclude zero.}
\label{tab:pertag-sys}
\end{table}

\begin{table}[htbp]
\centering
\small
\begin{adjustbox}{max width=\linewidth}
\begin{tabular}{ll rrr}
\toprule
Corpus (entries) & System & \textsc{Anon} & \textsc{Leak} & \textsc{Misled} \\
\midrule
\multirow{6}{*}{MedQA (2093)} & Input & 23 & 1785 & 285 \\
 & Global mask & 172 & 1420 & 501 \\
 & Global ladder & 285 & 1376 & 432 \\
 & INTACT & 13 & 1828 & 252 \\
 & SACC-mask & 164 & 1436 & 493 \\
 & SACC & 290 & 1377 & 426 \\
\midrule
\multirow{6}{*}{TAB (505)} & Input & 60 & 366 & 79 \\
 & Global mask & 66 & 364 & 75 \\
 & Global ladder & 77 & 354 & 74 \\
 & INTACT & 56 & 372 & 77 \\
 & SACC-mask & 87 & 345 & 73 \\
 & SACC & 90 & 343 & 72 \\
\midrule
\multirow{6}{*}{SynthPAI (351)} & Input & 19 & 325 & 7 \\
 & Global mask & 32 & 309 & 10 \\
 & Global ladder & 27 & 316 & 8 \\
 & INTACT & 19 & 325 & 7 \\
 & SACC-mask & 32 & 310 & 9 \\
 & SACC & 32 & 310 & 9 \\
\bottomrule
\end{tabular}
\end{adjustbox}
\caption{Entries in each state ($\delta=.05$) under the development adversary, for the attributes in each
corpus's primary measure (counts, not shares; corpus totals in the row label).}
\label{tab:states}
\end{table}
% <<< GENERATED pertag

\paragraph{Unclipped scale.}
Table~\ref{tab:signed} sets $L$ beside the mean of the unclipped residual $(\post-\prior)/(1-\prior)$,
balanced in the same way. The unclipped scale credits a misled adversary more than the clip does, because a posterior
below the prior lowers the mean instead of counting as zero. Three readings change between the scales; in each the clip
creates a separation in \sacc{}'s favour that the unclipped scale does not show: on MedQA, global masking and SACC-mask, both less private than \sacc{} on
$L$, are not separable from it on the unclipped scale, and on TAB neither is SACC-mask's deficit on $L$. On MedQA both systems leave more entries misled than
\sacc{} (Table~\ref{tab:states}), and on \discE{}, which charges misdirection, both are less private than \sacc{}
(Table~\ref{tab:contrasts}). The two-sided reading of the main text is the one that separates these cases.

% >>> GENERATED signed
\begin{table*}[tbp]
\centering
\small
\setlength{\tabcolsep}{3.4pt}
\begin{adjustbox}{max width=\linewidth}
\begin{tabular}{l cc cc cc}
\toprule
& \multicolumn{2}{c}{MedQA} & \multicolumn{2}{c}{TAB} & \multicolumn{2}{c}{SynthPAI} \\
\cmidrule(lr){2-3} \cmidrule(lr){4-5} \cmidrule(lr){6-7}
System $-$ SACC & $\Delta L$ & $\Delta d$ & $\Delta L$ & $\Delta d$ & $\Delta L$ & $\Delta d$ \\
\midrule
Anonymised input & \textbf{+.207}\,{\scriptsize[+.192,+.222]} & \textbf{+.220}\,{\scriptsize[+.194,+.246]} & \textbf{+.133}\,{\scriptsize[+.090,+.179]} & \textbf{+.086}\,{\scriptsize[+.029,+.147]} & \textbf{+.112}\,{\scriptsize[+.052,+.179]} & \textbf{+.116}\,{\scriptsize[+.053,+.191]} \\
\midrule
Global mask & \textbf{+.018}\,{\scriptsize[+.002,+.030]} & $-$.009\,{\scriptsize[$-$.029,+.009]} & +.021\,{\scriptsize[$-$.031,+.051]} & +.007\,{\scriptsize[$-$.043,+.038]} & $-$.006\,{\scriptsize[$-$.019,\phantom{+}.000]} & $-$.006\,{\scriptsize[$-$.022,+.002]} \\
Global ladder & $-$.008\,{\scriptsize[$-$.021,+.003]} & $-$.007\,{\scriptsize[$-$.021,+.005]} & $-$.015\,{\scriptsize[$-$.070,+.016]} & $-$.013\,{\scriptsize[$-$.072,+.023]} & +.030\,{\scriptsize[$-$.026,+.090]} & +.031\,{\scriptsize[$-$.025,+.090]} \\
\midrule
INTACT \citep{pilan-etal-2025-truthful} & \textbf{+.218}\,{\scriptsize[+.197,+.238]} & \textbf{+.239}\,{\scriptsize[+.206,+.272]} & \textbf{+.153}\,{\scriptsize[+.107,+.202]} & \textbf{+.099}\,{\scriptsize[+.036,+.165]} & \textbf{+.108}\,{\scriptsize[+.048,+.176]} & \textbf{+.114}\,{\scriptsize[+.050,+.190]} \\
Adaptive \citep{loiseau-etal-2026-adaptive} & $-$.002\,{\scriptsize[$-$.047,+.044]} & +.013\,{\scriptsize[$-$.047,+.075]} & \textbf{+.181}\,{\scriptsize[+.128,+.248]} & \textbf{+.128}\,{\scriptsize[+.056,+.211]} & +.085\,{\scriptsize[$-$.002,+.172]} & +.089\,{\scriptsize[$-$.002,+.179]} \\
\midrule
SACC-mask & \textbf{+.024}\,{\scriptsize[+.012,+.035]} & $-$.003\,{\scriptsize[$-$.020,+.014]} & \textbf{+.022}\,{\scriptsize[+.006,+.039]} & +.015\,{\scriptsize[$-$.003,+.034]} & \phantom{+}.000\,{\scriptsize[\phantom{+}.000,\phantom{+}.000]} & \phantom{+}.000\,{\scriptsize[\phantom{+}.000,\phantom{+}.000]} \\
\midrule
\textcolor{black!45}{LLM rewrite} & \textcolor{black!45}{\textbf{$-$.026}\,{\scriptsize[$-$.052,$-$.003]}} & \textcolor{black!45}{\textbf{$-$.048}\,{\scriptsize[$-$.077,$-$.020]}} & \textcolor{black!45}{+.005\,{\scriptsize[$-$.061,+.072]}} & \textcolor{black!45}{+.004\,{\scriptsize[$-$.076,+.086]}} & \textcolor{black!45}{$-$.029\,{\scriptsize[$-$.153,+.082]}} & \textcolor{black!45}{$-$.035\,{\scriptsize[$-$.168,+.088]}} \\
\bottomrule
\end{tabular}
\end{adjustbox}
\caption{Clipped ($L$) and unclipped ($d$) leakage, system minus SACC; positive means the system leaks more. Brackets are 99\% intervals; bold estimates exclude zero.}
\label{tab:signed}
\end{table*}
% <<< GENERATED signed

% >>> GENERATED docwin
\paragraph{Documents.}
Averaging the clipped leakage score $L$ over each document's entries, \sacc{} lowers the score of 1,048 of the 1,096 MedQA documents whose score moves (of 1,165); 113 of the 132 TAB documents whose score moves (of 140); 50 of the 63 SynthPAI documents whose score moves (of 246), relative to the input it receives.
% <<< GENERATED docwin

\section{Lexical Coverage}
\label{app:coverage}
Table~\ref{tab:coverage} splits each corpus's entries by whether the input holds at least one cue of that attribute
that \sacc{} selects for that subject, that is, a cue in its lexicon whose binding status for the subject is direct,
relational or unresolved, and gives \sacc{}'s gain over its input on each part. Almost every \attr{Sex} entry on
MedQA and TAB has such a cue, and the gain is concentrated there. Fewer than half of the \attr{Occupation} entries
have one, and even where one exists the gain is small on MedQA, whose \attr{Occupation} measure is uninterpretable because of the upstream tagger's
placeholders (Section~\ref{sec:setup}). On SynthPAI, few \attr{Sex} entries have a cue at all, and the \attr{Relationship} gain comes
entirely from entries with one. Entries without a cue gain nothing separable from zero anywhere.

% >>> GENERATED coverage
\begin{table}[htbp]
\centering
\small
\setlength{\tabcolsep}{2.4pt}
\begin{adjustbox}{max width=\linewidth}
\begin{tabular}{ll rc cc}
\toprule
Corpus & Attr. & Entries & With cue & $\Delta L$, cue & $\Delta L$, no cue \\
\midrule
MedQA & Sex & 1,513 & .962 & \textbf{+.426}\,{\scriptsize[+.396,+.457]} & \phantom{+}.000\,{\scriptsize[\phantom{+}.000,\phantom{+}.000]} \\
MedQA & Occ & 643 & .429 & +.005\,{\scriptsize[$-$.015,+.030]} & +.003\,{\scriptsize[$-$.007,+.016]} \\
\midrule
TAB & Sex & 244 & .963 & \textbf{+.248}\,{\scriptsize[+.163,+.344]} & --- \\
TAB & Occ & 261 & .441 & \textbf{+.055}\,{\scriptsize[+.019,+.110]} & +.006\,{\scriptsize[$-$.003,+.016]} \\
\midrule
SynthPAI & Sex & 54 & .074 & --- & +.001\,{\scriptsize[\phantom{+}.000,+.004]} \\
SynthPAI & Occ & 233 & .228 & \textbf{+.008}\,{\scriptsize[\phantom{+}.000,+.023]} & +.009\,{\scriptsize[\phantom{+}.000,+.040]} \\
SynthPAI & Rel & 64 & .469 & \textbf{+.441}\,{\scriptsize[+.193,+.683]} & +.001\,{\scriptsize[\phantom{+}.000,+.003]} \\
\bottomrule
\end{tabular}
\end{adjustbox}
\caption{Share of entries whose input holds a cue \sacc{} selects for that subject and attribute, and \sacc{}'s gain over its input (input minus \sacc{}) on entries with and without one. Brackets are 99\% intervals; bold estimates exclude zero; --- marks fewer than ten entries.}
\label{tab:coverage}
\end{table}
% <<< GENERATED coverage

\section{Utility Comparisons}
\label{app:utility}

\begin{figure}[htbp]
\centering
\safegraphic{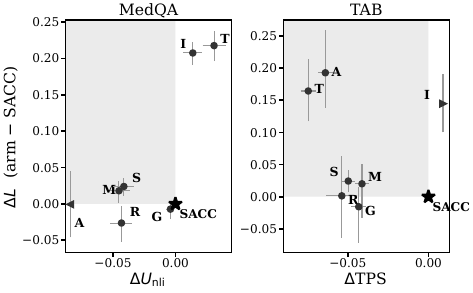}
\caption{\textbf{Paired privacy--utility differences relative to \sacc{} (star).}
Horizontal: content (95\% for \unli{}, 99\% for TPS); vertical: $L$
(99\% intervals), over shared entries with balanced
\attr{Sex} orders. Shading marks worse privacy and utility.
Content is \unli{} on MedQA and TPS on TAB, where
\unli{} is near its floor; axes differ across panels.
I: input; M: global mask; G: global ladder;
T: INTACT; A: Adaptive (same-reference entries);
S: SACC-mask; R: LLM rewrite. Arrows denote values
outside the scale (Adaptive on MedQA: $-.237$;
input TPS minus \sacc{}: $+.155$).}
\label{fig:paired}
\end{figure}

Table~\ref{tab:utility-app} gives the paired utility differences behind Table~\ref{tab:main}, \sacc{} minus each system.
On MedQA and TAB, \sacc{} scores higher than every other editor on both measures. The exceptions are the systems that
remove no attribute evidence, the input and INTACT, which keep more allowed content on MedQA by \unli{}, and SACC-mask on
TAB, which \unli{} does not separate from \sacc{}; there \unli{} scores the masked text slightly above the input. On SynthPAI, where \sacc{} rewrites nothing, its differences from the
other editors are small, and the two whole-document rewriters, Adaptive and the LLM rewrite, keep less by \unli{}. The
LLM rewrite and the published Adaptive on MedQA are compared on the documents each was run on.

\paragraph{The clinical task.}
Each MedQA vignette ends with a four-option exam question, whose options and key we take from the public MedQA release.
GPT-4o-mini answers it on each system's text, in four rotations of the options (Table~\ref{tab:downstream}). It answers
73.0\% of the questions on \sacc{}'s text and on the anonymised input, and 74.1\% on the source text: the
de-identifier costs about a point, and no editor changes the rate measurably on top of it, including on the 288 vignettes
that name a sex-specific clinical term. All three predictions pre-specified before the run held. From the question and its options alone, without the vignette, it answers 42.8\%, so the test depends on the vignette. On the 35 vignettes without a sex-specific term whose answer on the source text changes, in at least two of the four rotations, when the sex of the people in it is swapped, it answers 40.0\% after \sacc{} and 45.0\% on the input (difference $-.050$, 95\% interval $[-.121,+.014]$); both predictions pre-specified for these two checks held, but the subset is too small to separate the editors from one another.

% >>> GENERATED downstream
\begin{table}[htbp]
\centering
\small
\setlength{\tabcolsep}{2.6pt}
\begin{adjustbox}{max width=\linewidth}
\begin{tabular}{l c ccc}
\toprule
& & \multicolumn{3}{c}{System $-$ \sacc{}} \\
\cmidrule(lr){3-5}
Text & Accuracy & All & Sex-specific term (288) & Sex-dependent (35) \\
\midrule
Source text & .741 & \textbf{+.011}\,{\scriptsize[+.003,+.019]} & +.013\,{\scriptsize[$-$.004,+.031]} & +.043\,{\scriptsize[$-$.064,+.157]} \\
Anonymised input & .730 & \phantom{+}.000\,{\scriptsize[$-$.006,+.005]} & +.009\,{\scriptsize[$-$.003,+.021]} & +.050\,{\scriptsize[$-$.014,+.121]} \\
Global mask & .728 & $-$.002\,{\scriptsize[$-$.009,+.005]} & +.006\,{\scriptsize[$-$.008,+.021]} & +.014\,{\scriptsize[$-$.086,+.129]} \\
Global ladder & .729 & $-$.001\,{\scriptsize[$-$.005,+.002]} & +.003\,{\scriptsize[$-$.005,+.009]} & +.021\,{\scriptsize[$-$.021,+.071]} \\
SACC-mask & .727 & $-$.003\,{\scriptsize[$-$.009,+.004]} & +.004\,{\scriptsize[$-$.010,+.017]} & +.036\,{\scriptsize[$-$.064,+.143]} \\
\rowcolor{oursrow}SACC & .730 &  &  &  \\
LLM rewrite & .721 & $-$.006\,{\scriptsize[$-$.013,\phantom{+}.000]} & $-$.003\,{\scriptsize[$-$.016,+.010]} & $-$.039\,{\scriptsize[$-$.117,+.039]} \\
\midrule
Question only & .428 & --- & --- & --- \\
\bottomrule
\end{tabular}
\end{adjustbox}
\caption{MedQA exam questions answered on each system's text by GPT-4o-mini (four option rotations per text; the LLM rewrite on its 1{,}189 documents). Sex-specific term: vignettes that name a sex-specific clinical term. Sex-dependent: vignettes without such a term whose answer on the source text changes, in at least two of the four rotations, when the sex of the people in it is swapped. Question only: the exam question and its options without the vignette. Brackets are 95\% document-bootstrap intervals; bold estimates exclude zero, paired within a system.}
\label{tab:downstream}
\end{table}

% >>> GENERATED utilityapp
\begin{table*}[tbp]
\centering
\small
\setlength{\tabcolsep}{3.4pt}
\begin{adjustbox}{max width=\linewidth}
\begin{tabular}{l cc cc cc}
\toprule
& \multicolumn{2}{c}{MedQA} & \multicolumn{2}{c}{TAB} & \multicolumn{2}{c}{SynthPAI} \\
\cmidrule(lr){2-3} \cmidrule(lr){4-5} \cmidrule(lr){6-7}
System & $U_{\mathrm{nli}}$ & TPS & $U_{\mathrm{nli}}$ & TPS & $U_{\mathrm{nli}}$ & TPS \\
\midrule
Anonymised input & \textbf{$-$.014}\,{\scriptsize[$-$.021,$-$.006]} & --- & +.005\,{\scriptsize[$-$.003,+.015]} & --- & \phantom{+}.000\,{\scriptsize[$-$.006,+.006]} & --- \\
\midrule
Global mask & \textbf{+.045}\,{\scriptsize[+.037,+.054]} & \textbf{+.052}\,{\scriptsize[+.050,+.054]} & \textbf{+.028}\,{\scriptsize[+.017,+.040]} & \textbf{+.041}\,{\scriptsize[+.037,+.045]} & \phantom{+}.000\,{\scriptsize[$-$.001,\phantom{+}.000]} & \phantom{+}.000\,{\scriptsize[\phantom{+}.000,+.002]} \\
Global ladder & \textbf{+.004}\,{\scriptsize[+.001,+.007]} & \textbf{+.004}\,{\scriptsize[+.002,+.005]} & \textbf{+.023}\,{\scriptsize[+.011,+.035]} & \textbf{+.043}\,{\scriptsize[+.038,+.048]} & $-$.002\,{\scriptsize[$-$.008,+.003]} & \textbf{+.009}\,{\scriptsize[+.001,+.017]} \\
\midrule
INTACT \citep{pilan-etal-2025-truthful} & \textbf{$-$.031}\,{\scriptsize[$-$.040,$-$.022]} & \textbf{+.040}\,{\scriptsize[+.038,+.043]} & \textbf{+.026}\,{\scriptsize[+.010,+.042]} & \textbf{+.074}\,{\scriptsize[+.070,+.079]} & +.011\,{\scriptsize[$-$.005,+.029]} & $-$.008\,{\scriptsize[$-$.017,+.002]} \\
Adaptive \citep{loiseau-etal-2026-adaptive} & \textbf{+.237}\,{\scriptsize[+.220,+.253]} & \textbf{+.074}\,{\scriptsize[+.072,+.076]} & \textbf{+.032}\,{\scriptsize[+.014,+.051]} & \textbf{+.064}\,{\scriptsize[+.059,+.068]} & \textbf{+.064}\,{\scriptsize[+.025,+.104]} & $-$.005\,{\scriptsize[$-$.018,+.005]} \\
\midrule
SACC-mask & \textbf{+.041}\,{\scriptsize[+.033,+.050]} & \textbf{+.052}\,{\scriptsize[+.050,+.054]} & \phantom{+}.000\,{\scriptsize[$-$.007,+.007]} & \textbf{+.050}\,{\scriptsize[+.045,+.054]} & \phantom{+}.000\,{\scriptsize[\phantom{+}.000,\phantom{+}.000]} & \phantom{+}.000\,{\scriptsize[\phantom{+}.000,\phantom{+}.000]} \\
\midrule
\textcolor{black!45}{LLM rewrite} & \textcolor{black!45}{\textbf{+.043}\,{\scriptsize[+.035,+.052]}} & \textcolor{black!45}{\textbf{+.043}\,{\scriptsize[+.040,+.045]}} & \textcolor{black!45}{\textbf{+.025}\,{\scriptsize[+.011,+.040]}} & \textcolor{black!45}{\textbf{+.054}\,{\scriptsize[+.044,+.063]}} & \textcolor{black!45}{\textbf{+.141}\,{\scriptsize[+.104,+.177]}} & \textcolor{black!45}{$-$.002\,{\scriptsize[$-$.011,+.006]}} \\
\bottomrule
\end{tabular}
\end{adjustbox}
\caption{Utility differences, SACC minus system; positive means SACC keeps more. \unli{} with 95\% and TPS with 99\% intervals; bold estimates exclude zero. The input's TPS is 1 by definition.}
\label{tab:utility-app}
\end{table*}
% <<< GENERATED utilityapp

\section{Robustness of the Evaluation}
\label{app:robust}
Three checks test whether the results depend on how they were measured, each pre-specified before it was scored. The
first two ran on the texts before the binder fix of Appendix~\ref{app:history}, which changes 114 MedQA and 24 TAB
documents; the reasoning adversary ran on the fixed texts.

\paragraph{One reference for every system.}
Table~\ref{tab:fixedref} gives every entry one reference, built once from the anonymised input and given to the adversary
for every system's text, which removes a paraphraser's weaker-reference advantage. \sacc{} still lowers its input's
leakage on every corpus and measure under the development adversary and Qwen. On MedQA Adaptive's lead ($-.054$ with
output-built references) shrinks to $-.024$, not separable; this reading predates the binder fix. Re-scoring just the
114 changed documents on the current text confirms that the version gap is immaterial: \sacc{}'s own MedQA $L$ moves by $-.007$, and
Adaptive's shrunk lead moves from $-.024$ to $-.017$, remaining not separable and reversing no verdict (0.09 USD; the
reference match, not the version difference, drives the $-.054\to-.024$ shift). On TAB Adaptive still leaks more than \sacc{}, and the LLM
rewrite is not separable from \sacc{} on any corpus. Under Llama \sacc{}'s TAB gain over its input is not separable from zero, as in
Table~\ref{tab:transfer}.

\paragraph{Other upstream anonymisers.}
Table~\ref{tab:upstream} applies \sacc{} on TAB after INTACT (span filling) and the published Adaptive (paraphrase), with
the light rewrite setting. Under the development adversary and Qwen \sacc{} lowers $L$ of both outputs and rewriting adds
to masking; under the development adversary \discE{} does not separate, so part of the gain moves the adversary to a wrong
value. Under Llama \sacc{} lowers $L$ and \discE{} after INTACT and only \discE{} after Adaptive, so the pre-specified
prediction that it lowers $L$ after both under both held-out families fails for Adaptive under Llama.

\paragraph{A reasoning adversary.}
GPT-4.1-mini, reasoning in at most five sentences before its answer letter, was run on 360 sampled entries per corpus with
two candidate orders (2.24 USD; Table~\ref{tab:reasoning}). It often declines to answer on edited text: on a fifth of
\sacc{}'s MedQA entries and a quarter of its TAB entries, against 2\% and 7\% of the input's. Dropping declined answers,
as pre-specified, \sacc{} lowers its input's MedQA $L$ ($+.095$ $[.044,.147]$), not separably on TAB, and the global ladder
is more private on MedQA ($-.049$ $[-.095,-.001]$), against the pre-specified prediction. Counting a declined answer as no
information, a reading added after the run and before any contrast was read, \sacc{} lowers its input's MedQA $L$ by .250
$[.204,.296]$, no editor is separably more private, and the TAB reduction is not separable from zero. Under neither reading is \sacc{}
separably more private than SACC-mask, although rewriting makes this adversary decline more often. The gap between the
two abstention readings is the largest single-choice effect in our study, which is why Section~\ref{sec:recs} asks for
both.

% >>> GENERATED reasoning
\begin{table*}[tbp]
\centering
\small
\setlength{\tabcolsep}{3.0pt}
\begin{adjustbox}{max width=\linewidth}
\begin{tabular}{l ccc ccc}
\toprule
& \multicolumn{3}{c}{MedQA} & \multicolumn{3}{c}{TAB} \\
\cmidrule(lr){2-4} \cmidrule(lr){5-7}
System & $\Delta L$, answers only & $\Delta L$, abstention $=\prior$ & Abstains & $\Delta L$, answers only & $\Delta L$, abstention $=\prior$ & Abstains \\
\midrule
Anonymised input & \textbf{+.095}\,{\scriptsize[+.044,+.147]} & \textbf{+.250}\,{\scriptsize[+.204,+.296]} & .019 & +.057\,{\scriptsize[$-$.030,+.173]} & +.090\,{\scriptsize[$-$.001,+.193]} & .068 \\
Global mask & $-$.053\,{\scriptsize[$-$.128,+.006]} & +.006\,{\scriptsize[$-$.063,+.055]} & .170 & $-$.052\,{\scriptsize[$-$.162,+.023]} & $-$.040\,{\scriptsize[$-$.130,+.012]} & .157 \\
Global ladder & \textbf{$-$.049}\,{\scriptsize[$-$.095,$-$.001]} & $-$.020\,{\scriptsize[$-$.051,+.013]} & .216 & +.039\,{\scriptsize[$-$.063,+.152]} & $-$.002\,{\scriptsize[$-$.072,+.058]} & .264 \\
SACC-mask & +.003\,{\scriptsize[$-$.049,+.051]} & +.026\,{\scriptsize[$-$.010,+.061]} & .170 & $-$.024\,{\scriptsize[$-$.071,+.018]} & $-$.005\,{\scriptsize[$-$.050,+.051]} & .154 \\
LLM rewrite & $-$.061\,{\scriptsize[$-$.138,+.004]} & $-$.020\,{\scriptsize[$-$.087,+.032]} & .199 & --- & --- & --- \\
\midrule
\rowcolor{oursrow}SACC & & & .217 & & & .252 \\
\bottomrule
\end{tabular}
\end{adjustbox}
\caption{A reasoning adversary, GPT-4.1-mini answering after a short chain of thought, on 360 sampled entries per corpus (two candidate orders): system minus \sacc{} on $L$ (positive: the system leaks more; 99\% intervals), first as pre-specified, dropping the orders in which the adversary declines to answer, then counting a declined answer as no information. Abstains: share of the adversary's replies that decline.}
\label{tab:reasoning}
\end{table*}
% <<< GENERATED reasoning

% >>> GENERATED fixedref
\begin{table*}[tbp]
\centering
\small
\setlength{\tabcolsep}{2.6pt}
\begin{adjustbox}{max width=\linewidth}
\begin{tabular}{ll cc cc cc}
\toprule
& & \multicolumn{2}{c}{Input} & \multicolumn{2}{c}{Adaptive} & \multicolumn{2}{c}{LLM rewrite} \\
\cmidrule(lr){3-4} \cmidrule(lr){5-6} \cmidrule(lr){7-8}
Adversary & Corpus & $\Delta L$ & $\Delta$\textsc{Disc} & $\Delta L$ & $\Delta$\textsc{Disc} & $\Delta L$ & $\Delta$\textsc{Disc} \\
\midrule
\multirow{3}{*}{GPT-4o-mini} & MedQA & \textbf{+.200}\,{\scriptsize[+.184,+.216]} & \textbf{+.087}\,{\scriptsize[+.070,+.104]} & $-$.024\,{\scriptsize[$-$.061,+.010]} & $-$.018\,{\scriptsize[$-$.050,+.009]} & $-$.001\,{\scriptsize[$-$.061,+.038]} & +.020\,{\scriptsize[$-$.044,+.068]} \\
 & TAB & \textbf{+.136}\,{\scriptsize[+.097,+.180]} & \textbf{+.061}\,{\scriptsize[+.026,+.102]} & \textbf{+.188}\,{\scriptsize[+.135,+.251]} & \textbf{+.084}\,{\scriptsize[+.037,+.146]} & +.049\,{\scriptsize[$-$.014,+.107]} & +.018\,{\scriptsize[$-$.057,+.093]} \\
 & SynthPAI & \textbf{+.112}\,{\scriptsize[+.052,+.179]} & \textbf{+.089}\,{\scriptsize[+.036,+.143]} & +.085\,{\scriptsize[$-$.002,+.172]} & +.070\,{\scriptsize[$-$.004,+.141]} & $-$.029\,{\scriptsize[$-$.153,+.082]} & +.001\,{\scriptsize[$-$.139,+.120]} \\
\midrule
\multirow{3}{*}{Llama-3.1-8B} & MedQA & \textbf{+.149}\,{\scriptsize[+.126,+.173]} & \textbf{+.044}\,{\scriptsize[+.024,+.068]} & +.037\,{\scriptsize[\phantom{+}.000,+.074]} & $-$.015\,{\scriptsize[$-$.047,+.018]} & $-$.018\,{\scriptsize[$-$.071,+.018]} & $-$.009\,{\scriptsize[$-$.069,+.052]} \\
 & TAB & +.047\,{\scriptsize[$-$.002,+.100]} & +.014\,{\scriptsize[$-$.048,+.074]} & \textbf{+.087}\,{\scriptsize[+.025,+.159]} & \textbf{+.065}\,{\scriptsize[+.015,+.136]} & +.036\,{\scriptsize[$-$.041,+.112]} & +.024\,{\scriptsize[$-$.069,+.122]} \\
 & SynthPAI & \textbf{+.108}\,{\scriptsize[+.046,+.161]} & \textbf{+.091}\,{\scriptsize[+.021,+.149]} & +.079\,{\scriptsize[$-$.001,+.150]} & +.066\,{\scriptsize[$-$.017,+.141]} & $-$.011\,{\scriptsize[$-$.119,+.079]} & +.065\,{\scriptsize[$-$.048,+.156]} \\
\midrule
\multirow{3}{*}{Qwen3-30B} & MedQA & \textbf{+.221}\,{\scriptsize[+.204,+.240]} & \textbf{+.039}\,{\scriptsize[+.021,+.059]} & +.032\,{\scriptsize[$-$.003,+.066]} & $-$.018\,{\scriptsize[$-$.050,+.015]} & $-$.025\,{\scriptsize[$-$.081,+.008]} & $-$.028\,{\scriptsize[$-$.092,+.035]} \\
 & TAB & \textbf{+.122}\,{\scriptsize[+.090,+.157]} & \textbf{+.043}\,{\scriptsize[+.014,+.075]} & \textbf{+.191}\,{\scriptsize[+.143,+.255]} & +.038\,{\scriptsize[$-$.034,+.109]} & +.045\,{\scriptsize[$-$.004,+.100]} & +.002\,{\scriptsize[$-$.045,+.052]} \\
 & SynthPAI & \textbf{+.098}\,{\scriptsize[+.040,+.151]} & \textbf{+.086}\,{\scriptsize[+.016,+.145]} & +.092\,{\scriptsize[$-$.058,+.221]} & +.052\,{\scriptsize[$-$.024,+.124]} & $-$.010\,{\scriptsize[$-$.113,+.078]} & +.077\,{\scriptsize[$-$.005,+.149]} \\
\bottomrule
\end{tabular}
\end{adjustbox}
\caption{Every system scored with one subject reference, built from the anonymised input and the same for all systems: system minus \sacc{} (SynthPAI: minus SACC-mask, which \sacc{} equals there). Positive: the system leaks more. Brackets are 99\% intervals; bold estimates exclude zero. Scored on the texts before the binder fix (Appendix~\ref{app:history}).}
\label{tab:fixedref}
\end{table*}
% <<< GENERATED fixedref

% >>> GENERATED upstream
\begin{table*}[tbp]
\centering
\small
\setlength{\tabcolsep}{3.0pt}
\begin{adjustbox}{max width=\linewidth}
\begin{tabular}{ll ccc c}
\toprule
& & \multicolumn{3}{c}{Upstream output minus \sacc{}} & SACC-mask minus \sacc{} \\
\cmidrule(lr){3-5} \cmidrule(lr){6-6}
Adversary & Upstream & $\Delta L$ & $\Delta$\textsc{Disc} & $\Delta$\textsc{Leak} & $\Delta L$ \\
\midrule
\multirow{2}{*}{GPT-4o-mini} & INTACT & \textbf{+.093}\,{\scriptsize[+.055,+.136]} & +.018\,{\scriptsize[$-$.010,+.053]} & \textbf{+.040}\,{\scriptsize[+.002,+.081]} & \textbf{+.017}\,{\scriptsize[+.002,+.033]} \\
 & Adaptive & \textbf{+.111}\,{\scriptsize[+.070,+.156]} & +.009\,{\scriptsize[$-$.019,+.042]} & +.025\,{\scriptsize[$-$.010,+.065]} & \textbf{+.028}\,{\scriptsize[+.001,+.053]} \\
\midrule
\multirow{2}{*}{Llama-3.1-8B} & INTACT & \textbf{+.080}\,{\scriptsize[+.024,+.141]} & \textbf{+.057}\,{\scriptsize[+.007,+.111]} & $-$.043\,{\scriptsize[$-$.109,+.020]} & $-$.006\,{\scriptsize[$-$.028,+.013]} \\
 & Adaptive & +.025\,{\scriptsize[$-$.017,+.081]} & \textbf{+.069}\,{\scriptsize[+.024,+.125]} & \textbf{$-$.094}\,{\scriptsize[$-$.179,$-$.017]} & \textbf{+.015}\,{\scriptsize[\phantom{+}.000,+.032]} \\
\midrule
\multirow{2}{*}{Qwen3-30B} & INTACT & \textbf{+.094}\,{\scriptsize[+.061,+.129]} & \textbf{+.035}\,{\scriptsize[\phantom{+}.000,+.086]} & +.033\,{\scriptsize[$-$.021,+.098]} & \textbf{+.047}\,{\scriptsize[+.027,+.068]} \\
 & Adaptive & \textbf{+.124}\,{\scriptsize[+.087,+.163]} & +.029\,{\scriptsize[$-$.028,+.073]} & +.035\,{\scriptsize[$-$.028,+.093]} & \textbf{+.037}\,{\scriptsize[+.016,+.061]} \\
\bottomrule
\end{tabular}
\end{adjustbox}
\caption{\sacc{} applied after two other upstream anonymisers on TAB: its gain over the text it receives, and the gain of rewriting over masking. Positive: \sacc{} leaks less. Brackets are 99\% intervals; bold estimates exclude zero. Scored on the texts before the binder fix (Appendix~\ref{app:history}).}
\label{tab:upstream}
\end{table*}
% <<< GENERATED upstream

\section{Configuration and Further Baselines}
\label{app:final}
The agreement setting was chosen among three candidates (no repair, light, full) by five-fold cross-fitting over
documents on the development adversary's $L$; all five folds chose full on MedQA and light on TAB (SynthPAI has nothing to
rewrite). Table~\ref{tab:config} compares the settings: with the light setting everywhere no privacy measure changes and
the content scores do not fall; the full setting improves surface quality on MedQA, removing almost all agreement errors
and leftover redactions, but is less private on TAB, which is why the light setting is released there. With the light setting
\sacc{}'s MedQA \unli{} is .003 below the ladder's ($[-.008,+.002]$), so that lead comes from the full repair.

% >>> GENERATED finalization
\begin{table}[htbp]
\centering
\small
\setlength{\tabcolsep}{2.2pt}
\begin{adjustbox}{max width=\linewidth}
\begin{tabular}{ll ccccc rr}
\toprule
Corpus & Setting & $L$ & \textsc{Disc} & \textsc{Leak} & $U_{\mathrm{nli}}$ & TPS & Redactions & Agr.\ errors \\
\midrule
\multirow{3}{*}{MedQA} & light everywhere & .644 & .911 & .739 & .708 & .857 & 1720 & 812 \\
 & full everywhere & .642 & .904 & .735 & .710 & .847 & 40 & 1 \\
 & released & .642 & .904 & .735 & .710 & .847 & 40 & 1 \\
\midrule
\multirow{3}{*}{TAB} & light everywhere & .420 & .807 & .587 & .321 & .843 & 421 & 8 \\
 & full everywhere & .426 & \textbf{.834} & \textbf{.605} & .319 & .845 & 175 & 0 \\
 & released & .420 & .807 & .587 & .321 & .843 & 421 & 8 \\
\midrule
\multirow{3}{*}{SynthPAI} & light everywhere & .705 & .835 & .785 & .777 & .790 & 120 & 0 \\
 & full everywhere & .705 & .835 & .785 & .777 & .790 & 120 & 0 \\
 & released & .705 & .835 & .785 & .777 & .790 & 120 & 0 \\
\bottomrule
\end{tabular}
\end{adjustbox}
\caption{Agreement setting, on the texts before the two rewrite-step fixes and the binder fix of Appendix~\ref{app:history}, which together move $L$ by up to .007 (Tables~\ref{tab:fixeffect} and~\ref{tab:fix3effect}); the released rows therefore differ slightly from Table~\ref{tab:main}. Redactions: redactions left in place; Agr.\ errors: agreement errors found by the dependency check. Bold: differs from the released configuration.}
\label{tab:config}
\end{table}
% <<< GENERATED finalization

On all 1{,}189 MedQA documents outside its prompt's development set (0.19 USD; Table~\ref{tab:llmfull}), the same-input
LLM rewrite is more private than \sacc{} on $L$ under all three families, through \attr{Occupation}; on \attr{Sex} the
development adversary finds it less private, and no family separates the two on \discE{}. \sacc{} keeps more content
(\unli{} $+.043$ $[.035,.052]$; TPS $+.043$ $[.040,.045]$) while editing 1.3\% of characters against 2.8\%. Of the
pre-specified predictions, that the $L$ difference would not be separable failed and that \sacc{} would keep more held.

% >>> GENERATED llmfull
\begin{table*}[tbp]
\centering
\small
\setlength{\tabcolsep}{3.0pt}
\begin{adjustbox}{max width=\linewidth}
\begin{tabular}{l r ccccc c}
\toprule
Adversary & Entries & $\Delta L$ & $\Delta$\textsc{Disc} & $\Delta L$, same ref. & $\Delta L$, \attr{Sex} & $\Delta L$, \attr{Occ} & Non-inferior at .02 \\
\midrule
GPT-4o-mini & 1,999 & \textbf{$-$.026}\,{\scriptsize[$-$.052,$-$.003]} & +.013\,{\scriptsize[$-$.012,+.035]} & $-$.012\,{\scriptsize[$-$.037,+.012]} & \textbf{+.039}\,{\scriptsize[+.021,+.058]} & \textbf{$-$.092}\,{\scriptsize[$-$.138,$-$.048]} & no / no \\
Llama-3.1-8B & 1,573 & \textbf{$-$.045}\,{\scriptsize[$-$.071,$-$.020]} & $-$.002\,{\scriptsize[$-$.028,+.025]} & \textbf{$-$.025}\,{\scriptsize[$-$.053,$-$.003]} & $-$.004\,{\scriptsize[$-$.026,+.017]} & \textbf{$-$.085}\,{\scriptsize[$-$.131,$-$.043]} & no / no \\
Qwen3-30B & 1,597 & \textbf{$-$.038}\,{\scriptsize[$-$.062,$-$.013]} & $-$.017\,{\scriptsize[$-$.041,+.008]} & $-$.017\,{\scriptsize[$-$.041,+.004]} & +.007\,{\scriptsize[$-$.009,+.023]} & \textbf{$-$.083}\,{\scriptsize[$-$.128,$-$.041]} & no / no \\
\bottomrule
\end{tabular}
\end{adjustbox}
\caption{The same-input LLM rewrite on every MedQA document outside its prompt-development set, against \sacc{}: LLM rewrite minus \sacc{} (negative: the LLM rewrite leaks less). Non-inferior: whether \sacc{}'s deficit is shown to be below .02 on all entries / on same-reference entries. Brackets are 99\% intervals.}
\label{tab:llmfull}
\end{table*}
% <<< GENERATED llmfull

\paragraph{The redaction token.}
Part of rewriting's MedQA \unli{} gain is the NLI model's response to the token \texttt{[REDACTED]} itself: with a
differently spelled placeholder at the same sites, the gain shrinks to about 60\% of its size, while privacy does not move.

\paragraph{An existing gender-neutral rewriter.}
To separate \sacc{}'s selection from its own rewrite step, we replaced its rewrite table and agreement repair by the
rule-based NeuTral Rewriter \citep{vanmassenhove-etal-2021-neutral}, run unchanged with its own grammar correction on the
pronouns \sacc{} selects; every other selected cue stays redacted, since that system rewrites no sex-marked noun (SEL+NTR,
Table~\ref{tab:ntr}). It keeps most of rewriting's gain over masking under the development adversary ($L$: .021 of .024
on MedQA, .022 of .023 on TAB), is never separably more private than \sacc{} under any adversary, and under the
development adversary is less private on MedQA's sex entries, where \sacc{} also rewrites nouns. Its content scores are close (MedQA \unli{} .702 against .706, TPS .852 against .846),
so the gain lies in what is selected and admitted rather than in the particular rewriter.

% >>> GENERATED ntr
\begin{table*}[tbp]
\centering
\small
\setlength{\tabcolsep}{3.0pt}
\begin{adjustbox}{max width=\linewidth}
\begin{tabular}{ll ccc c}
\toprule
& & \multicolumn{3}{c}{SEL+NTR $-$ \sacc{}} & SACC-mask $-$ SEL+NTR \\
\cmidrule(lr){3-5} \cmidrule(lr){6-6}
Adversary & Corpus & $\Delta L$ & $\Delta$\textsc{Disc} & $\Delta$\textsc{Leak} & $\Delta L$ \\
\midrule
\multirow{2}{*}{GPT-4o-mini} & MedQA & +.003\,{\scriptsize[$-$.006,+.011]} & +.013\,{\scriptsize[$-$.002,+.027]} & \textbf{+.010}\,{\scriptsize[+.001,+.018]} & \textbf{+.021}\,{\scriptsize[+.011,+.030]} \\
 & TAB & +.001\,{\scriptsize[$-$.007,+.008]} & +.014\,{\scriptsize[$-$.010,+.041]} & +.004\,{\scriptsize[$-$.011,+.021]} & \textbf{+.022}\,{\scriptsize[+.004,+.041]} \\
\midrule
\multirow{2}{*}{Llama-3.1-8B} & MedQA & +.005\,{\scriptsize[$-$.002,+.011]} & $-$.004\,{\scriptsize[$-$.024,+.018]} & $-$.006\,{\scriptsize[$-$.023,+.012]} & $-$.005\,{\scriptsize[$-$.013,+.004]} \\
 & TAB & $-$.001\,{\scriptsize[$-$.007,+.005]} & $-$.022\,{\scriptsize[$-$.066,+.017]} & $-$.007\,{\scriptsize[$-$.043,+.025]} & \textbf{$-$.022}\,{\scriptsize[$-$.042,$-$.005]} \\
\midrule
\multirow{2}{*}{Qwen3-30B} & MedQA & \phantom{+}.000\,{\scriptsize[$-$.005,+.005]} & $-$.002\,{\scriptsize[$-$.016,+.012]} & $-$.004\,{\scriptsize[$-$.015,+.006]} & \textbf{+.009}\,{\scriptsize[+.003,+.017]} \\
 & TAB & \phantom{+}.000\,{\scriptsize[$-$.010,+.009]} & $-$.003\,{\scriptsize[$-$.055,+.035]} & $-$.007\,{\scriptsize[$-$.061,+.023]} & \textbf{+.042}\,{\scriptsize[+.024,+.061]} \\
\bottomrule
\end{tabular}
\end{adjustbox}
\caption{\sacc{}'s selection with an existing gender-neutral rewriter (SEL+NTR: the rule-based NeuTral Rewriter of \citealp{vanmassenhove-etal-2021-neutral} on the pronouns \sacc{} selects, every other selected cue redacted), scored in one pass per adversary on every document. Positive: the first system leaks more. Brackets are 99\% intervals; bold estimates exclude zero.}
\label{tab:ntr}
\end{table*}
% <<< GENERATED ntr

\paragraph{A lexicon from another register.}
The lexicons are the only register-specific component, and MedQA's adds clinical sex nouns and role nouns to the
generic third-person lists. To see how much of the MedQA result rests on them, we rebuilt \sacc{} on MedQA with TAB's
lexicon in both candidate pools and with the agreement setting chosen on the other corpora, leaving the binder, scope
rule and rewrite table unchanged (SACC-LODO, Table~\ref{tab:lodo}); with MedQA's own lexicon the same code reproduces
the reported texts byte for byte. The generic lexicon changes the edits on five documents only: the clinical sex nouns
are already masked in the anonymised input, and the binder leaves most clinical role nouns unedited. Under the
development adversary SACC-LODO lowers its input's $L$ by .205 [.188, .222], against .207 for \sacc{}, and it is not
separably less private than \sacc{} under any adversary, contrary to our pre-specified expectation that the clinical nouns
would matter. The setting chosen on MedQA improves content retention rather than privacy: \unli{} is .007 higher [.003, .012] and TPS
.010 lower than with the other corpora's setting (on the fixed texts; Table~\ref{tab:config}, on the texts before the
fixes, shows a .002 difference). This holds out the lexicon and the setting only; the binder rules and
rewrite table were developed on all three corpora.

% >>> GENERATED lodo
\begin{table}[tbp]
\centering
\small
\setlength{\tabcolsep}{3.0pt}
\begin{adjustbox}{max width=\linewidth}
\begin{tabular}{l r ccc c}
\toprule
& & \multicolumn{3}{c}{SACC-LODO $-$ \sacc{}} & SACC-mask \\
\cmidrule(lr){3-5} \cmidrule(lr){6-6}
Adversary & Entries & $\Delta L$ & $\Delta$\textsc{Disc} & $\Delta$\textsc{Leak} & $\Delta L$ \\
\midrule
GPT-4o-mini & 2,156 & +.002\,{\scriptsize[$-$.008,+.010]} & +.010 & +.007 & \phantom{+}.000 \\
Llama-3.1-8B & 1,719 & \phantom{+}.000\,{\scriptsize[$-$.001,+.001]} & +.005 & +.005 & \phantom{+}.000 \\
Qwen3-30B & 1,714 & \phantom{+}.000\,{\scriptsize[\phantom{+}.000,+.001]} & \phantom{+}.000 & \phantom{+}.000 & \phantom{+}.000 \\
\bottomrule
\end{tabular}
\end{adjustbox}
\caption{MedQA with TAB's generic lexicon and the agreement setting chosen on the other corpora (SACC-LODO), against the reported \sacc{}, in one pass per adversary on every document; SACC-mask: the same lexicon swap for SACC-mask. Positive: the swapped system leaks more. Brackets: 99\% interval of SACC-LODO $-$ \sacc{} on $L$; bold estimates exclude zero.}
\label{tab:lodo}
\end{table}
% <<< GENERATED lodo

\section{Judged Audit}
\label{app:judges}
Three LLM judges read 180 items drawn from a sample fixed in advance (60 per corpus), each a passage with one of its
subjects and the propositions it states, on the released texts. They judge which of the propositions a text still states
(81 passages with reference facts) and whether an edited passage carries a grammatical or fluency fault (140 to 145
passages). GPT-4.1 was designated the primary judge before any output: it is the strongest judge in our configuration and
the only one that is neither an adversary nor the LLM rewrite's model. \mbox{GPT-4o-mini} and \mbox{Qwen3-30B-A3B} are reported
beside it, and no per-item majority is formed. Each judge also reads the unedited input as a control: on it, a valid
judge keeps at least .70 of the reference facts and no fewer than on an edited text, and flags the original words at
\sacc{}'s edit positions no more often than the edits themselves. Table~\ref{tab:judgev2} gives the levels.

% >>> GENERATED judgev2
\begin{table}[htbp]
\centering
\small
\setlength{\tabcolsep}{2.4pt}
\begin{adjustbox}{max width=\linewidth}
\begin{tabular}{l ccc ccc l}
\toprule
& \multicolumn{3}{c}{Facts kept} & \multicolumn{3}{c}{Passages with a fault} & \\
\cmidrule(lr){2-4} \cmidrule(lr){5-7}
Judge & Input & Mask & \sacc{} & Input & Mask & \sacc{} & Control \\
\midrule
GPT-4.1 (primary) & .665 & .566 & .624 & .032 & .044 & .016 & fails G1 \\
GPT-4o-mini & .905 & .876 & .840 & .241 & .121 & .257 & fails G2 \\
Qwen3-30B-A3B & .873 & .801 & .816 & .195 & .180 & .288 & passes \\
\bottomrule
\end{tabular}
\end{adjustbox}
\caption{Judged audit on 81 passages with reference facts and 139 to 145 cue-edit passages (passages a judge's reply could not be parsed for are left out); Mask is SACC-mask. The control asks the unedited input to keep at least .70 of the facts and no fewer than the edited texts (G1), and to carry no more faults than they do (G2).}
\label{tab:judgev2}
\end{table}
% <<< GENERATED judgev2

Two of the three judges fail the control, in opposite directions. GPT-4.1 finds only .665 of the reference facts in the
unedited input, so its fact labels are too strict to measure loss. \mbox{GPT-4o-mini} flags the original words at \sacc{}'s
edit positions about as often as \sacc{}'s rewrites (.241 against .257), and twice as often as the redactions, so its
flags do not separate an edit from the text around it. \mbox{Qwen3-30B-A3B} passes. Under it, \sacc{} keeps as many facts as
SACC-mask ($+.015$, 99\% interval $[-.000,+.038]$) and has more flagged passages than SACC-mask ($+.108$,
$[+.039,+.182]$) and than the input ($+.094$, $[+.013,+.176]$). Most of these flags fall on epicene pronouns in standard
use: on MedQA and TAB, 36 of its 60 agreement flags pair \emph{they} with a plural verb and none pairs it with a singular
verb, and 105 of its 138 possessive or reference flags are on a span that holds an epicene pronoun, such as
\emph{their lungs are clear}. Because the judge designated in advance fails the control, we draw no conclusion from the judges;
read on its own, the one judge that passes finds no loss of facts and a fluency cost that lies mostly in singular
\emph{they} itself. Under GPT-4.1, \sacc{} has fewer flagged passages than SACC-mask ($-.027$, $[-.061,-.003]$); under
\mbox{GPT-4o-mini} it has more ($+.136$, $[+.070,+.208]$) and as many as the input. The judges agree on 73\% to 87\% of fact
labels. An earlier audit with \mbox{GPT-4o-mini} and \mbox{Qwen3-30B-A3B}, on the texts released before the rewrite-step fix and
without the input control, found about twice as many flagged passages for \sacc{} as for SACC-mask (.312 / .268 against
.165 / .133), with the same composition: of the spans flagged as agreement errors, 6 of 72 and 0 of 112 contain
\emph{they} followed by a singular verb, and 81 of 122 and 21 of 32 possessive or reference flags are the bare word
\emph{their}.

\section{Worked Examples}
\label{app:examples}
Tables~\ref{tab:example-medqa} and~\ref{tab:example-tab} show one MedQA document and one TAB passage under six systems.
They were chosen by a fixed rule, not by hand: among the documents of the LLM rewrite's first samples (200 of MedQA,
91 of TAB) on which \sacc{}'s text differs from the global ladder's, with MedQA restricted to inputs of 350 to 650 characters so that a vignette fits, one
document is drawn with a seeded generator. MedQA vignettes are shown whole; for TAB, the passage runs from 15 words before
the first word that \sacc{} and the global ladder edit differently to 30 words after it, and each system's text is aligned
to that span of the input. They illustrate behaviour and say nothing about averages.

In the MedQA vignette, the two masking systems redact every pronoun and both mentions of the brother, and the global
ladder and \sacc{} rewrite them to epicene forms. Both repair the verb after \emph{they}, but only \sacc{}, with the full
agreement setting, also repairs the second verb of the clause (\emph{They have no similar symptoms and are in good
health}), where the ladder leaves \emph{and is}. The LLM rewrite redacts the pronouns instead of rewriting them, keeps
\emph{brother} twice and also redacts the patient's age. In the TAB passage, both global editors edit
\emph{Government}, a use of an occupational term that refers to no person, and the ladder's replacement breaks the
sentence's agreement; \sacc{} and SACC-mask leave it. The LLM rewrite redacts the title \emph{Ambassador} of the
government's agent, which \sacc{}'s lexicon does not list.

% >>> GENERATED examples
\begin{table*}[tbp]
\centering
\footnotesize
\setlength{\tabcolsep}{4pt}
\begin{adjustbox}{max width=\linewidth}
\begin{tabular}{l p{0.80\textwidth}}
\toprule
System & Text \\
\midrule
Anonymised input & A 35-year-old [SEX] comes to the [OCCUPATION] for genetic counseling prior to conception. Her younger brother has mild developmental delay, pale complexion, and has to limit his dietary intake of phenylalanine. She has no similar symptoms and is in good health. Her parents are healthy. Which of the following is the most appropriate assessment of her [OCCUPATION] status for the disease affecting her brother? \\
\midrule
Global mask & A 35-year-old [SEX] comes to the [OCCUPATION] for genetic counseling prior to conception. \textbf{[REDACTED]} younger \textbf{[REDACTED]} has mild developmental delay, pale complexion, and has to limit \textbf{[REDACTED]} dietary intake of phenylalanine. \textbf{[REDACTED]} has no similar symptoms and is in good health. \textbf{[REDACTED]} parents are healthy. Which of the following is the most appropriate assessment of \textbf{[REDACTED]} [OCCUPATION] status for the disease affecting \textbf{[REDACTED]} \textbf{[REDACTED]?} \\
\midrule
Global ladder & A 35-year-old [SEX] comes to the [OCCUPATION] for genetic counseling prior to conception. \textbf{Their} younger \textbf{sibling} has mild developmental delay, pale complexion, and has to limit \textbf{their} dietary intake of phenylalanine. \textbf{They} \textbf{have} no similar symptoms and is in good health. \textbf{Their} parents are healthy. Which of the following is the most appropriate assessment of \textbf{their} [OCCUPATION] status for the disease affecting \textbf{their} \textbf{sibling?} \\
\midrule
SACC-mask & A 35-year-old [SEX] comes to the [OCCUPATION] for genetic counseling prior to conception. \textbf{[REDACTED]} younger \textbf{[REDACTED]} has mild developmental delay, pale complexion, and has to limit \textbf{[REDACTED]} dietary intake of phenylalanine. \textbf{[REDACTED]} has no similar symptoms and is in good health. \textbf{[REDACTED]} parents are healthy. Which of the following is the most appropriate assessment of \textbf{[REDACTED]} [OCCUPATION] status for the disease affecting \textbf{[REDACTED]} \textbf{[REDACTED]?} \\
\midrule
\rowcolor{oursrow}SACC & A 35-year-old [SEX] comes to the [OCCUPATION] for genetic counseling prior to conception. \textbf{Their} younger \textbf{sibling} has mild developmental delay, pale complexion, and has to limit \textbf{their} dietary intake of phenylalanine. \textbf{They} \textbf{have} no similar symptoms and \textbf{are} in good health. \textbf{Their} parents are healthy. Which of the following is the most appropriate assessment of \textbf{their} [OCCUPATION] status for the disease affecting \textbf{their} \textbf{sibling?} \\
\midrule
LLM rewrite & A \textbf{[REDACTED]} [SEX] comes to the [OCCUPATION] for genetic counseling prior to conception. \textbf{[REDACTED]} younger brother has mild developmental delay, pale complexion, and has to limit \textbf{[REDACTED]} dietary intake of phenylalanine. \textbf{[REDACTED]} has no similar symptoms and is in good health. \textbf{[REDACTED]} parents are healthy. Which of the following is the most appropriate assessment of \textbf{[REDACTED]} [OCCUPATION] status for the disease affecting \textbf{[REDACTED]} brother? \\
\bottomrule
\end{tabular}
\end{adjustbox}
\caption{MedQA document \texttt{MEDQA-01206}, drawn by the fixed rule of this appendix from 25 candidates. Bold: words the system wrote that are not in the input.}
\label{tab:example-medqa}
\end{table*}

\begin{table*}[tbp]
\centering
\footnotesize
\setlength{\tabcolsep}{4pt}
\begin{adjustbox}{max width=\linewidth}
\begin{tabular}{l p{0.80\textwidth}}
\toprule
System & Text \\
\midrule
Anonymised input & \ldots{} European Commission of Human Rights (``the Commission'') under former Article 25 by the applicant. The Government of Austria are represented by their Agent, Ambassador [PERSON], Head of the International Law Department at the Federal Ministry of Foreign Affairs. The applicant's request to the Court referred \ldots \\
\midrule
Global mask & \ldots{} European Commission of Human Rights (``the Commission'') under former Article 25 by the applicant. The \textbf{[REDACTED]} of Austria are represented by their Agent, Ambassador [PERSON], Head of the International Law Department at the Federal Ministry of Foreign Affairs. The applicant's request to the Court referred \ldots \\
\midrule
Global ladder & \ldots{} European Commission of Human Rights (``the Commission'') under former Article 25 by the applicant. The \textbf{public} \textbf{official} of Austria are represented by their Agent, Ambassador [PERSON], Head of the International Law Department at the Federal Ministry of Foreign Affairs. The applicant's request to the Court referred \ldots \\
\midrule
SACC-mask & \ldots{} European Commission of Human Rights (``the Commission'') under former Article 25 by the applicant. The Government of Austria are represented by their Agent, Ambassador [PERSON], Head of the International Law Department at the Federal Ministry of Foreign Affairs. The applicant's request to the Court referred \ldots \\
\midrule
\rowcolor{oursrow}SACC & \ldots{} European Commission of Human Rights (``the Commission'') under former Article 25 by the applicant. The Government of Austria are represented by their Agent, Ambassador [PERSON], Head of the International Law Department at the Federal Ministry of Foreign Affairs. The applicant's request to the Court referred \ldots \\
\midrule
LLM rewrite & \ldots{} European Commission of Human Rights (``the Commission'') under former Article 25 by the applicant. The Government of Austria are represented by their Agent, \textbf{[REDACTED]} [PERSON], Head of the International Law Department at the Federal Ministry of Foreign Affairs. The applicant's request to the Court referred \ldots \\
\bottomrule
\end{tabular}
\end{adjustbox}
\caption{TAB document \texttt{TAB-e666e}, drawn by the fixed rule of this appendix from 91 candidates. Bold: words the system wrote that are not in the input.}
\label{tab:example-tab}
\end{table*}
% <<< GENERATED examples

\section{Version History}
\label{app:history}
The reported system is described once, in Section~\ref{sec:method} and Appendices~\ref{app:taskeditor} and~\ref{app:impl}. Three implementation defects, two in the rewrite step and one in the binder, were found and fixed in two rounds, each pre-specified before it was scored. Every result uses the texts fixed in both rounds except these, each marked where it appears: the configuration comparison (Table~\ref{tab:config}) and the reference counts of Table~\ref{tab:reftypes}, scored before both rounds, and the single-reference, upstream and held-out LLM-rewrite comparisons (Tables~\ref{tab:fixedref}, \ref{tab:upstream} and~\ref{tab:llmheld}), scored before the binder fix. The single-reference reading also supplies the subject-reference row of Table~\ref{tab:choices} and the reference panel of Figure~\ref{fig:choices}, which set a contrast on the fixed texts beside one on the texts before the binder fix (bounded in Appendix~\ref{app:robust}). Figure~\ref{fig:multiverse} reads both references on the texts before the fix: on one version the reference changes 3 of 45 all-pairs verdicts, all involving Adaptive, whereas mixing versions doubled the count, because the fix alone changes three TAB \leakE{} verdicts. This appendix records what the fixes changed.

% >>> GENERATED examplesnote
\paragraph{Two defects of the rewrite step, fixed.}
The texts released before this evaluation had two defects in the rewrite step, both corrected in every result reported here. First, the test that chooses between the possessive and the pronoun read only a following word of letters, so possessive \emph{his} and \emph{her} before a placeholder or a number became \emph{theirs} and \emph{them} (136 and 125 times on MedQA, 7 and 0 on TAB, none on SynthPAI). Second, the rewrite table was consulted by surface form, so a marital term could become \emph{spouse} or \emph{partner}, which still states a relationship status and is therefore not admissible (21 times on TAB, never elsewhere); those cues now stay redacted. The fix changes only these words. Table~\ref{tab:fixeffect} gives its effect under each adversary.
% <<< GENERATED examplesnote

Over each adversary's whole cohort (Table~\ref{tab:fixeffect}), on $L$ no estimate exceeds .003 in absolute value. On
\discE{} and \leakE{} one cell, Llama on TAB, moves by $+.006$ with an upper bound of .024; a prediction pre-specified
before the fix was scored had excluded a move of that size, and we record it as failed. \unli{} and TPS change by at
most .002. With the released texts, the global ladder's lead over \sacc{} on MedQA was separable on $L$ and \leakE{};
after this fix it was separable on $L$ alone, and after the binder fix below on no measure.

% >>> GENERATED fixeffect
\begin{table*}[tbp]
\centering
\small
\setlength{\tabcolsep}{3.4pt}
\begin{adjustbox}{max width=\linewidth}
\begin{tabular}{ll r ccc}
\toprule
Adversary & Corpus & Changed & $\Delta L$ & $\Delta$\textsc{Disc} & $\Delta$\textsc{Leak} \\
\midrule
\multirow{2}{*}{GPT-4o-mini} & MedQA & 642 & \phantom{+}.000\,{\scriptsize[$-$.003,+.004]} & $-$.003\,{\scriptsize[$-$.009,+.002]} & $-$.003\,{\scriptsize[$-$.009,+.002]} \\
 & TAB & 89 & +.001\,{\scriptsize[$-$.001,+.002]} & $-$.001\,{\scriptsize[$-$.008,+.005]} & $-$.002\,{\scriptsize[$-$.009,+.003]} \\
\midrule
\multirow{2}{*}{Llama-3.1-8B} & MedQA & 655 & $-$.002\,{\scriptsize[$-$.004,+.001]} & +.001\,{\scriptsize[$-$.010,+.014]} & +.001\,{\scriptsize[$-$.008,+.013]} \\
 & TAB & 82 & +.001\,{\scriptsize[\phantom{+}.000,+.002]} & +.006\,{\scriptsize[\phantom{+}.000,+.024]} & +.006\,{\scriptsize[\phantom{+}.000,+.025]} \\
\midrule
\multirow{2}{*}{Qwen3-30B} & MedQA & 616 & $-$.002\,{\scriptsize[$-$.006,\phantom{+}.000]} & $-$.004\,{\scriptsize[$-$.015,+.004]} & $-$.002\,{\scriptsize[$-$.011,+.003]} \\
 & TAB & 91 & $-$.003\,{\scriptsize[$-$.007,+.001]} & $-$.001\,{\scriptsize[$-$.009,+.004]} & $-$.001\,{\scriptsize[$-$.011,+.005]} \\
\bottomrule
\end{tabular}
\end{adjustbox}
\caption{Fixed minus released \sacc{} over the whole cohort of each adversary (positive: the fixed texts leak more). Changed: entries on documents the fix altered. Brackets are 99\% intervals; bold estimates exclude zero.}
\label{tab:fixeffect}
\end{table*}
% <<< GENERATED fixeffect

\paragraph{The binder fix.}
The first audit of Appendix~\ref{app:impl} found sex and relationship nouns naming a protected relative sent to another
person. The mention-form rule repairs this; it changes 114 MedQA and 24 TAB documents, and
pre-specified gates checked that nothing else changes. It lowers MedQA $L$ under all three adversaries (by .007, .012 and
.007, intervals clear of zero) and moves no TAB $L$ separably (Table~\ref{tab:fix3effect}); under Qwen it raises TAB's
leaking rate ($+.024$, $[+.001,+.052]$), and on MedQA it lowers SACC-mask's \unli{} by .006, beyond a pre-specified bound of
.005. Three main-text readings changed with it: the global ladder's MedQA lead on $L$ ($-.015$, $[-.030,-.004]$) is no
longer separable, \sacc{} became separably more private than global masking on MedQA $L$, and \sacc{}'s TAB reduction of
its input's \leakE{} rate is no longer separable from zero. The comparisons with the LLM rewrite on its samples under the held-out
adversaries (Table~\ref{tab:llmheld}) predate the fix.

% >>> GENERATED fix3effect
\begin{table*}[tbp]
\centering
\small
\setlength{\tabcolsep}{3.0pt}
\begin{adjustbox}{max width=\linewidth}
\begin{tabular}{ll r ccc c}
\toprule
& & & \multicolumn{3}{c}{\sacc{}} & SACC-mask \\
\cmidrule(lr){4-6} \cmidrule(lr){7-7}
Adversary & Corpus & Changed & $\Delta L$ & $\Delta$\textsc{Disc} & $\Delta$\textsc{Leak} & $\Delta L$ \\
\midrule
\multirow{2}{*}{GPT-4o-mini} & MedQA & 448 & \textbf{$-$.007}\,{\scriptsize[$-$.012,$-$.003]} & $-$.002\,{\scriptsize[$-$.009,+.004]} & \textbf{$-$.006}\,{\scriptsize[$-$.013,\phantom{+}.000]} & \textbf{$-$.006}\,{\scriptsize[$-$.011,$-$.003]} \\
 & TAB & 121 & +.003\,{\scriptsize[$-$.007,+.016]} & $-$.014\,{\scriptsize[$-$.052,+.002]} & +.014\,{\scriptsize[$-$.005,+.043]} & +.007\,{\scriptsize[$-$.009,+.029]} \\
\midrule
\multirow{2}{*}{Llama-3.1-8B} & MedQA & 528 & \textbf{$-$.012}\,{\scriptsize[$-$.019,$-$.006]} & $-$.003\,{\scriptsize[$-$.014,+.006]} & \textbf{$-$.013}\,{\scriptsize[$-$.024,$-$.003]} & \textbf{$-$.012}\,{\scriptsize[$-$.019,$-$.006]} \\
 & TAB & 103 & +.020\,{\scriptsize[$-$.004,+.050]} & +.011\,{\scriptsize[\phantom{+}.000,+.034]} & +.029\,{\scriptsize[\phantom{+}.000,+.070]} & +.011\,{\scriptsize[$-$.002,+.031]} \\
\midrule
\multirow{2}{*}{Qwen3-30B} & MedQA & 452 & \textbf{$-$.007}\,{\scriptsize[$-$.011,$-$.003]} & +.006\,{\scriptsize[\phantom{+}.000,+.012]} & $-$.003\,{\scriptsize[$-$.009,+.002]} & \textbf{$-$.009}\,{\scriptsize[$-$.013,$-$.005]} \\
 & TAB & 112 & +.018\,{\scriptsize[\phantom{+}.000,+.043]} & +.001\,{\scriptsize[$-$.018,+.023]} & \textbf{+.024}\,{\scriptsize[+.001,+.052]} & +.017\,{\scriptsize[$-$.001,+.042]} \\
\bottomrule
\end{tabular}
\end{adjustbox}
\caption{The binder fix (a cue that names a protected relative is that relative's cue), fixed minus before over the whole cohort of each adversary (negative: the fixed texts leak less). Changed: entries on the documents the fix alters. Brackets are 99\% intervals; bold estimates exclude zero.}
\label{tab:fix3effect}
\end{table*}
% <<< GENERATED fix3effect

\paragraph{Held-out families before both rounds.}
A pass on the texts before both rounds, with \sacc{} in the full setting on MedQA, had
read the gain of rewriting over masking as separable under Llama on MedQA ($+.038$) and as null under Llama on TAB; on the reported texts it is
null on MedQA and reversed on TAB.

\section{Human Evaluation}
\label{app:human}
\paragraph{Design.}
Automatic content scores do not see grammaticality, and two of three LLM judges failed their controls
(Appendix~\ref{app:judges}), so people judged passages around \sacc{}'s edits. Thirty MedQA and thirty TAB documents on
which \sacc{}'s text differs from the global ladder's were drawn with a fixed seed; from each, one of \sacc{}'s edit
sites was drawn, and the input sentence holding it, with one sentence on each side (at most 120 words), was aligned to
\sacc{}'s, SACC-mask's and the global ladder's text. Annotators saw the input passage and the three versions in a random
order, with the words that differ listed, never a system name or score, and labelled each version on two scales:
\textsc{Facts}, whether it still states all (2), some (1) or little (0) of the input's content, where losing a word that
only states a person's sex, relationship status or occupation counts as no loss; and \textsc{Grammar}, no error (3), a minor
error (2) or an error that hinders reading (1), where singular \emph{they} with plural agreement counts as grammatical.
One passage in seven was an instructed attention check, and every annotator passed all of them. Three volunteers,
fluent in English and not involved in this work, annotated the passages after reading the guidelines summarised here.
They were recruited personally by one of the authors from outside the project and were not paid. Each passage was labelled by all three, and contrasts are paired by passage with two-sided 95\%
cluster-bootstrap intervals; agreement is Krippendorff's $\alpha$ \citep{krippendorff-2004-content}. The tests fixed before
annotation were that \sacc{} keeps more \textsc{Facts} than the global ladder and than SACC-mask, and that its
\textsc{Grammar} is non-inferior to SACC-mask's at a margin of 0.2. The passages were labelled in two halves of thirty;
each half was first analysed with two annotators, and the third annotator's labels, added afterwards, change no test's
outcome.

\paragraph{Results.}
\label{sec:human}

% >>> GENERATED human
\begin{table*}[tbp]
\centering
\small
\setlength{\tabcolsep}{3pt}
\begin{adjustbox}{max width=\linewidth}
\begin{tabular}{l ccc ccc}
\toprule
& \multicolumn{3}{c}{\textsc{Facts}} & \multicolumn{3}{c}{\textsc{Grammar}} \\
\cmidrule(lr){2-4} \cmidrule(lr){5-7}
System & All & MedQA & TAB & All & MedQA & TAB \\
\midrule
SACC-mask & 1.90 & 1.91 & 1.89 & 3.00 & 3.00 & 3.00 \\
\rowcolor{oursrow}SACC & 1.94 & 1.99 & 1.89 & 2.88 & 2.87 & 2.90 \\
Global ladder & 1.70 & 1.69 & 1.71 & 2.45 & 2.47 & 2.43 \\
\midrule
\sacc{} $-$ global ladder & +.24\,{\scriptsize[+.12,+.36]} & +.30\,{\scriptsize[+.13,+.47]} & +.18\,{\scriptsize[$-$.01,+.37]} & +.43\,{\scriptsize[+.26,+.61]} & +.40\,{\scriptsize[+.22,+.59]} & +.47\,{\scriptsize[+.18,+.78]} \\
\sacc{} $-$ SACC-mask & +.04\,{\scriptsize[+.01,+.08]} & +.08\,{\scriptsize[+.01,+.16]} & .00\,{\scriptsize[.00,.00]} & $-$.12\,{\scriptsize[$-$.20,$-$.05]} & $-$.13\,{\scriptsize[$-$.27,$-$.03]} & $-$.10\,{\scriptsize[$-$.23,.00]} \\
\bottomrule
\end{tabular}
\end{adjustbox}
\caption{Human evaluation of 60 passages around \sacc{}'s edits (30 MedQA, 30 TAB), each version labelled by three annotators: means, then paired differences with 95\% intervals. \textsc{Facts} 0--2, \textsc{Grammar} 1--3; Krippendorff's $\alpha$ .83 and .92.}
\label{tab:human}
\end{table*}
% <<< GENERATED human

Table~\ref{tab:human} summarises the mean ratings and paired differences. \sacc{} preserves more content than the global ladder,
with the largest improvements on MedQA, and achieves higher grammaticality on both corpora. Annotators primarily attribute
content loss under the ladder to over-generalised non-person nouns (e.g., \emph{clinic} $\rightarrow$ \emph{workplace}, \emph{army}
$\rightarrow$ \emph{public servant}) and the resulting degraded noun phrases. Compared to SACC-mask, rewriting retains
slightly more information (separable on MedQA, not on TAB) but is marginally less
grammatical, so it fails the pre-specified non-inferiority test. The remaining errors are mostly agreement issues
(e.g., occasional singular verbs following \emph{they}), which do not arise under redaction.

\fi

\end{document}